\documentclass[11pt]{article}
\usepackage{algorithm}
\usepackage{verbatim}
\usepackage[noend]{algpseudocode}

\usepackage{amsmath}
\usepackage{tikz}
\usetikzlibrary{positioning}
\usepackage{amsmath,amsfonts,amsthm}
\usepackage{xspace}
\usepackage{xcolor}
\usepackage{url}
\usepackage{caption}
\usepackage{subcaption}
\usepackage{tikz}
\usepackage{paralist}
\usepackage{libertine}
\usepackage{dsfont}
\usepackage[
pagebackref,
colorlinks=true,
urlcolor=blue,
linkcolor=blue,
citecolor=blue,
]{hyperref}
\usepackage[nameinlink]{cleveref}
\usepackage{enumitem}
\usepackage{todonotes}

\usepackage{color-edits} 
\addauthor{rdk}{blue}
\addauthor{pco}{purple}

\usepackage[margin=1in]{geometry}
\usepackage{txfonts}
\usepackage[T1]{fontenc}

\theoremstyle{definition}
\newtheorem{definition}{Definition}[section]

\newtheorem{theorem}{Theorem}[section]
\newtheorem{lemma}[theorem]{Lemma}

\newtheorem{corollary}[theorem]{Corollary}

\newtheorem{question}[theorem]{Question}

\DeclareMathOperator{\sgn}{sign}
\newcommand{\ue}[2]{{#1}^{(#2)}}

\newcommand{\eps}{\varepsilon}

\newcommand{\ind}{\mathbb{I}}

\newcommand{\II}{\mathbb{I}} 
\DeclareMathOperator*{\E}{\mathbb{E}}
\newcommand{\var}{\mathrm{Var}}

\newcommand{\calerr}{\mathrm{calerr}}   
\newcommand{\Int}{\mathcal{I}}      
\newcommand{\Ber}{\mathrm{Ber}}

\newcommand{\sprinstance}{\textsf{SPRInstance}}      
\newcommand{\spr}{\mathrm{SPR}}     
\newcommand{\opt}{\mathrm{opt}}     
\newcommand{\uspr}{\mathrm{USPR}}     

\newcommand{\maxfish}[1]{{\color{red} Max: {#1}}}
\newcommand{\noah}[1]{}
\newcommand{\ep}{\epsilon}
\newcommand{\BN}{\mathbb{N}}
\newcommand{\BR}{\mathbb{R}}
\newcommand{\MQ}{\mathcal{Q}}
\newcommand{\MP}{\mathcal{P}}

\newcommand{\Rbias}{R_{\mathsf{bias}}}
\newcommand{\Rvar}{R_{\mathsf{var}}}
\newcommand{\Var}{\mathrm{Var}}

\newcommand{\PlayerP}{Player-P\xspace}
\newcommand{\PlayerL}{Player-L\xspace}
\newcommand{\calE}{E}
\newcommand{\interval}{\mathrm{interval}}
\newcommand{\simulateGame}{\texttt{simulateGame}}
\newcommand{\probub}{\mathrm{prob}}
\newcommand{\biasub}{\mathrm{bias}}

\newcommand{\calD}{E}
\newcommand{\tRbias}{\tilde{R}_{\mathsf{bias}}}
\newcommand{\tRvar}{\tilde{R}_{\mathsf{var}}}

\newcommand{\AlgA}{\textsf{A}\xspace}
\newcommand{\AlgB}{\textsf{B}\xspace}
\newcommand{\recentB}{\text{recentB}\xspace}
\newcommand{\cnt}{\text{count}\xspace}
\newcommand{\placeSign}{\textsf{label}\xspace}
\newcommand{\initialize}{\textsf{initialize}\xspace}
\newcommand{\prevHalf}{\text{prevHalf}}
\newcommand{\cntHalf}{\text{countHalf}}
\newcommand{\phase}{\text{phase}}
\newcommand{\half}{\text{half}}
\newcommand{\remainingSigns}{\textsf{remainingSigns}}
\newcommand{\signs}{\textsf{remainingSigns}}
\newcommand{\executionSteps}{\textsf{executionSteps}}
\newcommand{\vep}{\varepsilon}

\makeatletter
\AtBeginDocument{%
  \providecommand*{\theHALG@line}{}%
  \renewcommand*{\theHALG@line}{\thealgorithm.\arabic{ALG@line}}%
  \renewcommand*{\ALG@step}{%
    \refstepcounter{ALG@line}%
    \expandafter\@cref@getprefix\cref@currentlabel\@nil\cref@currentprefix%
    \xdef\cref@currentprefix{\cref@currentprefix}%
    \addtocounter{ALG@rem}{1}%
    \ifthenelse{\equal{\arabic{ALG@rem}}{\ALG@numberfreq}}%
      {\setcounter{ALG@rem}{0}\alglinenumber{\arabic{ALG@line}}}%
      {}%
  }%
  \crefname{ALG@line}{line}{lines}%
  \Crefname{ALG@line}{Line}{Lines}%
}
\makeatother

\title{Breaking the $T^{2/3}$ Barrier for Sequential Calibration}
\date{}
\author{Yuval Dagan \\ {\small Tel Aviv University} \\ \small{\url{ydagan@tauex.tau.ac.il}} \and Constantinos Daskalakis \\ {\small MIT \& Archimedes AI} \\ \small{\url{costis@csail.mit.edu}} \and Maxwell Fishelson \\ {\small MIT} \\ \small{\url{maxfish@mit.edu}} \and Noah Golowich \\ {\small MIT} \\ \small{\url{nzg@mit.edu}} \and Robert Kleinberg \\ {\small Cornell University} \\ \small{\url{rdk@cs.cornell.edu}} \and Princewill Okoroafor \\ {\small Cornell University} \\ \small{\url{pco9@cornell.edu}}}

\begin{document}

\begin{titlepage}
\clearpage\maketitle
\thispagestyle{empty}



\begin{abstract}
A set of probabilistic forecasts is \emph{calibrated} if each prediction of the forecaster closely approximates the empirical distribution of outcomes on the subset of timesteps where that prediction was made. We study the fundamental problem of online calibrated forecasting of binary sequences under the standard $\ell_1$ calibration error metric, which was initially studied by \cite{foster1998asymptotic}. They derived an algorithm with $O(T^{2/3})$ calibration error after $T$ time steps, and showed a lower bound of $\Omega(T^{1/2})$. These bounds remained stagnant for two decades, until \cite{qiao2021stronger} improved the lower bound to $\Omega(T^{0.528})$ by introducing a combinatorial game called \emph{sign preservation} and showing that lower bounds for this game imply lower bounds for calibration.

In this paper, we give the first improvement to the $O(T^{2/3})$ upper bound on calibration error of \cite{foster1998asymptotic}.
We do this by introducing a variant of \cite{qiao2021stronger}'s game that we call {\em sign preservation with reuse (SPR)}. We prove that the relationship between SPR and calibrated forecasting is bidirectional: not only do lower bounds for SPR translate into lower bounds for calibration, but algorithms for SPR also translate into new algorithms for calibrated forecasting. We then give an improved \emph{upper bound} for the SPR game, which implies, via our equivalence, 
a forecasting algorithm with calibration error $O(T^{2/3 - \eps})$ for some $\eps > 0$, improving \cite{foster1998asymptotic}'s upper bound for the first time. Using similar ideas, we then prove a slightly stronger lower bound than that of \cite{qiao2021stronger}, namely $\Omega(T^{0.54389})$. Our lower bound is obtained by an \emph{oblivious} adversary, marking the first $\omega(T^{1/2})$ calibration lower bound for oblivious adversaries.



\end{abstract}

\end{titlepage}


\section{Introduction}
The notion of \emph{calibration} when forecasting an event is a fundamental concept in statistical theory with longstanding theoretical \cite{foster1998asymptotic,dawid1982bayesian,oakes1985self,murphy1967verification} and empirical \cite{murphy1977reliability,brier1950verification} foundations. A forecaster who repeatedly predicts probabilities for some event to occur (e.g., a weather forecaster predicting the probability of rain each day) is said to be \emph{calibrated} if their forecasts match up with the proportion of times the event actually occurs. Producing calibrated forecasts is a very natural desideratum: for instance,  of the days when a weather forecaster predicts a ``$10\%$ chance of rain'', we would hope that it rains on roughly $10 \%$ of them.

Calibration has found a plethora of applications in modern machine learning. For instance, a recently proposed criterion for evaluating algorithmic fairness of classifiers is that the classifer's predictions be calibrated on all subgroups (e.g., by ethnicity). This idea is encapsulated by the concept of \emph{multicalibration} \cite{herbert2018multicalibration}, and there has been a flurry of recent work to obtain algorithms satisfying multicalibration as well as related notions, such as omniprediction \cite{kleinberg2017inherent,pleiss2017fairness,shabat2020sample,jung2020moment,garg2024oracle,zhao2020individual}.  
From a more empirical perspective, there has been significant interest in evaluating the degree to which modern machine learning models generate calibrated predictions, such as for image classification \cite{kuleshov2015calibrated,guo2017calibration,kumar2019verified,minderer2021revisiting}. Additionally, producing calibrated prognoses has attracted much interest in the study of medicine \cite{jiang2012calibrating,crowson2016assessing}, and calibration has found many other uses as well \cite{murphy1967verification,kleinberg2023ucalibration,blasiok2023when}. 
We emphasize that calibration is a distinct notion from accuracy: it is possible for perfectly calibrated forecasts to be either perfectly accurate or highly inaccurate.\footnote{For example, for some time horizon $T$, if it rains on the first $T/2$ days and then does not rain on the last $T/2$ days, then both of the following forecasters are perfectly calibrated: (a) predicting $100\%$ chance of rain on the first $T/2$ days and $0\%$ chance of rain on the last $T/2$ days, and (b) predicting $50\%$ chance of rain on all days. However, the first forecaster is more accurate.} 

\paragraph{Sequential calibration} In this paper, we study the classical \emph{sequential calibration} problem, which dates back to some of the earliest work on calibration \cite{foster1998asymptotic,dawid1982bayesian}. The problem can be described as the following game between two players, a \emph{forecaster} and an \emph{adversary}, which operates over some number $T \in \BN$ of time steps. 
At each {time step} $t \in [T]$, the forecaster chooses a \emph{prediction} $p_t \in [0,1]$ and the adversary independently chooses an \emph{outcome} $y_t \in \{0,1\}$, and each then observes the other player's choice. The prediction $p_t$ should be interpreted as the probability the forecaster believes the outcome will be $y_t = 1$. The forecaster's goal is to minimize the overall \emph{calibration error} of their predictions against a worst-case adversary, defined as follows: 
\begin{align}
\calerr(T) := \sum_{p \in [0,1]} |m_T(p) - p \cdot n_T(p)|\label{eq:def-calerr},
\end{align}
where $n_T(p)$ is the total number of times the forecaster predicts $p$, and $m_T(p)$ is the number of those time steps for which the outcome was ``1''.\footnote{Note that the sum in \Cref{eq:def-calerr} is well-defined since the forecaster only predicts at most $T$ distinct values in $[0,1]$.} 

In this work, we focus on the $\ell_1$ calibration error in \Cref{eq:def-calerr}. Many other notions of calibration and related ``distance from calibration'' measures have also been studied, including $\ell_2$-calibration \cite{fishelson2025fullswap}, KL-calibration \cite{luo2025simultaneous}, distance-to-calibration and smooth measures \cite{kakade2008deterministic,qiao2024distance,arunachaleswaran2025elementary}, and decision-theoretic measures \cite{kleinberg2023ucalibration,hu2024predict,haghtalab2024truthfulness,qiao2025truthfulness,hartline2025perfectly}.

Aside from being perhaps the simplest model with which to study calibration, the sequential calibration problem has found direct applications to several of the areas mentioned above: for instance, using a connection between the notion of \emph{swap omniprediction}\footnote{Roughly speaking, a swap omnipredictor is a predictor which yields good predictions for every loss function in a given class, in a swap regret sense \cite{gopalan2023swap}.} and multicalibration, \cite{garg2024oracle} obtain lower bounds for online swap omniprediction algorithms by reducing to the vanilla sequential calibration problem. In particular, the lower bound of \cite{qiao2021stronger} implies (and our lower bound strengthens) lower bounds for online swap omniprediction, in a black-box manner.
Despite the centrality of sequential calibration, 
the following fundamental question has gone unanswered after over two decades: \emph{what is the optimal bound that can be obtained on $\calerr(T)$, as defined in \Cref{eq:def-calerr}?}

The best-known upper bound on $\calerr(T)$ was established by \cite{foster1998asymptotic} (see also \cite{hart2023minimax}), who showed that there is an algorithm for the forecaster which guarantees expected calibration error of at most $O(T^{2/3})$, against any adversary. The best known lower bound was $\Omega(T^{1/2})$ (which follows from a standard anticoncentration argument) until a recent result of \cite{qiao2021stronger}, who improved this slightly to $\Omega(T^{.528})$. 
The lower bound of \cite{qiao2021stronger} was proved by introducing the \emph{sign-preservation game}, which is a multi-round game between two players who repeatedly place and remove the signs (i.e., $+,-$) from a 1-dimensional grid (see \Cref{sec:spr}). The sign-preservation game has the advantage of being combinatorial in nature, which allowed \cite{qiao2021stronger} to (a) show bounds on the possible outcomes attainable in the sign-preservation game, and (b) show how these bounds translate into lower bounds for the sequential calibration problem. In this paper, one of our focuses is on improving calibration error bounds in the opposite direction:
\begin{question}
  \label{que:calerr}
\sloppy Is there a forecaster which guarantees expected calibration error $O(T^{2/3 - \ep})$, for some constant $\ep > 0$?
\end{question}



\paragraph{Our contributions} The sign-preservation game was a useful technical tool in \cite{qiao2021stronger} for obtaining lower bounds on calibration error, but it was unclear how fundamental it is. Our first contribution shows that it is fundamental in the following sense: the possible outcomes attainable in (a slight variant of) the sign-preservation game \emph{exactly characterize} the answer to \Cref{que:calerr}. In more detail, we consider a variant of the sign-preservation game (see \Cref{sec:spr}) in which a 1-dimensional grid consisting of $n$ \emph{cells} is given. At each of $s$ \emph{rounds}, one player chooses a cell $j \in [n]$, and the other player decides whether to place a $+$ or $-$ in that cell. They may also remove any $-$ signs to the left of $j$ or $+$ signs to the right of $j$. We let $\opt(n,s)$ be the maximum number of signs remaining at the end, assuming that the former player (who chooses $j$) tries to maximize this quantity and the latter player tries to minimize it.
Our first result, \Cref{col:main_eq}, tells us that the asymptotic behavior of $\opt(n,n)$ \emph{exactly characterizes}  the answer to \Cref{que:calerr}. 
\begin{theorem}[Equivalence]\label{col:main_eq}
\sloppy If there exists $\eps > 0$ such that for all $n \in \BN$, $\opt(n,n) \leq O(n^{1-\eps})$, then there exists a forecaster that guarantees  calibration error of $O(T^{\frac{2}{3} - \frac{\eps}{18}})$.
If instead $\opt(n,n^\alpha) \geq \Omega(n^\beta)$ for some constants $\alpha, \beta > 0$ and all $n \in \BN$, then there exists an adversary that ensures calibration error of at least $\tilde \Omega(T^{\frac{\beta+1}{\alpha+2}})$. 
\end{theorem}
Note that plugging in $\alpha = \beta = 1$ to the second part of the above theorem would yield a lower bound on calibration error of $\tilde \Omega(T^{2/3})$. As it turns out, such a statement for $\alpha = \beta = 1$ does not hold: our main result shows that $\opt(n,n) \leq O(n^{1-\ep})$ for some $\ep > 0$, which answers \Cref{que:calerr} in the affirmative.
\begin{theorem}[Upper bound for calibration]
  \label{thm:cal-ub}
There is $\eps > 0$ so that for all $n \in \BN$, $\opt(n,n) \leq O(n^{1-\eps})$. In particular, there is a forecaster guaranteeing calibration error of $O(T^{\frac 23 - \frac{\eps}{18}})$. 
\end{theorem}
The fact that $\opt(n,n) \leq O(n^{1-\eps})$ implies that there is a forecaster achieving calibration error of $O(T^{\frac{2}{3} - \frac{\eps}{18}})$ (per \Cref{col:main_eq}) is new to this work, though it uses similar intuitions to the work of \cite{qiao2021stronger} which showed the converse direction; see \Cref{sec:equivalence} for more details.  
The technical bulk of the proof of \Cref{thm:cal-ub} is establishing that $\opt(n,n) \leq O(n^{1-\eps})$. At a high level, the proof recursively constructs a strategy for the player deciding on the sign ($+$ or $-$) for each cell, using recursion both in the number of cells $n$ and the number of rounds $s$. A key insight is as follows: if the placement of a sign in, say, the right half of the cells (i.e., cells $[n/2,n]$) causes the deletion of many minus signs in the left half, then in following rounds we can \emph{bias} the signs placed towards being minus signs. This allows us to decrease the number of signs remaining at the end since (a) many minus signs were deleted and (b) of the remaining signs, fewer than ``expected'' were plus signs. We refer the reader to \Cref{sec:spr-intuition} for a more detailed proof overview.

The calibration adversary constructed in the proof of the \emph{lower bound} part of \Cref{col:main_eq}, as well as the one of \cite{qiao2021stronger}, is \emph{adaptive}, in the sense that its choices of outcomes are allowed to depend on past predictions of the forecaster. A weaker notion of adversary which has been mentioned in early works on calibration \cite{foster1998asymptotic} as well as having analogues in many other online learning settings \cite{cesabianchi2006prediction}, is that of an \emph{oblivious} adversary, whose outcomes \emph{cannot} depend on the forecaster's past predictions. Our final result gives the first oblivious adversary guaranteeing $\omega(T^{1/2})$ expected calibration error, and in fact a bound which is stronger than the $\Omega(T^{.528})$ bound of \cite{qiao2021stronger}: 
\begin{theorem}[Oblivious calibration adversary]
  \label{col:obl_lower}
There exists an \emph{oblivious} adversary for calibration which forces any forecaster to incur expected calibration error at least $\Omega(T^{0.54389})$. 
\end{theorem}
The proof of \Cref{col:obl_lower} proceeds by proving a lower bound for the sign-preservation game  
via an \emph{oblivious} strategy which satisfies some additional properties. Notice that the quantitative bound in \Cref{col:obl_lower} is stronger than the $\Omega(T^{0.528})$ bound of \cite{qiao2021stronger} (which was only shown for an adaptive adversary).  



\section{Preliminaries}

\subsection{Online Prediction and the $l_1$-calibration error}
We formally introduce the sequential calibration setting, which consists of a game between a forecaster and Nature (the adversary). The forecaster first chooses a finite set $P \subset [0,1]$ from which to select predictions. 
At each timestep $t \in [T]$, the forecaster makes a \emph{prediction} $p_t \in P$ and the adversary chooses an \emph{outcome} $y_t \in \{0,1\}$ without any knowledge of $p_t$. In choosing $p_t$, the forecaster may use knowledge of the adversary's past choices of outcomes $y_1, \ldots, y_{t-1}$. We consider two types of adversary:
\begin{itemize}[leftmargin=0.4cm]
\item An \emph{adaptive adversary} may use the full history $H_t = \{ (p_i, y_i) \}_{i=1}^{t-1}$ of outcomes  and predictions prior to time step $t$, when choosing $y_t$.
\item An \emph{oblivious adversary} can not use the learner's predictions $p_1, \ldots, p_{t-1}$ when choosing $y_t$, and so the distribution of $y_t$ can only depend on $y_1, \ldots, y_{t-1}$. Thus, an oblivious adversary may be equivalently represented as a joint distribution over tuples $(y_1, \ldots, y_T) \in \{0,1\}^T$. 
\end{itemize}
The forecaster aims to minimize the $\ell_1$-calibration error, denoted by $\calerr(T)$, of their predictions over the $T$ timesteps of the interaction. Formally, for $t \in [T]$, 
$
    \calerr (t) = \sum_{p \in P} | m_t (p) - n_t (p) \cdot p| ,
$
where $m_t(p) = \sum_{i=1}^t y_i \cdot \II [p_i = p]$ is the sum of outcomes when prediction $p$ was made, and $n_t(p) = \sum_{i=1}^t \II[p_i = p]$ is the number of timesteps when prediction $p$ was made. To track the calibration error over time, it is useful to define the quantity $\calE_t (p) = n_t (p) \cdot p - m_t(p)$, the (signed) error associated with prediction $p$ after $t$ timesteps. Since this value can be positive or negative, denote $\calE_t^+ (p) = \max \{ \calE_t (p), 0\}$ and $\calE_t^- (p) = \max \{ -\calE_t (p), 0\}$.
Then, $\calerr(t)$ can be equivalently written as
$
    \calerr(t)
=   \sum_{p \in P}|\calE_t(p)|
=   \sum_{p \in P}\calE^{+}_t(p) + \sum_{p \in P}\calE^{-}_t(p).
$

We use the binary sign convention $\sgn(x)=+1$ for $x\ge0$ and $\sgn(x)=-1$ for $x<0$.

\subsection{Sign-Preservation Game with Reuse}
\label{sec:spr}
We present the following deterministic two-player sequential game, called {\em Sign Preservation with Reuse (SPR)},
\xspace which generalizes the Sign-Preservation Game introduced by \cite{qiao2021stronger} by allowing the adversary to choose a cell multiple times as long as the cell is empty. An instance of $\spr$ with parameters $n,s\in \mathbb{N}$, denoted by $\sprinstance (n,s)$, proceeds as follows. There are two players, \PlayerP (for ``pointer'') and \PlayerL (for ``labeler''). At the beginning of the game, there are $n$ empty cells numbered $1, 2, \ldots, n$. The game consists of at most $s$ rounds, and in each round:
\begin{enumerate}[leftmargin=0.4cm]
    \item \PlayerP may terminate the game immediately.
    \item Otherwise, \PlayerP chooses (``points at'') an empty cell numbered $j \in [n]$.
    \item (Sign Removal) After knowing the value of $j$, \PlayerL may remove any minus (``$-$'') signs in cells to the left of cell $j$ or any plus (``$+$'') signs in cells to the right of cell $j$. (\PlayerL may remove signs from any subset of the indicated cells, including possibly the empty set.)
    \item (Sign Placement) \PlayerL places a \emph{sign} (either ``$+$'' or ``$-$'') into cell $j$.
\end{enumerate}
\PlayerP's goal is to maximize the number of preserved signs at the end of the game, while \PlayerL aims to minimize this number. Note that if we restrict \PlayerP to choose a cell at most once throughout the game and require \PlayerL to remove all signs that can be removed in each round, this game becomes the sign preservation game in \cite{qiao2021stronger}. 
We allow randomized strategies for \PlayerP and  \PlayerL. Define $\opt(n,s)$ to be the maximum number of expected preserved signs in $\sprinstance (n,s)$ (where the expectation is over the randomness in the players' strategies), assuming both players play optimally. 
Since SPR is a finite-horizon, perfect-information game with deterministic dynamics, the value $\opt(n,s)$ is unchanged if we restrict both players to deterministic strategies (e.g., by standard backward induction).

\sloppy An \emph{adaptive strategy} for \PlayerP in $\sprinstance(n,s)$ assigns to each history a distribution over the currently empty cells and the option to terminate. A strategy for \PlayerL observes that history and the current pointed cell, and specifies a distribution over a legally removable subset of signs together with the sign to place in the pointed cell.
In contrast, an \emph{oblivious strategy} for \PlayerP cannot use the past sign choices of \PlayerL. Formally, an oblivious strategy for \PlayerP is  specified by a joint distribution over tuples $(k_1, \ldots, k_s) \in [n]^s$.\footnote{For simplicity, we only consider oblivious strategies which do not terminate the game early.} We will only consider adaptive strategies for \PlayerL and therefore drop the term ``adaptive'' when describing \PlayerL's strategies.

\section{Overview of \Cref{thm:cal-ub}: sign-preservation upper bound}
\label{sec:spr-intuition}

A key technical contribution of this work is a strategy $\AlgA$ for Player-L in the SPR game with $n$ cells and $n$ time steps (i.e., $\sprinstance(n,n)$) which ensures that the number of pluses and minuses remaining at the end are each at most $O(n^{1-\eps})$, thus establishing \Cref{thm:cal-ub}.  Before diving into the technical details of the strategy (presented in full in \Cref{sec:signflipping-ub}), we provide here some intuition behind the algorithm and its proof.

As a first attempt, we try to construct an algorithmic strategy $\AlgA_{n}$ that takes the number of cells $n$ as a parameter and ensures the following property. For any $t \in \BN$ and sign $\sigma \in \{+1,-1 \}$, let the number of signs of type $\sigma$ remaining after $t$ time steps be denoted by $\signs(\AlgA_n, t, \sigma)$. We aim to show that there are some positive constants $\alpha, \beta$ satisfying $\alpha + \beta < 1$ so that for every power-of-two board size $n$ and every $t\in\BN$, it holds that
\begin{align}\label{eq:intuition-induction}
  \signs(\AlgA_n, t, \sigma) \leq n^\alpha  t^\beta.
\end{align}
Choosing $t=n$ gives the desired result for power-of-two board sizes. Padding to the next power of two extends the bound to arbitrary board sizes, as shown in the proof of \Cref{thm:cal-ub}. 

The key idea is to accomplish this task recursively.  We want $\AlgA_n$ to instantiate two copies of $\AlgA_{n/2}$: one for the \emph{left half} of cells (i.e., cells $1, 2, \ldots, n/2$) and one for the \emph{right half} of the cells (i.e., cells $n/2+1, \ldots n$).  When Player-P chooses a cell in the left half of $[n]$, $\AlgA_n$ places the sign chosen by the left instance of  $\AlgA_{n/2}$, and similarly for the right half.  Assuming that, over the course of $t$ rounds, Player-P chooses a cell in the left half $\tau$ times and in the right half $t-\tau$ times, we would have by induction that, for each sign $\sigma \in \{-1,+1\}$, 
\begin{equation}
  \label{eq:intuition-tau}
    \signs(\AlgA_n,t,\sigma) \leq (n/2)^{\alpha}\left( \tau^\beta + (t-\tau)^\beta \right).
  \end{equation}
  To show how to use \Cref{eq:intuition-tau}, we consider a few cases. 
\paragraph{Case 1: imbalanced choices} Let us suppose first that there is an imbalance in the number of times that Player-P chooses each half. In particular, if either $\tau \leq t/3$ or $\tau \geq 2t/3$, then it is straightforward to see using concavity of the function $\tau \mapsto \tau^\beta + (t-\tau)^\beta$ that for each $\sigma \in \{ +, -\}$, 
\begin{equation}\label{eq:unbal-example}
    \signs(\AlgA_n,t,\sigma) \leq (1/2)^{\alpha}\left( (1/3)^\beta + (2/3)^\beta \right)n^\alpha t^\beta.
\end{equation}
It is straightforward to choose $\alpha, \beta$ so that  $(1/2)^{\alpha}\left( (1/3)^\beta + (2/3)^\beta \right) \leq 1$ (e.g., $\alpha = \beta = 0.49$), which thus  completes the inductive step \Cref{eq:intuition-induction}. (This case is formalized in \Cref{lem:large-ratio}.)

\paragraph{Case 2: balanced choices} Now suppose that Player-P \emph{does} point roughly equally to both halves ($t/3 \leq \tau \leq 2t/3$). In the worst case, we have $\tau = t/2$, in which case the inductive hypothesis \Cref{eq:intuition-tau} gives
\begin{equation}\label{eq:almost-good}
    \signs(\AlgA_n,t,+1) \leq 2(1/2)^{\alpha}(1/2)^\beta n^\alpha t^\beta = 2^{1-\alpha-\beta} n^\alpha t^\beta,
\end{equation}
which only gives the desired bound \Cref{eq:intuition-induction} if $\alpha+\beta \geq 1$. Note that if we can improve upon \Cref{eq:almost-good} by \emph{any} constant factor (e.g., replace the right-hand side with $0.99 \cdot 2^{1-\alpha-\beta}n^\alpha t^\beta$), then the right-hand side of \Cref{eq:almost-good} would be upper-bounded by $n^\alpha t^\beta$ for some $\alpha + \beta < 1$. Roughly speaking, doing so is our goal. 

To achieve this goal, we must take advantage of the fact that signs get deleted!  We consider the following two sub-cases:

\paragraph{Case 2a: halves remain roughly balanced throughout} Suppose that Player-P chooses a cell in the left half at some time step $s_1$ when at least $t/6$ signs have been placed in the right half, and that Player-P chooses a cell in the right half at some time step $s_2$ when at least $t/6$ signs have been placed in the left half. At time step $s_1$, all $+$'s in the right half are removed, and at time step $s_2$, all $-$'s in the left half are removed. This means that for each sign $\sigma \in \{ +1, -1 \}$, for a constant fraction of the steps (i.e., at least $t/6$ steps), we only have to ``account for'' signs on either the right half (if $\sigma = +$) or the left half (if $\sigma = -$). This effectively allows us to improve \Cref{eq:almost-good} by a constant factor; the details are formalized in \Cref{lem:b-perp}. 

\paragraph{Case 2b: halves are very unbalanced at some point}
Suppose the previous case does not occur: in particular, suppose that there is no time step when Player-P chooses a cell in the left half when at least $t/6$ signs have already been placed in the right half. Let $s$ be the last time step during which a sign was placed in the left half. Since we have assumed $t/3 \leq \tau \leq 2t/3$, it follows that at time step $s$, at least $t/3$ cells have already been placed in the left half but fewer than $t/6$ cells have been placed in the right half. 
Notice that the placement of a sign in the right half at step $s+1$ causes all minus signs in the left half to be deleted. The main dilemma is as follows: since \emph{no cells} chosen by Player-P following step $s$ are in the left half, we cannot ensure that, e.g., a constant fraction of plus signs in the right half are deleted (as we did in the previous case). 

The astute reader will notice that up until this point, we have not used in any way the particular choice of signs played by Player-L. (Indeed, as we have not yet described the base case $n=1$ of the recursion, we have not yet fully specified the Player-L strategy.) To deal with the present case, we do need Player-L to choose which sign to place in a particular way: roughly speaking, our strategy is for Player-L to ``bias'' the signs following step $s$ ``towards'' being minus signs. This bias will allow us to improve the inductive bound \Cref{eq:intuition-tau} for the right half instance $\AlgA_{n/2}$ for $\sigma = +1$ by a constant factor. Moreover, after step $s+1$, all minus signs contributed by the at least $t/3$ earlier left-half placements are absent. Those rounds therefore contribute no surviving minus signs, which offsets the additional bias towards minus signs in the right half. Altogether, we will be able to improve upon \Cref{eq:almost-good} for each $\sigma \in \{+1,-1\}$ by a constant factor, as desired. This intuition is formalized in \Cref{lem:phase4}. 

\emph{How do we ensure this bias towards minus signs following step $s$?} 
The main idea is to \emph{generalize} the recursive algorithm $\AlgA_n$ to take as input an additional \emph{bias parameter} $\rho \in \BR$. One should interpret the bias parameter as a sort of ``hint'' which indicates to the algorithm $\AlgA_n$ (a) which sign it should tend to place (determined by the sign of the bias parameter), and (b) how strongly this effect propagates to recursive sub-algorithms (determined by the magnitude of the bias parameter, which will be at most $O(\log n)$).  Formally, for every power-of-two board size $n$ representing the number of cells covered by the algorithm instance and bias parameter $\rho$, we actually construct an algorithmic strategy $\AlgA_{n,\rho}$.  Instead of \Cref{eq:intuition-tau}, we will ensure that for number of rounds $t$ and sign $\sigma \in \{-1,+1\}$,
\begin{align}
  \label{eq:signs-bias}
  \signs(\AlgA_{n,\rho},t,\sigma) \leq n^\alpha t^\beta 2^{\rho \sigma}.
\end{align}
The algorithm $\AlgA_{n,\rho}$ operates in the same manner to $\AlgA_n$ as described above, with the following key difference: in Case 2b above, after the completion of step $s$ (namely, the last time step during which a sign was placed in the left half),\footnote{In a symmetric case, we also need to consider the time step $s'$ which is the last time step during which a sign was placed in the right half. We omit the details in this overview.} we \emph{re-initialize} the instance of $\AlgA$ corresponding to the right half with a \emph{smaller} bias parameter. In particular, if the right half was originally initialized as $\AlgA_{n/2, \rho}$, then we re-initialize it as $\AlgA_{n/2, \rho-\lambda}$ for some universal constant $\lambda > 0$. Moreover, for a ``leaf instance'' of $\AlgA$, namely one of the form $\AlgA_{1, \rho}$ (i.e., so that it controls a single cell), it will always return the sign $\mathrm{sign}(\rho)$, with ties broken arbitrarily if $\rho = 0$. In this manner, the re-initialization of $\AlgA_{n, \rho}$ with smaller bias $\rho-\lambda$ ultimately propagates down to the leaves, which will induce more signs placed following step $s$ to be minus signs.

Finally, we discuss one additional aspect of the algorithm which the above discussion overlooks: \emph{How does the procedure $\AlgA$ know what the step $s$ is, namely the last time step during which a sign was placed in the left half?} 
In general, due to the adversarial nature of Player-P, it is impossible to guess which cells Player-P will choose in the future, and thus the quantity $s$ \emph{cannot} be determined. Our procedure for Player-L overcomes this by using a sort of doubling trick which repeatedly ``guesses'' values for $s$ which increase geometrically. If a previous guess turns out to be incorrect, it re-initializes the $\AlgA_{n,\rho}$ instance with a new ``guess'' which is larger by a constant factor. We refer the reader to \Cref{sec:signflipping-ub} for details. 

\section{Equivalence between Calibration and Sign Preservation}
\label{sec:equivalence}
In this section, we prove \Cref{col:main_eq}, showing an equivalence between the optimal calibration error and the sign-preservation (with reuse) game. 
This result is a direct corollary of \Cref{thm:main_upper}, which shows that an $O(n^{1-\eps})$ upper bound on $\opt(n,n)$, for a fixed $\eps>0$, yields an $O(T^{2/3-\eps/18})$ upper bound for calibration, and of \Cref{thm:adapt_lower}, which shows that an $\Omega(n)$ lower bound on $\opt(n,n)$ translates into a $\tilde\Omega(T^{2/3})$ lower bound for calibration. The formal proof of \Cref{col:main_eq} is presented in \Cref{sec:equivalence_proof}. We present these two latter results in the below subsections.

\subsection{Upper Bounding $\ell_1$-calibration error using Sign Preservation}\label{sec:upper}
In this section, we show the existence of a forecaster that minimizes calibration error by following the strategy of \PlayerL in the sign preservation game. In particular, we show the following:


\begin{theorem}\label{thm:main_upper}
Suppose that $f:\BR_{>0}\to\BR_{>0}$ satisfies $f(\alpha)\le\min\{\alpha,1\}$ for every $\alpha>0$, and that there is a constant $C_0>0$ such that
\[
\opt(n,n^\alpha)\le C_0n^{f(\alpha)}
\]
for every integer $n\ge2$ and every $\alpha>0$. Define
\[
\gamma:=\sup_{\alpha>0}\frac{f(\alpha)}{\alpha+1}.
\]
The assumption on $f$ ensures that $0<\gamma\le1/2$. Then there is a forecaster (\Cref{alg:upper}, with a suitable choice of $h$) that guarantees expected calibration error
\[
O\!\left(T^{\frac{3-2\gamma}{5-4\gamma}}\log T\right)
\]
against any adaptive adversary. If $\gamma=1/2$, the forecaster guarantees $O(T^{2/3})$ expected calibration error.
\end{theorem}


We first overview the best-known upper bound of $O(T^{2/3})$ \cite{foster1998asymptotic,hart2023minimax} on calibration error, and give some intuition for how the SPR game comes into play. In particular, we follow the proof of \cite{hart2023minimax}, which uses the minimax theorem to show that we can assume at each time step $t$ that the forecaster observes $e_t \in [0,1]$, the  expected value of the binary outcome $y_t$ conditioned on the history $H_t = (p_1, y_1), \ldots, (p_{t-1}, y_{t-1})$.\footnote{As such, this proof is non-constructive in nature; so will be our forecaster establishing \Cref{thm:main_upper}.} The forecaster of \cite{hart2023minimax} rounds $e_t$ to the nearest multiple $p_t$ of $T^{-1/3}$ and predicts $p_t$. Overall, the calibration error may be bounded by the sum of (a) the ``biases'' from rounding each $e_t$ (which amounts to error at most $O(T^{-1/3}) \cdot T \leq O(T^{2/3})$), and (b) a ``variance'' term which results from variation in the predictions at each point in $\{0, T^{-1/3}, \ldots, 1\}$ (which, by a standard concavity argument, amounts to error of at most $T^{1/3} \cdot O(\sqrt{T^{2/3}}) = O(T^{2/3})$).

\vspace{-0.22cm}
\paragraph{``Covering up'' past errors} While the analysis above is tight \cite{qiao2021stronger}, a forecaster which observes $e_t$ as above can attempt to decrease the number of distinct predictions they make below $T^{1/3}$ (which will decrease the variance error), without increasing the bias error. In particular, if the forecaster has incurred positive calibration error on some point $p$ prior to time step $t$ (i.e., $E_{t-1}(p) > 0$), then if, at step $t$, $e_t > p$, it can predict $p_t = p$, which gives $E_t(p)-E_{t-1}(p)$ negative conditional expectation. In this way, the forecaster can ``cover up'' past calibration errors. Of course, the forecaster can only do this if the value of $e_t$ is greater than some $p$ with positive calibration error (or less than some $p$ with negative calibration error). Dividing up the interval $[0,1]$ into intervals, labeling each interval by the ``typical'' sign of calibration error for points $p$ in that interval, and letting the values $e_t$ correspond to the actions of \PlayerP, we arrive at a setting which closely resembles the SPR game. We will prove \Cref{thm:main_upper} by showing that a strategy for \PlayerL which certifies an upper bound on $\opt(n,s)$ in the SPR game allows us to construct a forecaster which can ``cover up'' the calibration error for many past predictions. Roughly speaking, this follows since signs that are \emph{removed} in the SPR game correspond to predictions where we successfully cover up past calibration error. 

We proceed to describe how a forecaster can simulate a Player-L strategy so that its bias error is bounded in terms of the signs remaining in the simulated SPR games, together with a constant allowance for each distinct cell and game used.  To do so, the forecaster must be able to deliberately bias their calibration error in a specific direction, akin to Player-L's ability to assign either a \( + \) or \( - \) sign in a given cell. 

\vspace{-0.22cm}
\paragraph{Biasing Calibration Error in a Specific Direction}
To explain the intuition behind our approach, we consider an epoch-based adversarial framework inspired by the lower bound construction of Qiao and Valiant. Here, the time steps \(1, 2, \ldots,  T \) are divided into \( m \) equally spaced epochs, for some $m \in \BN$. At the beginning of each epoch, the adversary announces that they will generate outcomes \( y \) by sampling from a Bernoulli distribution with an expectation \( e \), where \( e \in \{ 0, 1/k, 2/k, \ldots, 1 \}\), for the next \( \frac{T}{m} \) time steps. Assume that \( \frac{T}{mk} > \sqrt{\frac{T}{m}} \). 
Under these conditions, the forecaster can manipulate their calibration error by shifting predictions deliberately as follows: if the adversary announces a probability \( \frac{j}{k} \), the forecaster can predict either \( \frac{j+1}{k} \) or \( \frac{j-1}{k} \) consistently for the entire epoch duration. By the end of the epoch, this intentional bias contributes \( \frac{T}{m} \times \frac{1}{k} \) to the calibration error in a location ``near to'' $j/k$. Since \( \frac{T}{mk} > \sqrt{\frac{T}{m}} \), this bias term surpasses the variance term, which scales as \( \sqrt{\frac{T}{m}} \).

The forecaster also has the ability to eliminate calibration errors in accordance with the rules of the game. Specifically, in a new epoch, when the adversary announces a likelihood \( \frac{j}{k} \), the forecaster can first address errors from previous epochs by predicting the value with error that can be removed in expectation. This corresponds to the ``sign removal'' component of the game.

By combining these strategies, the forecaster can charge the bias associated with preserved signs to their number times the bias per epoch, $T/(mk)$. In the formal algorithm below, erased signs can leave bounded residual bias, so we also include a constant allowance for each distinct cell and game used.

\vspace{-0.22cm}
\paragraph{Adapting to the behavior of the adversary}
While the above approach seems effective, it contains a fundamental flaw. In the sign-preservation game with reuse, each epoch can place at most one sign in a given cell. This constraint implies that if, in a later epoch, the adversary announces a probability $j/k$ that has already been announced previously—without the corresponding error having been removed—this would not constitute a valid move for Player-P. Consequently, it becomes ambiguous how the forecaster should proceed. If they continue predicting according to the currently placed sign, the adversary can accrue additional error without increasing the total count of signs.

One potential solution is to start a new game and proceed as if playing with an empty cell. The forecaster could still predict the same probability values, but the new game structure provides a formal means of tracking the error magnitude, its location, and the potential for error reduction. However, initializing too many games is not desirable. This raises the question: \textit{how can we ensure the existence of at least one game with an empty cell, without needing to start numerous games? Moreover, we seek a strategy that does not rely on knowing $m$ in advance.}

\paragraph{Setup for \Cref{alg:upper}}
Accordingly, we propose a forecaster that runs $O(\log^2 T)$ different instances of the \emph{reuse sign-preservation game} as \PlayerL and decides what probability values to predict based on the strategy of \PlayerL. Let $h$ be an integer satisfying $1\le h\le\log_2 T-1$, to be specified below. As above, we assume the forecaster observes $e_t \in [0,1]$, the expected value of $y_t$ conditioned on the history $H_t = (p_1, y_1), \ldots, (p_{t-1}, y_{t-1})$. Although they know $e_t$, it may not be optimal to predict $e_t$. Instead, the forecaster chooses $p_t$ to be one of two prediction values associated with a selected cell of a simulated sign-preservation game (roughly, one near the left endpoint of the cell's interval and one near the right endpoint).  More precisely, for each \emph{discretization level}  $\frac{1}{2^{i+1}}$ for $i \in [\log T]$, the forecaster simulates $2 h$ instances of the game -- two instances (``even'' and ``odd'') for every possible value of $j \in [i+1,i+h]$, which controls the number of rounds. Letting $\tau := \log T$,
both instances for the values $(i,j)$ have $2^i$ cells. The labeler's guarantee is applied to the reduced transcript, with its round budget specified in \Cref{cor:bound-opt}. Each cell corresponds to a dyadic interval of length $2^{-(i+1)}$, with the endpoint convention specified below. The cells of the even instance correspond to even interval indices $m$, and the cells of the odd instance correspond to odd interval indices.

To summarize, for each $i \in [\log T]$, $j \in [i+1,i+h]$, and $\ell \in \{0,1\}$, we simulate an instance $G_{i,j,l}$ on $2^i$ cells, with the labeler's round budget specified in \Cref{cor:bound-opt}. 
For each $i$, partition $[0,1]$ into the intervals $[m/2^{i+1},(m+1)/2^{i+1})$, for $0 \leq m < 2^{i+1}$, including the right endpoint $1$ in the final interval. All interval expressions below use this convention.
Given an instance $G_{i,j,l}$ and a cell $c \in [2^i]$, we associate the cell with the interval whose index is $m=2(c-1)+l$. For a sign $s \in \{+,-\}$, define
\begin{itemize}[leftmargin=0.4cm]
\item $\interval(c, G_{i,j,l}) := \left[\frac{2(c-1)+l}{2^{i+1}}, \frac{2(c-1)+l+1}{2^{i+1}}\right)$, with the endpoint convention above.
\item $\probub(c,s,G_{i,j,l}) := \begin{cases}\frac{\max\{0,2(c-1)+l-1\}}{2^{i+1}} & \text{if }s=+ \\ \frac{\min\{2(c-1)+l+2,2^{i+1}\}}{2^{i+1}} & \text{if }s=-.\end{cases}$
\end{itemize}
Conversely, for $i \in [\log T]$, $j \in [i+1,i+h]$, and $e_t \in [0,1]$, define $\mathrm{cell}(i,j,e_t)$ as follows. Set
\[
m=\min\{\lfloor 2^{i+1}e_t\rfloor,2^{i+1}-1\},\qquad
l=m\bmod 2,\qquad c=(m-l)/2+1.
\]
Then $\mathrm{cell}(i,j,e_t)$ returns $(c,G_{i,j,l})$, whose associated interval contains $e_t$.

Since our forecaster simulates the \emph{reuse sign-preservation game (SPR)}, we define the $\simulateGame$ subroutine to be the following procedure: it takes in a cell $c$ of an instance $G_{i,j,l}$ and simulates one round of that SPR game. The state of the game $G_{i,j,l}$ is updated according to the rules of the game. That is, \PlayerP chooses cell $c$, all minus signs to the left of $c$ and plus signs to the right of $c$ are erased, and \PlayerL places a sign in cell $c$. 

For every cell $c$ of an instance $G$, \Cref{alg:upper} maintains two counters, $\biasub^+(c,G)$ and $\biasub^-(c,G)$, initialized to zero. These record the bias contributions at the two fixed predictions $\probub(c,+,G)$ and $\probub(c,-,G)$, respectively. More precisely, let $\mathcal T_t(c,s,G)$ be the set of rounds $u\leq t$ on which the algorithm selects the pair $(c,G)$ and uses prediction sign $s$, whether for bias placement or bias removal. Identifying the signs $+$ and $-$ with $+1$ and $-1$, define
\[
\biasub_t^s(c,G):=\sum_{u\in\mathcal T_t(c,s,G)}s(e_u-p_u),\qquad s\in\{+,-\}.
\]
Thus the plus counter accumulates $e_u-p_u$ and the minus counter accumulates $p_u-e_u$ on their respective rounds. A counter is updated only on rounds assigned to its own cell and instance, even if another cell predicts the same probability. We will show in \Cref{lem:sign_bias} that both counters remain nonnegative.
Throughout the algorithm and its analysis, let $\biasub_t(c,G)$ denote the signed maximum
\[
\biasub_t(c,G):=\begin{cases}
\biasub_t^+(c,G),&\text{if }\biasub_t^+(c,G)\geq\biasub_t^-(c,G),\\
-\biasub_t^-(c,G),&\text{otherwise}.
\end{cases}
\]
We omit the time subscript when referring to the current state.
At each round $t$ of the calibration game, \Cref{alg:upper} performs the following steps:  First, (\Cref{line:big-bias-for}), it attempts to find some instance $G$ whose cell $c$ corresponding to $e_t$ has the property that there is some cell $\bar c < c$ with highly negative bias or some cell $\bar c > c$ with highly positive bias. If so, then we predict the probability value corresponding to such cell $\bar c$ (per $\probub$), which decreases the qualifying bias component and cannot increase the magnitude of the signed maximum. If we cannot find any such $\bar c$, then (in \Cref{line:small-bias-for}) then we use the $\simulateGame$ procedure to place a sign in an appropriate instance (namely, one of relatively low bias), and use the value of the sign to slightly bias our prediction with respect to the value of $e_t$ (per $\probub$). The idea behind this step is that the $\simulateGame$ procedure is giving us a way of purposefully biasing our prediction to make it easier to ``cancel out'' in future rounds. 
\begin{algorithm}[h!]
\caption{Calibration Algorithm for \Cref{thm:main_upper}}
\label{alg:upper}
\begin{algorithmic}[1]
\Require A \PlayerL strategy for the game $\spr(n,s)$ for all values of $n,s$; integer parameter $1\le h\le\tau-1$.
\State Initialize $\biasub^+(c,G)=\biasub^-(c,G)=0$ for every instance $G$ and cell $c$ of that instance.
\For{each round $t \in [T]$} 
\State Observe $e_t \in [0,1]$
\For{$i \in [\log T]$}\label{line:big-bias-for} \Comment{perform ``bias removal'' if possible}
\For{$j \in [i+1,i + h]$}
\State Set $c, G \leftarrow \mathrm{cell}(i,j, e_t)$
\If{$\exists \ \bar{c} < c$ with $\biasub(\bar{c},G) < -1$ \textbf{or} $\exists \ \bar{c} > c$ with $\biasub(\bar{c},G) > 1$}\label{line:big-bias-if}
\State Set $s \leftarrow \mathrm{sign}(\biasub(\bar{c},G))$
\State Predict $p_t := \probub(\bar{c},s,G)$ \Comment{decrease the selected bias component}
\State $\biasub^s(\bar c,G)\leftarrow\biasub^s(\bar c,G)+s(e_t-p_t)$
\State Terminate round $t$
\EndIf
\EndFor
\EndFor

\For{$i \in [\log T]$}\label{line:small-bias-for} \Comment{perform ``bias placement'' if no ``bias'' to remove}
\For{$j \in [i+1,i + h]$}
\State Set $c, G \leftarrow \mathrm{cell}(i,j, e_t)$
\If{$|\biasub(c,G)| < 2^{j-i}$ }\label{line:small-bias-if}
\If{cell $c$ is empty}
\State $\simulateGame(c, G)$ \Comment{places sign in cell $c$, performs sign removal in SPR}
\EndIf
\State Set $s \leftarrow $sign in cell $c$
\State Predict $p_t := \probub(c,s,G)$ \Comment{predict to obtain negative or positive bias}
\State $\biasub^s(c,G)\leftarrow\biasub^s(c,G)+s(e_t-p_t)$
\State Terminate round $t$
\EndIf
\EndFor
\EndFor

\EndFor
    \end{algorithmic}
  \end{algorithm}

  \paragraph{Proof Overview for \Cref{thm:main_upper}}
Let $P$ be a fixed finite grid containing every possible prediction of \Cref{alg:upper}, and define
\[
B_p:=\sum_{t=1}^T\ind[p_t=p](e_t-p),
\qquad
M_p:=\sum_{t=1}^T\ind[p_t=p](e_t-y_t).
\]
By the triangle inequality,
\[
\E[\calerr(T)]
\leq \E\sum_{p\in P}|B_p|+\E\sum_{p\in P}|M_p|.
\]
We call these the bias and variance terms.
\begin{itemize}[leftmargin=0.4cm]
    \item To bound the variance term, the martingale-difference property gives $\E\sum_{p\in P}M_p^2\leq T$. By \Cref{lem:distinct_intervals}, the number of distinct prediction values used on every run is at most $K=O(2^{(\tau-h)/2})$. Cauchy--Schwarz on each run, followed by Jensen's inequality, therefore gives
    \[
    \E\sum_{p\in P}|M_p|\leq\sqrt{KT}
    =O(2^{3\tau/4-h/4}).
    \]
    The details appear in \Cref{lem:calerr_upper}.
    \item  
      In \Cref{lem:sign_bias}, we show that the bias error is at most $\sum_{c,G}|\biasub(c,G)|+N_T$, where $N_T$ is the number of distinct cell and instance pairs used by the algorithm. The proof keeps track of the two prediction values associated with each pair and shows that the smaller component is always at most one. The same lemma shows that $|\biasub(c,G)|\leq 1$ for an empty cell and $|\biasub(c,G_{i,j,l})|\leq 2^{j-i}+1$ otherwise. Thus, for each game $G=G_{i,j,l}$, its bias contribution is at most $2^{j-i}S_G+2N_G$, where $S_G$ is its number of preserved signs and $N_G$ is its number of distinct used cells. The preserved-sign bound uses \Cref{lem:useful,cor:bound-opt}, while \Cref{lem:distinct_intervals} bounds the used-cell contribution. In \Cref{lem:calerr_upper}, we combine these bounds and absorb the residual contribution into the term of order $2^{3\tau/4-h/4}$.
    \item We then use \Cref{lem:calerr_upper} to prove \Cref{thm:main_upper} by setting $h$ appropriately to balance the total expected bias error with the total expected variance error.
\end{itemize}

\subsection{Lower Bounding $\ell_1$-calibration error using Sign Preservation}\label{sec:lower}

In this section we show the existence of an (adaptive) adversary in the calibration game to lower bound the calibration error in terms of the performance of a given strategy of \PlayerP in the SPR game. In particular, we show the following:

\begin{theorem}\label{thm:adapt_lower}
  Suppose that there are constants $C_0 ,\alpha, \beta > 0$ so that $\opt(n, n^\alpha) \geq C_0 \cdot n^\beta$ for all $n \in \BN$.  
  Then, there exists an (adaptive) adversary  that uses \Cref{alg:adapt_lower} to force any forecaster to incur $\tilde\Omega (T^\frac{\beta + 1}{\alpha + 2})$ expected calibration error.
\end{theorem}

\paragraph{Setup for \Cref{alg:adapt_lower}}
We propose an adversary that runs a single instance of the Sign-Preservation Game with Reuse. The adversary uses a deterministic optimal strategy of \PlayerP for this instance, which guarantees at least $C_0n^\beta$ preserved signs against every legal sequence of labeler actions and certifies $\opt(n,n^\alpha) \geq C_0 \cdot n^\beta$. 
This adversary will simulate \PlayerL of an instance of SPR.  It does so by dividing the timesteps $t \in [T]$ into at most $n^\alpha$ \emph{epochs}, each corresponding to a single round of the SPR game. 

Fix any timestep $t_0 \in [T]$ which corresponds to the beginning of the epoch corresponding to some round $r$ of the SPR game. 
For a fixed cell $i \in [n]$, we define the following quantities: 
\begin{enumerate}[leftmargin=0.4cm]
\item $\mu^*_i = \frac{1}{3} + \frac{2i - 1}{6n}$.
\item $\Int_i = [l_i, r_i) := \left[\frac{1}{3} + \frac{i - 1}{3n}, \frac{1}{3} + \frac{i}{3n}\right) $. Note that $\mu^*_i$ is the midpoint of $\Int_i$.
\item Let $\Phi_{t}(i)$ denote the total ``negative calibration error'' in cells to the left of $i$ and ``positive calibration error'' in cells to the right of $i$:
\[ \Phi_t (i) = \sum_{p \in P \cap [0, {l}_i)}\calD^{-}_t(p) + \sum_{p \in P \cap [{r}_i,1]}\calD^{+}_t(p).\]
We will refer to the left summation above (i.e., over $p \in P \cap [0, l_i)$) as $\Phi_t^-(i)$ and the right one (i.e., over $p \in [r_i, 1]$) as $\Phi_t^+ (i)$.
\end{enumerate}
This adversary will simulate \PlayerL of an instance $G$ of $\text{SPR}(n,n^\alpha)$ as follows. If the given \PlayerP strategy selects cell $i$ on round $r$, the adversary draws the outcomes (in the calibration game) from a $\text{Ber} \left(\frac{1}{3} + \frac{2i - 1}{6n} \right)$ random variable until sign placement occurs.
\emph{Sign placement} is said to have occurred at a timestep $t$ of round $r$ starting at timestep $t_0$ if one of the following holds:
\begin{enumerate}[leftmargin=0.4cm]
    \item $\sum_{p\in P\cap\Int_i}|\calD_{t-1}(p)| \geq \theta$; \ i.e., there is $\geq \theta$ error in the interval corresponding to cell $i$ (\emph{Condition 1}).
    \item $\Phi_{t-1} (i) - \Phi_{t_0-1} (i) \geq \theta$; \ i.e., the potential $\Phi_{t-1}(i)$ has grown by a sufficient amount (\emph{Condition 2}).  
    \end{enumerate}
    The sign chosen in each case is specified in \Cref{alg:adapt_lower}.


\begin{algorithm}
\caption{Calibration adversary for \Cref{thm:adapt_lower}}\label{alg:adapt_lower}
\begin{algorithmic}[1]
\Require Value of $\alpha, \beta$
\State\label{line:set-ntheta} Set $n = (T/\ln^5 T)^{\frac{1}{\alpha + 2}}, \theta = \frac{1}{1440} \sqrt{\frac{T}{n^\alpha \ln T}}$ 
\State Initialize $\sprinstance (n, n^\alpha)$ with the fixed deterministic Pointer strategy
\State Set $t \gets 1$
\For{epoch $r \in [n^\alpha]$}
\State $t_0 \gets t$
\State If \PlayerP terminates the game in round $r$, then break
\State Let $i \in [n]$ be the cell chosen by the Pointer in round $r$
\State Remove all minus signs left of $i$ and all plus signs right of $i$
\While{sign placement condition is not satisfied}
  \State If $t>T$, stop
  \State Draw $y_t$ from $\text{Ber} \left( \mu_i^* \right)$\label{line:draw-outcome}
  \State Observe forecaster's prediction $p_t$
  \State $t \leftarrow t + 1$
\EndWhile
\If{sign placement Condition 1 is satisfied}
\State Set \PlayerL's sign to $+$ if $  \sum_{p \in P \cap \Int_i}\calD^{-}_{t-1} (p) \ge \sum_{p \in P \cap \Int_i}\calD^{+}_{t-1} (p)$ and $-$ otherwise\label{line:playerl-1}
\ElsIf{sign placement Condition 2 is satisfied}
\State Set \PlayerL's sign to $+$ if $ \Phi_{t-1}^- (i) - \Phi_{t_0-1}^- (i) \geq \Phi_{t-1}^+ (i) - \Phi_{t_0-1}^+ (i) $ and $-$ otherwise\label{line:playerl-2}
\EndIf
\EndFor
\State Set $\tau\gets t-1$
\State If $\sum_{p\in P}\calD_\tau^+(p)\ge\sum_{p\in P}\calD_\tau^-(p)$, output $0$ for all remaining rounds; otherwise output $1$
\end{algorithmic}
\end{algorithm}

To prove \Cref{thm:adapt_lower}, we show that the SPR simulation finishes at a time $\tau<T$ with probability $1-o(1)$, and that $m$ preserved signs then imply $\calerr(\tau)\ge m\theta/4$. Since the Pointer strategy guarantees $m\ge C_0n^\beta$ on every completed play, the final continuation gives $\calerr(T)\ge\calerr(\tau)/2$ and hence the stated expected lower bound. The preservation argument uses the fact that all legally removable signs are erased at the start of each epoch: while a minus in cell $i$ survives, subsequent pointers lie strictly to its left, and while a plus survives, they lie strictly to its right. Sign placement Condition 2 allows error accumulated outside $\Int_i$ to contribute to this argument. Full details appear in \Cref{sec:adapt_lower_proofs}.


\section{An upper bound for the sign-preservation game}
\label{sec:signflipping-ub}
In this section we prove \Cref{thm:cal-ub}, giving a strategy for \PlayerL in the SPR game which certifies $\opt(n,n) \leq O(n^{1-\eps})$ for some $\eps > 0$ and thereby yields a forecasting algorithm with calibration error of $O(T^{2/3 - \eps/18})$, by \Cref{col:main_eq}. 
\subsection{Analysis overview}
We first describe and analyze the recursive strategy for board sizes that are powers of two. The proof of \Cref{thm:cal-ub} below extends the bound to arbitrary board sizes by padding.
The strategy for \PlayerL in the sign preservation game is specified by two mutually recursive algorithms, \AlgA and \AlgB, defined in \Cref{alg:A,alg:B}, respectively. These algorithms constitute a tree-based strategy for \PlayerL, where the $n$ game cells serve as the leaves of the tree. Each interior node executes a sub-algorithm of \AlgA and \AlgB determined by the algorithm instances at its parent. 
For a power-of-two board of size $n$, the root instance of \AlgA, denoted by $\AlgA_0$, is initialized with $l=1,r=n,b=0$. For an arbitrary board of size $n$, we instead simulate the root on a virtual board of size $N=2^{\lceil\log_2 n\rceil}$, initialized with $l=1,r=N,b=0$, and use only its first $n$ cells.

\begin{algorithm}[!h]
  \caption{Definition of \AlgA}
  \label{alg:A}
  \begin{algorithmic}[1]
    \Require Parameters $l,r \in \mathbb{N}$ with $l\leq r$ and $r-l+1$ a power of two, $b \in \mathbb{Z}$. \emph{($[l,r]$ represents the interval of the game board corresponding to $\AlgA$, and $b$ represents a bias parameter controlling whether returned signs are biased towards $+1$ or $-1$.)} 

    \Function{\AlgA.\initialize}{$l,r,b$}
    \If{$l<r$}
    \State $\recentB \gets \AlgB.\initialize(l,r,b,1)$.
    \EndIf
    \State $\cnt \gets 0$.
    \EndFunction

    \Function{\AlgA.\placeSign}{$s$}\Comment{$s \in [l,r]$}
    \If{$l=r$}
    \State \Return $+1$ if $b\leq0$, and $-1$ otherwise.
    \Else
    \State $\cnt \gets \cnt + 1$.
    \State $\sigma \gets \recentB.\placeSign(s)$.
    \If{$\sigma = \perp$}
    \State $\recentB \gets \AlgB.\initialize(l,r,b,\cnt)$\label{line:b-reinit}
    \State $\cnt \gets 1$.
    \State \Return $\recentB.\placeSign(s)$.
    \Else
    \State\Return $\sigma$.
    \EndIf
    \EndIf
    \EndFunction
  \end{algorithmic}
\end{algorithm}

\begin{algorithm}[!h]
  \caption{Definition of \AlgB}
  \label{alg:B}
  \begin{algorithmic}[1]
    \Require Parameters $l,r \in \mathbb{N}$ with $l<r$ and $r-l+1$ a power of two, $b \in \mathbb{Z}, M \in \BN$. \emph{($[l,r]$ represents the sub-interval of the game board corresponding to this $\AlgB$ instance, $b$ represents the bias parameter, and $M$ represents a ``guess'' as to the number of steps $\AlgB$ is played for, which is used to implement a doubling trick.)}
    \Function{\AlgB.\initialize}{$l,r,b,M$}
    \State $m \gets \lfloor (l+r)/2 \rfloor$.
    \State $\AlgA[0] \gets \AlgA.\initialize(l,m,b)$,  $\AlgA[1] \gets \AlgA.\initialize(m+1,r,b)$. 
    \State $\prevHalf \gets -1$.
    \State $\cntHalf[0] \gets 0, \cntHalf[1] \gets 0$.
    \State $\phase \gets 1$. 
    \EndFunction

    \Function{\AlgB.\placeSign}{$s$}
    \State $\half \gets 0$ if $s \leq m$ else $\half \gets 1$.
    \State $\cntHalf[\half] \gets \cntHalf[\half] + 1$.
    \If{$\phase = 1$}
    \If{$\cntHalf[\half] = M$ and $M \leq \cntHalf[1-\half] \leq 2M$}
    \State $\phase \gets 2$
    \ElsIf{$\cntHalf[\half] = M$ and $\cntHalf[1-\half] > 2M$}
    \State $\phase \gets 3$
    \EndIf
    \ElsIf{$\phase = 2$}
     \If{$\half \neq \prevHalf$} \State\Return $\perp$.
    \EndIf
    \If{$\cntHalf[\half] = 2 \cdot \cntHalf[1-\half] + 1$}\State $\phase \gets 3$.
    \EndIf
    \ElsIf{$\phase=3$}
    \If{$\cntHalf[\half] = \lfloor \cntHalf[1-\half] / 2 \rfloor + 1$}
    \State $\phase \gets 4$.
    \If{$\half = 0$}
    \State \label{line:alg0-init} $\AlgA[0] \gets \AlgA.\initialize(l,m,b-1)$.
    \Else \Comment{$\half = 1$}
    \State \label{line:alg1-init} $\AlgA[1] \gets \AlgA.\initialize(m+1,r,b+1)$.
    \EndIf
    \EndIf
    \ElsIf{$\phase=4$}
    \If{$\cntHalf[\half] > \cntHalf[1-\half]$}\State\Return $\perp$.
    \EndIf
    \EndIf
    \State $\sigma \gets \AlgA[\half].\placeSign(s)$.
    \State $\prevHalf \gets \half$.
    \State \Return $\sigma$. 
    \EndFunction
  \end{algorithmic}
\end{algorithm}

\paragraph{Overall structure of \AlgA, \AlgB}
\sloppy Each of \AlgA, \AlgB has an initialization routine (i.e., $\AlgA.\initialize(l,r,b)$, $\AlgB.\initialize(l,r,b,M)$), and a labeling routine (i.e., $\AlgA.\placeSign(s), \AlgB.\placeSign(s)$). 
We say that an instance of \AlgA which is initialized via $\AlgA.\initialize(l,r,b)$ \emph{covers} a total of $n := r-l+1$ cells; namely, it covers the cells in $[l,r]$. We say that \AlgA is \emph{executed} for $t$ time steps if $\AlgA.\placeSign(s)$ returns a sign $\sigma \in \{-1,1\}$ (i.e., not $\perp$) for a total of $t$ time steps through the entire execution of the game, and write $\executionSteps(\AlgA) := t$. Finally, we call the parameter $b$ the \emph{bias} of \AlgA; positive $b$ biases returned signs towards $-1$, and negative $b$ towards $+1$. 
Given an instance \AlgA and $\sigma \in \{-1,1\}$, we let $\remainingSigns(\AlgA, \sigma)$ denote the number of signs of type $\sigma$ which are: (a) placed in one of the covered cells of \AlgA during the time steps in which it is executed, and (b) remain (according to the rules of the $\spr$ game) upon completion of \AlgA.\footnote{Notice that our notation pertaining to $\remainingSigns$ is a bit different in this section than in the overview in \Cref{sec:spr-intuition}. In particular, our bias parameter $b$ uses the opposite sign convention from the overview's parameter $\rho$, and we do not explicitly write the dependence of $t$ in $\remainingSigns$ (as we view this as being included in the specification of $\AlgA$, via $t = \executionSteps(\AlgA)$.) }

We use the same terminology for \AlgB: if an instance \AlgB is initialized via $\AlgB.\initialize(l,r,b,M)$, then we say it \emph{covers} the $n := r-l+1$ cells in the interval $[l,r]$, has \emph{bias} $b$, and is \emph{executed} for $t$ time steps if $\AlgB.\placeSign(s)$ returns a sign $\sigma \in \{-1,1\}$ (i.e., not $\perp$) for a total of $t$ time steps, and write $\executionSteps(\AlgB) = t$. The parameter $M \in \BN$ should be interpreted as a sort of ``guess'' which scales with the number of steps that $\AlgB.\placeSign(\cdot)$ will be called: it is used to implement a sort of doubling trick (as outlined in \Cref{sec:spr-intuition}). In particular, if $\AlgB.\placeSign(\cdot)$ is called many more than $M$ times, typically it will return $\perp$ at some point, which will cause $\AlgB.\initialize(\cdot)$ to be called in Line \ref{line:b-reinit} of \Cref{alg:A}. Upon re-initialization, the new value of $\cnt$ (i.e., the parameter $M$) will be at least a constant factor larger than the old value. 

Finally, we let $\remainingSigns(\AlgB, \sigma)$ denote the number of signs of type $\sigma$ which are placed in one of \AlgB's covered cells during the time steps in which it is executed, and remain upon completion of \AlgB.

A leaf instance of $\AlgA$ (i.e., $l=r$) has no descendants. Each nonleaf instance of $\AlgA$ maintains one or more instances of $\AlgB$ throughout its execution (namely, all values of $\AlgA.\recentB$ throughout its execution), each of which contains multiple instances of $\AlgA$ throughout their execution (namely, the initial values of $\AlgB.\AlgA[0], \AlgB.\AlgA[1]$, as well as the re-initialized value of $\AlgA$ in Line \ref{line:alg0-init} or Line \ref{line:alg1-init} of \Cref{alg:B}), and so on. We refer to all such instances of $\AlgB$ (respectively, $\AlgA$) as the \emph{$\AlgB$-descendents} (respectively, \emph{$\AlgA$-descendents}) of the parent $\AlgA$ instance. We refer to descendents of a parent $\AlgB$ instance in a similar manner. 

\paragraph{Overview of the labeling routines} Suppose the $\spr$ game is played for $t \in \BN$ timesteps; at each step where \PlayerL is called upon to place a sign in some cell $s \in [n]$, we place the sign returned by the root instance, namely $\AlgA_0.\placeSign(s)$. If $n=1$, the root is a leaf and returns $+1$ directly. Otherwise, this function returns the sign returned by $\recentB.\placeSign(s)$, where $\recentB$ is an instance of \AlgB maintained by $\AlgA_0$ (and re-initialized if necessary). In turn, each instance of \AlgB initialized via $\AlgB.\initialize(l,r,b,M)$ (\Cref{alg:B}) maintains two instances of \AlgA, denoted by $\AlgA[0], \AlgA[1]$, which cover the cells in the left and right halves of $[l,r]$, respectively. $\AlgB.\placeSign(s)$ returns the sign returned by $\AlgA[\half].\placeSign(s)$, where $\half$ is $0$ if $s$ is in the left half of $[l,r]$, and $1$ otherwise. In this manner, we have the following recursive structure: in the course of the outermost call $\AlgA_0.\placeSign(s)$, we call the $\placeSign(s)$ procedure for all active \AlgA, \AlgB instances on the root-to-leaf path starting at the leaf node determined by $s$. The recursion terminates at a leaf instance of \AlgA, which returns its sign directly. 

\paragraph{Summary of \AlgA} \Cref{alg:A} defines \AlgA. On input $(l,r,b)$, $\AlgA.\initialize$ sets $\cnt=0$ and, only when $l<r$, initializes a \AlgB instance with parameters $(l,r,b,1)$. A leaf instance ($l=r$) creates no \AlgB instance. When \AlgB returns $\perp$ on a $\AlgB.\placeSign$ call (which should be interpreted as \AlgB terminating), then \AlgA initializes a new \AlgB instance with parameters $(l, r, b, c)$ where $c$ is the current value of the counter.
 The procedure $\AlgA.\placeSign(s)$ is very simple: if \AlgA is a leaf instance (i.e., $l=r$), then it returns the opposite sign of its nonzero bias $b$, with $+1$ returned when $b=0$. Otherwise, it returns $\recentB.\placeSign(s)$ (after re-initializing $\recentB$, if applicable). 

\paragraph{Summary of \AlgB} \Cref{alg:B} defines \AlgB. On input $(l,r,b,M)$ with $l<r$, $\AlgB.\initialize$ initializes two \AlgA instances, denoted $\AlgA[0], \AlgA[1]$, for each half of the interval $[l,r]$ and maintains a counter for the number of times each of $\AlgA[0], \AlgA[1]$ is executed (i.e., the number of time steps \PlayerP chooses a cell in the left half or right half). Over the course of multiple calls to $\AlgB.\placeSign(\cdot)$, \AlgB passes through the following four phases: it remains in Phase 1 until both counters are at least $M$, and then it transitions to Phase 2 if the larger counter is not more than $2M$ and to Phase 3 otherwise. In Phase 2, it terminates (i.e., returns $\perp$) if \PlayerP switches halves (i.e., starts playing in a half different from the one that ended Phase 1), and otherwise it transitions to Phase 3 once the counter for its current half is just more than two times that of the other. In Phase 3, it transitions to Phase 4 when the counter for the current half becomes approximately half that of the other half; upon entering Phase 4, it re-initializes the \AlgA instance corresponding to the current half with an adjusted bias. In Phase 4, it continues placing signs as long as the counter for the current half does not exceed that of the other half (and returns $\perp$ once it would).   

We will prove the following lemma by induction:
\begin{lemma}
  \label{lem:a-induction}
There are constants $\alpha,\beta,\lambda,C>0$ satisfying $\alpha+\beta<1$ so that the following holds. Suppose an instance of \AlgA has bias $b\in\mathbb{Z}$, covers $n=2^d$ cells for a nonnegative integer $d$, and $t:=\executionSteps(\AlgA)$. Then for any $\sigma\in\{-1,1\}$,
\[
\remainingSigns(\AlgA,\sigma)
\leq C\lambda^{-b\sigma}n^\alpha t^\beta.
\]
\end{lemma}

In the analysis, we address local variables of instances \AlgA, \AlgB using the ``.'' notation, as in: $\AlgB.\cntHalf, \AlgB.\phase$, etc. Before proving \Cref{lem:a-induction}, we first observe that \Cref{thm:cal-ub} is an immediate consequence of it:
\begin{proof}[Proof of \Cref{thm:cal-ub}]
Fix a positive integer $n$, and set $N:=2^{\lceil\log_2 n\rceil}$, so that $n\leq N<2n$. Identify the original board with the first $n$ cells of a virtual board of size $N$, and initialize its root instance $\AlgA_0$ with $l=1,r=N,b=0$.

Whenever \PlayerP selects a cell $s\in[n]$, make the same selection on the virtual board. In both games, \PlayerL removes all minus signs to the left of $s$ and all plus signs to the right of $s$, and places the sign returned by $\AlgA_0.\placeSign(s)$. The additional cells are never selected.

Inductively, after every round the two boards agree on $[n]$, and all additional cells remain empty. Hence a cell selected legally on the original board is also empty on the virtual board. The embedding preserves left/right order, so the removals are legal and identical on the corresponding cells. Consequently, the preserved signs correspond exactly, including if \PlayerP terminates the game early.

Let $t\leq n$ be the number of rounds played. Since $N$ is a power of two, \Cref{lem:a-induction} gives, for each $\sigma\in\{-1,1\}$,
\[
\remainingSigns(\AlgA_0,\sigma)
\leq C N^\alpha t^\beta
\leq C N^\alpha n^\beta.
\]
Summing over the two signs yields
\[
\opt(n,n)
\leq 2C N^\alpha n^\beta
\leq 2^{1+\alpha}C n^{\alpha+\beta}
=2^{1+\alpha}C n^{1-\eps},
\]
where $\eps:=1-\alpha-\beta>0$. The calibration bound follows from \Cref{col:main_eq}.
\end{proof}

\paragraph{Ideas in the proof of \Cref{lem:a-induction}}
Before proceeding with a formal proof, we give some intuition behind the proof of \Cref{lem:a-induction}. We prove the lemma by induction on $d$, where $n=2^d$ is the number of cells covered by \AlgA, assuming the conclusion for instances covering fewer than $n$ cells whose sizes are powers of two. For this discussion, assume $n\geq2$; the case $n=1$ is handled directly by the leaf rule. Let $\AlgB_1,\ldots,\AlgB_k$ be the instances initialized during the execution of \AlgA, and write $t_i:=\executionSteps(\AlgB_i)$, so that $t=\executionSteps(\AlgA)=\sum_{i=1}^k t_i$. The completed instances $\AlgB_1,\ldots,\AlgB_{k-1}$ have geometrically increasing execution lengths; the final instance may stop earlier when the game ends. For intuition, first consider $k=1$, so that $t=t_k$. The full proof also accounts for the signs remaining from the completed instances.

We first consider the following ``naive'' argument, which does not suffice: suppose that $\AlgB_k$ never enters phase 4, meaning that $\AlgB_k.\AlgA[0], \AlgB_k.\AlgA[1]$ are never re-initialized. Letting $t_{k,h} := \executionSteps(\AlgB_k.\AlgA[h])$, so that $t_{k,0} + t_{k,1} = t_k = t$, then the inductive hypothesis yields that
\begin{align}
  \remainingSigns(\AlgA, \sigma) =& \remainingSigns(\AlgB_k, \sigma) \nonumber\\
  =& \remainingSigns(\AlgB_k.\AlgA[0],\sigma) + \remainingSigns(\AlgB_k.\AlgA[1])\nonumber\\
  \leq & C \lambda^{-b\sigma} \cdot (n/2)^\alpha \cdot (t_{k,0}^\beta + t_{k,1}^\beta) \leq C \lambda^{-b\sigma} n^\alpha t^\beta \cdot 2^{1-\alpha-\beta}\label{eq:naive-bound}.
\end{align}
In order for the above argument to establish the inductive step, we need $2^{1-\alpha-\beta} \leq 1$, i.e., $\alpha + \beta \geq 1$. Of course this is not useful, since in order to show $\opt(n,n) \leq O(n^{1-\eps})$ in the proof of \Cref{thm:cal-ub}, we crucially used that $\alpha + \beta < 1$.\footnote{It turns out that, due to a corner case, we need the preceding argument anyways in the proof of \Cref{lem:a-induction}; it is formalized in \Cref{lem:small-M} below.}

Thus, to prove \Cref{lem:a-induction}, we need to find a way to slightly decrease the term $2^{1-\alpha-\beta}$ in \Cref{eq:naive-bound} (i.e., by any constant factor less than 1). Roughly speaking, such a decrease occurs for one of the following reasons, depending on the state of $\AlgB_k$ after the last round on which $\AlgB_k.\placeSign(\cdot)$ is called:
\begin{itemize}
\item If $\AlgB_k$ never enters phase 4 and $t_{k,0}/t_{k,1} \not \in (1/2, 2)$, then the final inequality in \Cref{eq:naive-bound} can be improved to
  \begin{align}
C\lambda^{-b\sigma} (n/2)^\alpha (t_{k,0}^\beta + t_{k,1}^\beta) \leq C\lambda^{-b\sigma} n^\alpha t^\beta \cdot \frac{(1/3)^\beta + (2/3)^\beta}{2^\alpha}\nonumber,
  \end{align}
  and we now note that we can have $\frac{(1/3)^\beta + (2/3)^\beta}{2^\alpha} < 1$ for some $\alpha + \beta < 1$. (See \Cref{lem:large-ratio}.)  
\item Suppose $\AlgB_k$ enters phase 4; without loss of generality we may assume that it re-initializes $\AlgA[1]$ (in Line \ref{line:alg1-init}). The placement of the sign in $[m+1,r]$ upon entering phase 4 deletes all $-1$s placed in $[l,m]$. Using this fact, it is straightforward to improve the upper bound in \Cref{eq:naive-bound} for $\sigma = -1$ (\Cref{lem:phase4}).

  As for $\sigma = 1$, here we use the fact that $\AlgB_k.\AlgA[1]$ is re-initialized with bias $b+1$. Thus, the re-initialized instance of $\AlgB_k.\AlgA[1]$ can be shown to have a smaller number of remaining $+1$s after its execution, by a factor of roughly $1/\lambda$ (\Cref{lem:phase4}). We remark that this re-initialization with larger bias slightly increases the number of $-1$s remaining, but this does not cause an issue for the previous case ($\sigma = -1$) since there will be enough deleted $-1$s upon entering phase 4 to make up for this increase. 
\item If neither of the above cases occurs, the half-counts are balanced and $\AlgB_k$ is outside phase 4, so $t_k\leq6M$ by \Cref{lem:small-M-real}. If $k=1$, then $M=1$ and $t=t_k\leq6$, which is handled by the short-execution argument at the start of the proof of \Cref{lem:a-induction}. If $k\geq2$, the restart rule gives $M=t_{k-1}+1\leq3t_{k-1}/2$, because the completed instance $\AlgB_{k-1}$ has $t_{k-1}\geq2$. Consequently $t_k\leq9(t-t_k)$, or $t_k/t\leq9/10$. In this case, we use the stronger estimate for the completed instances: when an instance returns $\perp$, the placement on that round deletes one type of sign from one of the halves. These deletions compensate for the weaker bound needed for the balanced final instance.
\end{itemize}
To summarize, in all cases we are able to beat the naive bound in \Cref{eq:naive-bound} by a constant factor, which allows us to choose some $\alpha,\beta$ satisfying $\alpha + \beta < 1$. 

\subsection{Lemmas for inductive proof}
We establish \Cref{lem:a-induction} using induction on $d$, where $n=2^d$ is the number of cells covered by the instance of $\AlgA$. We choose
\begin{align}
\lambda = 1.5, \qquad C = 6 \cdot \lambda^3\label{eq:lambda-C},
\end{align}
and will choose the parameters $\alpha, \beta$ below in the proof of \Cref{lem:a-induction} (see \Cref{sec:a-induction-proof}). In this section, we establish several lemmas used to prove the inductive step of \Cref{lem:a-induction}. Thus, throughout this section, 
we fix $n=2^d$ for an integer $d\geq1$ and assume that \Cref{lem:a-induction} holds for every instance of \AlgA covering $2^j$ cells with $0\leq j<d$ (we refer to this assumption as the ``inductive hypothesis'').

Moreover, throughout this section, we fix an instance \AlgB which is initialized via $\AlgB.\initialize(l,r,b,M)$. We let $t := \executionSteps(B)$ denote the number of time steps for which \AlgB is executed, and $n = r-l+1$ denote the number of cells covered by \AlgB.
Since $m=\lfloor(l+r)/2\rfloor=l+n/2-1$, each child instance of \AlgA, including any re-initialized child, covers exactly $n/2=2^{d-1}$ cells. Thus the inductive hypothesis applies to every such child.

\begin{lemma}
  \label{lem:large-ratio}
  Suppose that, following the last round $\AlgB.\placeSign$ returns a sign, it holds that $\frac{\AlgB.\cntHalf[0]}{\AlgB.\cntHalf[1]} \not \in (1/2, 2)$ and $\AlgB.\phase \in \{1,2,3\}$. Then for each $\sigma \in \{-1,1\}$,
  \begin{align}
    \remainingSigns(\AlgB, \sigma) \leq C \lambda^{- b \cdot \sigma} n^\alpha t^\beta \cdot \frac{(1/3)^\beta + (2/3)^\beta}{2^\alpha}.\label{eq:large-ratio}
    \end{align}
\end{lemma}
\begin{proof}
  For $h \in \{0,1\}$, let $t_h$ denote the value of $\AlgB.\cntHalf[h]$ following the last round that $\AlgB.\placeSign$ returns a sign, so that $t_0 + t_1 = t$. By assumption we have $t_0/t_1 \not \in (1/2,2)$, meaning that $\max\{t_0, t_1\} \geq 2t/3$. Using the inductive hypothesis of \Cref{lem:a-induction} applied to $\AlgB.\AlgA[0], \AlgB.\AlgA[1]$ (which never change, as $\AlgB$ ends in phase 1, 2, or 3), we may compute
  \begin{align}
    \remainingSigns(\AlgB, \sigma) \leq & \remainingSigns(\AlgB.\AlgA[0],\sigma) + \remainingSigns(\AlgB.\AlgA[1], \sigma)\nonumber\\
    \leq & C \lambda^{-b\sigma} (n/2)^\alpha \cdot (t_0^\beta + t_1^\beta)\nonumber\\
    \leq & C\lambda^{-b\sigma} (n/2)^\alpha t^\beta \cdot ((1/3)^\beta + (2/3)^\beta)\nonumber,
  \end{align}
  where the final inequality holds by \Cref{lem:p-alpha-beta} with $p=1/3$. 
\end{proof}

\begin{lemma}
  \label{lem:phase4}
  Suppose that, following the last round in which $\AlgB.\placeSign$ returns a sign, it holds that $\AlgB.\phase = 4$. Then for each $\sigma \in \{-1,1\}$, and $\delta \in (0,1/3)$,
  \begin{align}
\remainingSigns(\AlgB, \sigma) \leq C \cdot \lambda^{-b \cdot \sigma} (n/2)^\alpha t^\beta \cdot \max \left\{ \frac{1+4\lambda/3}{4^\beta}, F_{\beta,\lambda}(\delta)\right\}\label{eq:phase4},
  \end{align}
  where $F_{\beta,\lambda}(\delta)$ is defined in \Cref{eq:f-delta-beta}. 
  If moreover \AlgB returns $\perp$, then in fact we have the following bound for each $\sigma \in \{-1,1\}$: 
  \begin{align}
&\remainingSigns(\AlgB, \sigma) \\
& \ \leq C \lambda^{-b\sigma} (n/2)^\alpha t^\beta \cdot \max \left\{ \frac{1+4\lambda/3}{4^\beta}, (1/2)^\beta + (1/4)^\beta + (1/4)^\beta/\lambda, (3/4)^\beta \right\}\label{eq:phase4-perp}. 
  \end{align}
\end{lemma}
We note that for $\beta$ sufficiently close to 1, the bound in \Cref{eq:phase4-perp} is stronger than that in \Cref{eq:phase4}. 
\begin{proof}[Proof of \Cref{lem:phase4}]
  Without loss of generality, let us suppose that $\half = 1$ during the round $\tau$ when $\AlgB$ sets $\phase\gets 4$, and $t_0$ denote the value of $\AlgB.\cntHalf[0]$ during round $\tau$. Write $t_1 := \lfloor t_0/2 \rfloor$; $t_1$ is the value of $\AlgB.\cntHalf[1]$ following round $\tau-1$, since by definition we must have $\AlgB.\cntHalf[1] = \lfloor \AlgB.\cntHalf[0]/2 \rfloor + 1$ when $\AlgB$ sets $\phase \gets 4$. Note that, if ever $\half = 0$ for some round $\tau' > \tau$, $\AlgB.\placeSign$ will return $\perp$. Thus, on all rounds following $\tau$ when $\AlgB.\placeSign$ returns a sign, we must have $\half = 1$. Moreover, the number of such rounds is given by $t_2 := t-t_1 - t_0$ and must satisfy $t_2 \leq \lceil t_0/2\rceil $ (as after $\lceil t_0/2\rceil $ rounds $\AlgB.\placeSign$ will return $\perp$).

  Let $\AlgB.\AlgA[0], \AlgB.\AlgA[1]$ denote the initial values of the sub-algorithms $\AlgA[0], \AlgA[1]$ (i.e., defined in $\AlgB.\initialize$), and let $\AlgB.\AlgA'[1]$ be the sub-algorithm initialized in Line \ref{line:alg1-init} when $\AlgB$ enters phase 4.

  \paragraph{Bounding the number of minuses} Note that all $-1$'s placed by $\AlgA[0]$ (which belong to cells in $[l,m]$) are deleted by the placement of the sign in round $\tau$ (which belongs to a cell in $[m+1, r]$). Thus, the inductive hypothesis yields that
  \begin{align}
    \remainingSigns(\AlgB, -1) \leq &  \remainingSigns(\AlgB.\AlgA[1], -1) + \remainingSigns(\AlgB.\AlgA'[1], -1) \nonumber\\
    \leq &   C \lambda^b (n/2)^\alpha t_1^\beta + C \lambda^{b+1} (n/2)^\alpha t_2^\beta \nonumber\\
    =& C \lambda^b (n/2)^\alpha \cdot (t_1^\beta + \lambda t_2^\beta)\nonumber\\
    \leq & C \lambda^b (n/2)^\alpha t^\beta \cdot \frac{1+4\lambda/3}{4^\beta} \label{eq:b-plus-4},
  \end{align}
where the final inequality uses \Cref{lem:plus-bound}. 

\paragraph{Bounding the number of pluses} In a similar manner, we may bound the number of $+1$'s remaining from $\AlgB$ as follows, using the inductive hypothesis:
\begin{align}
  \remainingSigns(\AlgB, +1) \leq & \ \remainingSigns(\AlgB.\AlgA[0], +1) + \remainingSigns(\AlgB.\AlgA[1], +1) \\
  &+ \remainingSigns(\AlgB.\AlgA'[1], +1) \nonumber\\
  \leq & C \lambda^{-b} (n/2)^\alpha t_0^\beta + C \lambda^{-b} (n/2)^\alpha t_1^\beta + C \lambda^{-b-1} (n/2)^\alpha t_2^\beta \nonumber\\
  = & C \lambda^{-b} (n/2)^\alpha \cdot (t_0^\beta + t_1^\beta + t_2^\beta/\lambda)\label{eq:pre-b-minus-4}\\
  \leq & C \lambda^{-b} (n/2)^\alpha t^\beta \cdot F_{\beta,\lambda}(\delta)\label{eq:b-minus-4},
\end{align}
where the final inequality uses \Cref{lem:minus-bound}. \emph{Note that this is the key point in the proof where we are using that $\AlgB.\AlgA'[1]$ is biasing signs towards $-$; indeed, this is the reason for the division by $\lambda > 1$ in the term $t_2^\beta/\lambda$ in \Cref{eq:pre-b-minus-4}, without which the proof would not work.}

Note that \Cref{eq:b-plus-4,eq:b-minus-4} establish \Cref{eq:phase4}.

\paragraph{Stronger bound if $\AlgB$ returns $\perp$} We proceed to establish \Cref{eq:phase4-perp}. Note that \Cref{eq:b-plus-4} establishes \Cref{eq:phase4-perp} for $\sigma = -1$, so it remains only to consider $\sigma = 1$. Let $\tau'$ denote the round when \AlgB returns $\perp$. We consider the following cases:
\begin{itemize}
\item If $\half = 1$ during round $\tau'$, then we must have $t_2 = t_0 - t_1 = \lceil t_0/2 \rceil$ and so $t= 2t_0$. In this case, using \Cref{eq:pre-b-minus-4}, we see that
  \begin{align}
    \remainingSigns(\AlgB, +1) \leq & C \lambda^{-b} (n/2)^\alpha (t_0^\beta + \lfloor t_0/2 \rfloor^\beta + \lceil t_0/2 \rceil^\beta/\lambda)\nonumber\\
    \leq &  C \lambda^{-b} (n/2)^\alpha (t_0^\beta + (t_0/2)^\beta + (t_0/2)^\beta/\lambda)\nonumber\\
    \leq & C \lambda^{-b} (n/2)^\alpha t^\beta \cdot ((1/2)^\beta + (1/4)^\beta + (1/4)^\beta/\lambda)\label{eq:perp-case-a},
  \end{align}
  where the second inequality uses the fact that the function $p \mapsto p^\beta + (1-p)^\beta/\lambda$ is increasing for $p \in [0,1/2]$. 
\item If $\half = 0$ during round $\tau'$, then the sign placed in $[l,m]$ on round $\tau'$ removes all plus signs in $[m+1,r]$ placed by $\AlgB.\AlgA[1]$ and $\AlgB.\AlgA'[1]$. Thus, we have
  \begin{align}
    \remainingSigns(\AlgB, +1) \leq & \remainingSigns(\AlgB.\AlgA[0], +1) \nonumber\\
    \leq & C \lambda^{-b} (n/2)^\alpha t_0^\beta \leq C \lambda^{-b} (n/2)^\alpha t^\beta (3/4)^\beta \label{eq:perp-case-b},
  \end{align}
  where the final inequality uses that $t_0 \leq 3t/4$ as a consequence of $t_1 =\lfloor t_0/2 \rfloor$. (Note that we also use here that $t_1 \geq 1$, which is a consequence of the fact that \AlgB has entered phase 3 and that $M \geq 1$.) 
\end{itemize}
Maximizing over \Cref{eq:perp-case-a,eq:perp-case-b} yields the desired conclusion in \Cref{eq:phase4-perp}. 
\end{proof}

\begin{lemma}
  \label{lem:b-perp}
  Suppose that \AlgB returns $\perp$. Then for each $\sigma \in \{-1,1\}$,
  \begin{align}
&\remainingSigns(\AlgB, \sigma) \\
& \ \leq C \cdot \lambda^{-b \cdot \sigma} (n/2)^\alpha t^\beta \cdot \max \left\{ (3/4)^\beta, (1/2)^\beta + (1/4)^\beta + (1/4)^\beta/\lambda, (1+4\lambda/3)/4^\beta \right\}\label{eq:b-perp}
  \end{align}
\end{lemma}
\begin{proof}
  We consider two cases involving the manner in which \AlgB returns $\perp$:
  \paragraph{Case 1: \AlgB returns $\perp$ while in phase 2} Consider the round $\tau$ which is the final round that $\AlgB.\placeSign$ returns a sign (i.e., not $\perp$), and for $h \in \{0,1\}$, let $t_h$ denote the value of $\AlgB.\cntHalf[h]$ following round $\tau$. Since we must have $\AlgB.\phase = 2$ following round $\tau$, it holds that $t_0/t_1 \in [1/2, 2]$. Noting the $t = t_0 + t_1$ in this case, it follows that $\max\{t_0, t_1 \} \leq 2t/3$.

  By symmetry, it is without loss of generality to suppose that $\half = 0$ during the round $\tau_0$ on which $\AlgB$ sets $\phase \gets 2$. Since we have assumed that \AlgB returns $\perp$ while in phase 2, it must be the case that at some round $\tau_1$ after setting $\phase \gets 2$, we have $\half = 1$, at which point \AlgB returns $\perp$. The placement of the sign at round $\tau_0$ removes all $+1$s in $[m+1, r]$ placed by $\AlgB.\AlgA[1]$, and the placement of the sign at round $\tau_1$ removes all $-1$s in $[l,m]$ placed by $\AlgB.\AlgA[0]$. 
Thus, for each $\sigma \in \{-1,1\}$, all remaining signs of type $\sigma$ lie either in $[l,m]$ or $[m+1,r]$, meaning that we have
  \begin{align}
    \remainingSigns(\AlgB, \sigma) \leq & \max\{  \remainingSigns(\AlgB.\AlgA[0], \sigma), \remainingSigns(\AlgB.\AlgA[1], \sigma) \} \nonumber\\
    \leq &  C \lambda^{-b\sigma} (n/2)^\alpha \max\{t_0, t_1 \}^\beta \leq C \lambda^{-b \sigma} (n/2)^\alpha t^\beta \cdot (2/3)^\beta\nonumber.
  \end{align}

  \paragraph{Case 2: $\AlgB$ returns $\perp$ while in phase 4} In this case, we simply apply the bound \Cref{eq:phase4-perp} of \Cref{lem:phase4}. 
\end{proof}

The below lemmas deal with the case that \AlgB does not satisfy the hypotheses of any of \Cref{lem:large-ratio,lem:phase4,lem:b-perp}. 
\begin{lemma}
  \label{lem:small-M}
Suppose that following the last round in which $\AlgB.\placeSign$ returns a sign, we have $\AlgB.\phase \neq 4$. Then for each $\sigma \in \{-1,1\}$, 
  \begin{align}
    \label{eq:small-M}
\remainingSigns(\AlgB, \sigma) \leq  C \lambda^{-b\sigma} n^\alpha t^\beta \cdot 2^{1-\alpha - \beta}.
  \end{align}
\end{lemma}
\begin{proof}
  Suppose that \AlgB is executed for a total of $t$ times overall; for $h \in \{0,1\}$, let $t_h$ denote the value of $\AlgB.\cntHalf$ following the final round $\AlgB.\placeSign$ returns a sign, so that $t_0 + t_1 = t$. By assumption,
  $\AlgB.\AlgA[0], \AlgB.\AlgA[1]$ are never re-initialized. Then by the inductive hypothesis (\Cref{lem:a-induction}) applied to the two instances $\AlgB.\AlgA[0], \AlgB.\AlgA[1]$, it holds that for each $\sigma \in \{-1,1\}$, 
  \begin{align}
\remainingSigns(\AlgB, \sigma) \leq C \lambda^{-b\sigma} (n/2)^\alpha \cdot (t_0^\beta + t_1^\beta) \leq C \lambda^{-b\sigma} n^\alpha t^\beta \cdot 2^{1-\alpha - \beta}\nonumber,
  \end{align}
  where the inequality uses \Cref{lem:2-alpha-beta}.

\end{proof}

\begin{lemma}
  \label{lem:small-M-real}
 Suppose that 
 following the last round in which $\AlgB.\placeSign$ returns a sign, we have $\frac{\AlgB.\cntHalf[0]}{\AlgB.\cntHalf[1]} \in (1/2, 2)$ and $\AlgB.\phase \neq 4$.
   Then $\executionSteps( \AlgB)\leq 6M$. 
 \end{lemma}
 \begin{proof}
  Note that, if $\AlgB.\phase = 3$ at the end of some round $t_0$, then it must be the case that $\frac{\AlgB.\cntHalf[0]}{\AlgB.\cntHalf[1]} \not\in (1/2,2)$ at the end of round $t_0$. (Indeed, to enter phase 3, it must be the case that for some $h \in \{0,1\}$, $\AlgB.\cntHalf[h] \geq 2 \cdot \AlgB.\cntHalf[1-h]$, and as soon as this ceases to be the case, \AlgB enters phase 4.)

  Therefore, letting $t_0$ denote the last round in which $\AlgB.\placeSign$ returns a sign, it must be the case that $\AlgB.\phase \in \{1,2\}$ at the end of round $t_0$. Note that, throughout the entirety of phase 1, there is some $h \in \{0,1\}$ for which $\AlgB.\cntHalf[h] \leq M$. Thus, if \AlgB is in phase 1 upon the end of round $t_0$, \AlgB is executed for at most $M + 2M = 3M$ steps.

  If \AlgB is in phase 2 at the end of round $t_0$, then for some $h \in \{0,1\}$ we must have $\AlgB.\cntHalf[h] \leq 2M$ and $\AlgB.\cntHalf[1-h] \leq 4M$. Thus \AlgB is executed for at most $6M$ steps. 
 \end{proof}

 \subsection{Proof of \Cref{lem:a-induction}}
\label{sec:a-induction-proof}
 Recall our definition of $\lambda$ in \Cref{eq:lambda-C}; we remark that throughout this section we will only need to use the fact that $\lambda \in (1,2)$ (i.e., any other value of $\lambda$ in this interval suffices as well). 
 We introduce the following quantities depending on $\beta \in (0,1)$ and $\delta \in (0,1/3)$, corresponding to the quantities in \Cref{eq:large-ratio,eq:phase4,eq:b-perp,eq:small-M}, respectively (we remark that $F_{\beta,\lambda}(\delta)$ is defined in \Cref{eq:f-delta-beta}): 
\begin{align}
  D_{\ref{eq:large-ratio}} := & (1/3)^\beta + (2/3)^\beta \label{eq:def-d1}\\
  D_{\ref{eq:phase4}} := & \max \left\{ \frac{1+4\lambda/3}{4^\beta}, F_{\beta,\lambda}(\delta)\right\}\label{eq:def-d2}\\
  D_{\ref{eq:b-perp}} := & \max \left\{ (3/4)^\beta, (1/2)^\beta + (1/4)^\beta + (1/4)^\beta/\lambda, (1+4\lambda/3)/4^\beta \right\}\label{eq:def-d3}\\
  D_{\ref{eq:small-M}} := & 2^{1-\beta}\label{eq:def-d4}. 
\end{align}

 We are now ready to prove the (inductive step) of \Cref{lem:a-induction}. 
\begin{proof}[Proof of \Cref{lem:a-induction}]
Fix $\delta=1/100$, and choose $\beta\in(0,1)$ and $\vep>0$ as in \Cref{lem:bound-d-constants}. Decreasing $\vep$ if necessary, assume $\vep<1-\beta$, and set $\alpha:=1-\beta-\vep>0$. We prove the bound by induction on $d$, where $n=2^d$ is the number of cells covered by \AlgA. The case $t=0$ is immediate, so assume $t\geq1$.

\paragraph{Base cases}
If $n=1$, at most one sign remains. If $b\sigma>0$, the leaf rule never returns $\sigma$; otherwise $\lambda^{-b\sigma}\geq1$. In either case,
$\remainingSigns(\AlgA,\sigma)\leq C\lambda^{-b\sigma}n^\alpha t^\beta$.

We also handle every execution with $t\leq6$ directly. If $b\sigma\leq3$, then $C\geq6\lambda^3$ gives
\[
\remainingSigns(\AlgA,\sigma)\leq6
\leq C\lambda^{-b\sigma}n^\alpha t^\beta.
\]
Suppose instead that $b\sigma>3$. A bias changes only when an instance of \AlgB enters phase 4 and initializes a fresh child with bias differing by one. Entry into phase 3 requires at least $3M+1\geq4$ sign-returning rounds, and entry into phase 4 occurs on a later round. Thus each bias-changing initialization requires at least four earlier sign-returning rounds in its parent instance. Along the lineage of a particular returned sign, these preceding rounds for successive bias changes are disjoint, because each change initializes a fresh child. Two such changes would therefore require at least nine rounds, including the round returning the sign. Since $t\leq6$, every leaf that returns a sign has bias $b'$ satisfying $|b'-b|\leq1$. Hence $b'\sigma>0$, so that leaf cannot return $\sigma$. This proves $\remainingSigns(\AlgA,\sigma)=0$ and completes the short-execution case.

\paragraph{Completed instances}
Now suppose $n=2^d$ with $d\geq1$ and $t>6$, and assume the conclusion for every \AlgA instance covering $2^j$ cells with $0\leq j<d$. Let $\AlgB_1,\ldots,\AlgB_k$ be the instances initialized during the execution of \AlgA. Write $t_i:=\executionSteps(\AlgB_i)$ and let $M_i$ be the parameter passed to $\AlgB_i.\initialize$. Then
\[
t=\sum_{i=1}^k t_i,\qquad M_1=1,\qquad M_i=t_{i-1}+1\quad(2\leq i\leq k).
\]
The last identity includes the increment of $\AlgA.\cnt$ on the call that returns $\perp$. Each of $\AlgB_1,\ldots,\AlgB_{k-1}$ returns $\perp$, and an instance can do so only after at least $2M_i$ sign-returning rounds. Consequently $t_{i+1}\geq2M_{i+1}\geq2t_i$ for $1\leq i\leq k-2$. Applying \Cref{lem:b-perp,lem:geometric-ti} to these completed instances gives
\begin{align}
\sum_{i=1}^{k-1}\remainingSigns(\AlgB_i,\sigma)
&\leq C\lambda^{-b\sigma}(n/2)^\alpha
 D_{\ref{eq:b-perp}}\sum_{i=1}^{k-1}t_i^\beta \nonumber\\
&\leq C\lambda^{-b\sigma}(n/2)^\alpha
 \frac{D_{\ref{eq:b-perp}}}{2^\beta-1}
 \left(\sum_{i=1}^{k-1}t_i\right)^\beta.
\label{eq:remsigns-bi}
\end{align}
When $k=1$, both sums are empty and this inequality is immediate. No growth bound is required for $t_k$.

Write $p_k:=t_k/t\in[0,1]$. Let $\phi_k$ and $c_h$, for $h\in\{0,1\}$, be the values of $\AlgB_k.\phase$ and $\AlgB_k.\cntHalf[h]$ immediately after its last sign-returning round, so that $c_0+c_1=t_k$. We use the bound
$\remainingSigns(\AlgA,\sigma)\leq\sum_{i=1}^k\remainingSigns(\AlgB_i,\sigma)$
and distinguish two cases.

\paragraph{Case 1: $\phi_k=4$, or $\phi_k\ne4$ and $\max\{c_0,c_1\}\geq2\min\{c_0,c_1\}$}
If $\phi_k=4$, apply \Cref{lem:phase4}. Otherwise, apply \Cref{lem:large-ratio}; if one counter is zero, the same bound follows directly from the inductive hypothesis and $D_{\ref{eq:large-ratio}}\geq1$. Thus
\[
\remainingSigns(\AlgB_k,\sigma)
\leq C\lambda^{-b\sigma}(n/2)^\alpha t_k^\beta
\max\{D_{\ref{eq:large-ratio}},D_{\ref{eq:phase4}}\}.
\]
Combining this with \Cref{eq:remsigns-bi} yields
\begin{align}
\remainingSigns(\AlgA,\sigma)
\leq {}& C\lambda^{-b\sigma}(n/2)^\alpha t^\beta
\Bigg(
\frac{D_{\ref{eq:b-perp}}(1-p_k)^\beta}{2^\beta-1}
\nonumber\\
&\hspace{18mm}
+p_k^\beta\max\{D_{\ref{eq:large-ratio}},D_{\ref{eq:phase4}}\}
\Bigg).
\label{eq:case1-remsigns}
\end{align}
This estimate holds for every $p_k\in[0,1]$, including $p_k=1$ when $k=1$.

\paragraph{Case 2: $\phi_k\ne4$ and $\max\{c_0,c_1\}<2\min\{c_0,c_1\}$}
The half-counts are balanced and \Cref{lem:small-M-real} gives $t_k\leq6M_k$. If $k=1$, then $M_1=1$ and $t=t_k\leq6$, which has already been handled. Hence $k\geq2$. Since the completed instance $\AlgB_{k-1}$ has $t_{k-1}\geq2$, the restart rule gives
\[
M_k=t_{k-1}+1
\leq\frac32t_{k-1}
\leq\frac32\sum_{i=1}^{k-1}t_i
=\frac32(t-t_k).
\]
It follows that $t_k\leq9(t-t_k)$ and therefore $p_k\leq9/10$.

Since $\phi_k\ne4$, \Cref{lem:small-M} applies and gives
\[
\remainingSigns(\AlgB_k,\sigma)
\leq C\lambda^{-b\sigma}(n/2)^\alpha t_k^\beta
D_{\ref{eq:small-M}}.
\]
Combining this with \Cref{eq:remsigns-bi} yields
\begin{align}
\remainingSigns(\AlgA,\sigma)
\leq {}& C\lambda^{-b\sigma}(n/2)^\alpha t^\beta
\left(
\frac{D_{\ref{eq:b-perp}}(1-p_k)^\beta}{2^\beta-1}
+p_k^\beta D_{\ref{eq:small-M}}
\right).
\label{eq:case2-remsigns}
\end{align}

\paragraph{Wrapping up}
Apply the first maximum in \Cref{lem:bound-d-constants} to \Cref{eq:case1-remsigns}, using $p_k\in[0,1]$, and the second maximum to \Cref{eq:case2-remsigns}, using $p_k\in[0,9/10]$. In either case,
\[
\remainingSigns(\AlgA,\sigma)
\leq C\lambda^{-b\sigma}n^\alpha t^\beta
2^{1-\alpha-\beta-\vep}
=C\lambda^{-b\sigma}n^\alpha t^\beta.
\]
Since $\alpha+\beta=1-\vep<1$, this completes the induction.
\end{proof}


\section{An Oblivious Lower Bound }
\label{sec:oblivious-main}
In this section, we overview the proof of \Cref{col:obl_lower}, which gives an oblivious strategy for the adversary in the calibration game. Our approach for this task proceeds by constructing an oblivious strategy for \PlayerP in the sign preservation game which yields a lower bound on the expected number of signs remaining. We in fact construct such a strategy for \PlayerP which has the following two additional properties, formalized in \Cref{def:wc-pp}:  (a) the lower bound on the number of signs remaining is ``worst case'' in the sense that at each time step $i \in [s]$, conditioned on any history up to time step $i$, the sign placed at time step $i$ will remain with at least some given probability; and (b) it never re-uses a cell (i.e., it abides by the rules of the original Sign-Preservation Game in \cite{qiao2021stronger}).
\begin{definition}
  \label{def:wc-pp}
  Fix $s, n \in \BN$ and $\ep \in (0,1)$. An \emph{oblivious no-reuse strategy for \PlayerP in the sign-preservation game} with $n$ \emph{cells}, $s$ \emph{rounds},  and \emph{worst-case preservation probability} $\epsilon$ is a distribution $\MP$ over $[n]^{s}$, satisfying the following:
  \begin{enumerate}[leftmargin=0.4cm]
    \item For all $i \in [s]$ and $k_1', \ldots, k_i' \in [n]$ for which $\Pr_{ \MP}(k_1 = k_1', \ldots, k_i = k_i') > 0$, we have
      \begin{align}
        \label{eq:sign-upper}
    \Pr_{(k_1, \ldots, k_s) \sim \MP} \left[\forall j,~ i <j \le s:~ k_j > k_i \mid k_1=k_1',\dots,k_i=k_i',~s \ge i\right] \ge \epsilon \\
      \label{eq:sign-lower}
    \Pr_{(k_1, \ldots, k_s) \sim \MP}\left[\forall j,~ i <j \le s:~ k_j < k_i \mid k_1=k_1',\dots,k_i=k_i',~s \ge i\right] \ge \epsilon.
    \end{align}
\item With probability $1$ over $(k_1, \ldots, k_s) \sim \MP$, $k_1, \ldots, k_s$ are distinct. 
\end{enumerate}
\end{definition}
The distribution $\MP$ over $[n]^s$ corresponds to an oblivious strategy for \PlayerP in the natural way: \PlayerP draws a sample $(k_1, \ldots, k_s) \sim \MP$ and then plays $k_i \in [n]$ at each round $i \in [s]$. Note that  \Cref{eq:sign-upper,eq:sign-lower} ensure that, conditioned on the history $(k_1', \ldots, k_i')$, no matter which sign \PlayerL chooses to place in cell $k_i'$ at round $i$, it will remain with probability at least $\ep$. Thus, the existence of a distribution $\MP$ satisfying the conditions of \Cref{def:wc-pp} ensures that $\opt(n, s) \geq \ep \cdot s$. Our lower bound for calibration, however, will not directly use this lower bound on $\opt(n,s)$; instead, we directly use the properties of \PlayerP's strategy from \Cref{def:wc-pp} to derive the following theorem:
\begin{theorem}
  \label{thm:oblivious-lb}
    Suppose there is a no-reuse strategy for \PlayerP in the sign-preservation game with $n$ cells, $s$ rounds, and worst-case preservation probability at least $\epsilon$ (\Cref{def:wc-pp}). Then for any $T \in \BN$ there is an oblivious adversary for the calibration problem with $T$ rounds which causes any forecaster to experience an expected error of at least
$
    \Omega\left(
        \min\left(
            \epsilon \sqrt{T s},~ \frac{\epsilon T}{n}
        \right)
    \right).
$
  \end{theorem}
The proof of \Cref{col:obl_lower} follows by constructing a strategy for \PlayerP satisfying \Cref{def:wc-pp} for an appropriate choice of the parameters $\ep, s, T$ (see \Cref{lem:tree_spr}), and then applying \Cref{thm:oblivious-lb}. 
We remark that if $s = n^\alpha$ and $\ep = n^{\beta-\alpha}$ (which corresponds to the setup of \Cref{thm:adapt_lower} in that existence of an adversary obtaining worst-case preservation probability of $\ep = n^{\beta-\alpha}$ implies $\opt(n,n^\alpha) \geq n^\beta$), then the bound of \Cref{thm:oblivious-lb} yields expected error of $\Omega(T^{\frac{\beta+1}{\alpha+2}})$, matching the quantitative bound of \Cref{thm:adapt_lower} up to logarithmic factors. (We emphasize that neither theorem implies the other since both the assumptions and conclusion of \Cref{thm:oblivious-lb} are qualitatively stronger.) 

  \begin{algorithm}
    \caption{Calibration adversary for \Cref{thm:oblivious-lb}}
    \label{alg:oblivious-lb}
    \begin{algorithmic}[1]
      \Require Number of cells $n$, number of time steps $s$ for sign preservation game, preservation probability $\ep$, number of rounds $T$ for calibration. Distribution $\MP \in \Delta([n]^{s})$ representing oblivious adversary for sign preservation game.
\State If $T<s$, output $T$ independent fair bits and stop
\State Partition $[T]$ into $s$ consecutive batches whose sizes differ by at most one
\State Draw $(k_1, \ldots, k_s) \sim \MP$. 
\For{batch $i \in [s]$}
\For{round $t$ in batch $i$}
\State Choose $q_t := \frac 14 + \frac{k_i}{2n}$. 
\State Draw $X_t \sim \Ber(q_t)$, and play $X_t$ as the outcome at round $t$. 
\EndFor
\EndFor
    \end{algorithmic}
  \end{algorithm}

The adversary used to prove \Cref{thm:oblivious-lb} is shown in \Cref{alg:oblivious-lb}.  
When $T<s$, independent fair bits suffice, as shown in the proof. For $T\ge s$, split the $T$ rounds into $s$ consecutive batches whose sizes differ by at most one; each has at most $2T/s$ rounds. The adversary draws a sequence $(k_1, \ldots, k_s) \in [n]^{s}$ realizing the assumed lower bound for the sign-preservation game. 
At each iteration $t$ belonging to batch $i \in [s]$, the adversary chooses outcome $X_t \in \{0,1\}$ by setting $X_t = 1$ probability given by the value $\frac 14  + \frac{k_i}{2n}$ (denoted by $q_t$). In other words, at each round $t$ in batch $i \in [s]$, 
the conditional expectation of $X_t$ given the realization of $(k_1, \ldots, k_s)$ is $q_t = \frac 14 + \frac{k_i}{2n}$. 
We let the resulting joint distribution over $(q_1, \ldots, q_T) \in [0,1]^T$ be denoted by $\MQ$. Moreover, we let $v := \frac{3}{16}$, which satisfies that $\Var(X_t \mid q_t) \geq v$ for all $t$. 

  Consider any forecaster algorithm; denote its prediction at time $t$ by $p_t \in P$, where $P$ is the fixed finite prediction set. Note that $p_t$ may be expressed as a randomized function of $X_1, \ldots, X_{t-1}$. Recall that $\calE_t(p) = \sum_{i \in [t] :\ p_i = p} (p-X_i)$ denotes the signed, unnormalized calibration error of the prediction $p$, up to time $t$. We wish to lower bound the expectation of the calibration error $\sum_{p \in P} |\calE_t(p)|$. 

   For $p \in [0,1]$, define the random variable $F(p) \in [T]$ to be $F(p) := \max\bigl(\{1\}\cup\{t\in\{2,\ldots,T\}:\sgn(p-q_t)\ne\sgn(p-q_{t-1})\}\bigr)$. If $\sgn(p-q_t)$ are all equal, we take $F(p) = 1$ as a matter of convention. 

   \paragraph{Proof overview for \Cref{thm:oblivious-lb}} 
   To prove \Cref{thm:oblivious-lb}, we use a potential-based argument. Set $\Delta := \min\{1/(8n), \sqrt{s/(2T)} \}$, so that we aim to show a lower bound of $\Omega(\ep \Delta T)$ on calibration error. We define
   \begin{align}
\tRbias(x,a) := a \cdot x, \qquad  \tRvar(x) := \frac{\Delta x^2}{2}, \qquad \tilde R(x,a) := \tRbias(x,a) +  \tRvar(x)\label{eq:r-rough}.
   \end{align}
   We define a potential function $\Psi_t(p)$ associated to each point $p \in [0,1]$, which satisfies:\footnote{Technically, $\Psi_t(p)$ depends on the predictions, means, and outcomes through round $t$; we omit this dependence from our notation.}
   \begin{align}
\Psi_t(p) \approx \tilde R(E_t(p) \cdot \sgn(p-q_t), \Pr[F(p) \leq t \mid q_{1:t}])\label{eq:psit-rough}.
   \end{align}
   We emphasize that the actual definition of $\Psi_t(p)$ (in \Cref{eq:psit-define}) has a few additional components which are needed for technical reasons, hence the ``$\approx$'' in \Cref{eq:psit-rough}. 
   Overall, we will show that the expected calibration error is lower bounded by $\frac67\sum_{p \in P} \E[\Psi_T(p)]$ (in \Cref{lem:et-psit}). The bulk of the proof of \Cref{thm:oblivious-lb} consists in lower bounding this latter quantity by $\Omega(\ep \Delta v \cdot T) \geq \Omega(\ep \Delta T)$. In turn, to do so, we will show that  
   $\sum_p \Psi_t(p)$ increases in expectation by $\Omega(\ep \Delta v)$ from round $t-1$ to $t$.

  To separate the effect of the next outcome from the changing conditional preservation probability, the formal proof introduces $\Phi_t(p)$ immediately before the outcome and $\Psi_t(p)$ immediately after it. Write $H_t=(p_{1:t},q_{1:t},X_{1:t-1})$. Conditional on this history, the signed increment satisfies
\[
\E[(\calE_t(p)-\calE_{t-1}(p))\sgn(p-q_t)\mid H_t]
=\II[p_t=p]|p-q_t|.
\]
If $|p-q_t|\ge\Delta$, this drift, together with the positive minimum slope of the corrected potential, supplies the required expected increase. If $|p-q_t|\le\Delta$, the outcome variance and the quadratic part of the potential supply the increase; an additional term handles errors outside the quadratic region. These are conditional-expectation statements: an individual outcome can move the signed error in either direction, and crossing zero can increase absolute error even when the drift points toward zero. The precise estimate is \Cref{lem:phi-psi}.

Between outcomes, a crossing of $q_{t+1}$ past $p$ makes the conditional probability of $F(p)\le t$ zero. If there is no crossing, the conditional probabilities satisfy the tower law. The modified bias term uses these facts to ensure $\E[\Phi_{t+1}(p)\mid H_t,X_t]\ge\Psi_t(p)$, as proved in \Cref{lem:phi1-psi}. Summing these two types of increment over time and over the fixed finite set $P$ gives the desired lower bound.


\section{Oblivious lower bound for the Sign Preservation Game}
In \Cref{lem:tree_spr} below, we give an \emph{oblivious strategy} on \PlayerP which yields a lower bound on the expected number of signs remaining in the sign-preservation game (thus implying lower bounds on $\opt(n,s)$ for various values of $n,s$). In fact, we show a stronger lower bound, namely on the \emph{worst-case preservation probability}, which is needed to apply \Cref{thm:oblivious-lb} to obtain a corresponding oblivious strategy for the adversary in the sequential calibration game. 
\begin{lemma}\label{lem:tree_spr}
  For any $d,k \in \mathbb{N}$ with $d \ge k$, there exists an oblivious strategy for \PlayerP in the sign-preservation game with $n = \binom{d}{k} 2^{d-k}$ cells and $s = \binom{d}{k}$ timesteps which satisfies the following: the worst-case preservation probability for this oblivious strategy (per \Cref{def:wc-pp}) is at least $2^{-k}$. 
  In particular, the expected number of preserved signs is at least $s \cdot 2^{-k}$, i.e., $\opt(n, s) \geq s \cdot 2^{-k}$. 
\end{lemma}
\begin{proof}
  First we describe the oblivious strategy of \PlayerP. Let us associate to each time step $t \in [s]$ a unique  string $\ue{w}{t}=(\ue{w}{t}_1,\dots,\ue{w}{t}_d) \in \{0,1\}^d$ containing $d-k$ ones and $k$ zeros. We consider the lexicographic order on the strings $\ue{w}{t}$, so that for $t < t'$ we can write $\ue{w}{t} < \ue{w}{t'}$. 
  Similarly, we associate to each $i \in [n]$ a unique string $\ue{q}{i} \in \{-1,0,1\}^d$ with exactly $k$ zeros, again ordered lexicographically, so that $i < j$ if and only if $\ue{q}{i} < \ue{q}{j}$.

  We define a distribution $\MP$ over tuples $(\ue{q}{1}, \ldots, \ue{q}{s}) \in [n]^s$, where each $\ue{q}{t} \in \{-1,0,1\}^d$ ($t \in [s]$) is viewed as an element of $[n]$ as described above. In particular, $\ue{q}{t}$ represents the play of \PlayerP at step $t$. To draw a sample from $\MP$:
  \begin{itemize}
  \item 
    For each string $u \in \{0,1\}^{\leq d}$,\footnote{Here $\{0,1\}^{\leq d}$ denotes the union of $\{0,1\}^\ell$ for all $\ell \leq d$.} draw a Rademacher random variable $\xi_u \sim \mathrm{Unif}(\{-1,1\})$, so that all the $\xi_u$ variables are i.i.d.
  \item 
    Define $\ue{q}{t}$ by, for $\ell \in [d]$,  $\ue{q}{t}_\ell := \ue{w}{t}_\ell \xi_{\ue{w}{t}_{[1:\ell-1]}}$, where $\ue{w}{t}_{[1:\ell-1]}:=(\ue{w}{t}_1,\dots,\ue{w}{t}_{\ell-1})$. In words, $\ue{q}{t}$ is obtained from $\ue{w}{t}$ by multiplying each entry with a Rademacher random variable that is indexed by the previous string elements.
  \end{itemize}
  The zero coordinates of $\ue{q}{t}$ are exactly those of $\ue{w}{t}$. Since these zero patterns are distinct, the strategy never reuses a cell.
  Now we analyze the worst-case preservation probability of a sign (per \Cref{def:wc-pp}). Fix some sequence $\ue{q}{1},\dots,\ue{q}{t}$ and let us first lower bound the probability (under $\MP$) that for all $t' >t$, $\ue{q}{t'} > \ue{q}{t}$, conditioned on $\ue{q}{1},\dots,\ue{q}{t}$. 
  Denote by $\ell_1,\dots,\ell_k \in [d]$ the $k$ indices of 0 coordinates of $\ue{w}{t}$. We argue that if $\xi_{\ue{w}{t}_{[1:\ell_j-1]}}=1$ for all $j\in[k]$, then $\ue{q}{t'}>\ue{q}{t}$ for all $t'>t$. Indeed, fix $t'>t$. Let $\ell \in [d]$ be the smallest index for which $\ue{w}{t'}_\ell \ne \ue{w}{t}_\ell$. Since $\ue{w}{t'} > \ue{w}{t}$ in the lexicographic ordering, we must have $1 = \ue{w}{t'}_\ell>\ue{w}{t}_\ell = 0$. 
  Thus we may choose $j \in [k]$ so that $\ell=\ell_j$. 
  Note that $\ue{w}{t}_{[1:\ell-1]} = \ue{w}{t'}_{[1:\ell-1]}$ for all $\ell < \ell_j$ by definition of $\ell_j$ as the smallest index at which $\ue{w}{t'}, \ue{w}{t}$ differ.  Hence, for all $\ell < \ell_j$, we have $\ue{q}{t}_\ell = \ue{w}{t}_\ell \xi_{\ue{w}{t}_{[1:\ell-1]}} = \ue{w}{t'}_\ell \xi_{\ue{w}{t'}_{[1:\ell-1]}} = \ue{q}{t'}_\ell$, 
  and so by assumption that $\xi_{\ue{w}{t}_{[1:\ell_j-1]}}=1$, we have that $\ue{q}{t'}>\ue{q}{t}$ as required.

  It remains to estimate the probability that $\xi_{\ue{w}{t}_{[1:\ell_j-1]}}=1$ for all $j \in [k]$, conditioned on $\ue{q}{1},\dots,\ue{q}{t}$. Fix a zero position $\ell_j$ and its prefix $u=\ue{w}{t}_{[1:\ell_j-1]}$. For every earlier string $\ue{w}{i}$, including $i=t$, either its prefix of this length differs from $u$, or its coordinate at $\ell_j$ is zero: otherwise it would be lexicographically larger than $\ue{w}{t}$. Thus no earlier displayed coordinate uses $\xi_u$. Coordinates at other depths use coins indexed by strings of different lengths. The $k$ coins in question have distinct indices, so they remain independent unbiased Rademacher variables conditional on the entire revealed history. Consequently, the probability that all are $1$ is $2^{-k}$. Hence, we proved that conditioned on $\ue{q}{1},\dots,\ue{q}{t}$, with probability at least $2^{-k}$, $\ue{q}{t'}>\ue{q}{t}$ for all $t'>t$. A symmetric argument shows that with probability at least $2^{-k}$, $\ue{q}{t'}<\ue{q}{t}$ for all $t'>t$. Consequently, the worst-case preservation probability is at least $2^{-k}$.
\end{proof}

\section*{Acknowledgements}
CD is supported by NSF Awards CCF-1901292, DMS-2022448, and DMS2134108, a
Simons Investigator Award, and the Simons Collaboration on the Theory of Algorithmic Fairness. NG is supported by a Fannie \& John Hertz Foundation Fellowship and an NSF Graduate Fellowship. PO is supported by a Linkedin-Cornell Fellowship Grant.

\bibliographystyle{alpha}
\bibliography{ref}

@misc{qiao2021stronger,
      title={Stronger Calibration Lower Bounds via Sidestepping}, 
      author={Mingda Qiao and Gregory Valiant},
      year={2021},
      eprint={2012.03454},
      archivePrefix={arXiv},
      primaryClass={cs.LG}
}

@misc{hart2023minimax,
      title={Calibrated Forecasts: The Minimax Proof}, 
      author={Sergiu Hart},
      year={2023},
      eprint={2209.05863},
      archivePrefix={arXiv},
      primaryClass={econ.TH}
}

@article{foster1998asymptotic,
  title={Asymptotic calibration},
  author={Foster, Dean P and Vohra, Rakesh V},
  journal={Biometrika},
  volume={85},
  number={2},
  pages={379--390},
  year={1998}
}

@article{dawid1982bayesian,
author = { A. P.   Dawid },
title = {The Well-Calibrated Bayesian},
journal = {Journal of the American Statistical Association},
volume = {77},
number = {379},
pages = {605-610},
year  = {1982},
publisher = {Taylor & Francis},
doi = {10.1080/01621459.1982.10477856}
}

@article{brier1950verification,
  author = {Glenn W. Brier},
  title = {Verification of forecasts expressed in terms of probability},
  journal = {Monthly Weather Review},
  volume = {78},
  number = {1},
  pages = {1--3},
  year = {1950}
}

@inproceedings{guo2017calibration,
  author = {Chuan Guo and Geoff Pleiss and Yu Sun and Kilian Q. Weinberger},
  title = {On calibration of modern neural networks},
  booktitle = {International Conference on Machine Learning (ICML)},
  pages = {1321--1330},
  year = {2017}
}

@article{jung2020moment,
  author = {Christopher Jung and Changhwa Lee and Mallesh M. Pai and Aaron Roth and Rakesh Vohra},
  title = {Moment multicalibration for uncertainty estimation},
  journal = {arXiv preprint arXiv:2008.08037},
  year = {2020}
}

@inproceedings{kuleshov2015calibrated,
  author = {Volodymyr Kuleshov and Percy S. Liang},
  title = {Calibrated structured prediction},
  booktitle = {Advances in Neural Information Processing Systems (NIPS)},
  pages = {3474--3482},
  year = {2015}
}

@inproceedings{kumar2019verified,
  author = {Ananya Kumar and Percy S. Liang and Tengyu Ma},
  title = {Verified uncertainty calibration},
  booktitle = {Advances in Neural Information Processing Systems (NeurIPS)},
  pages = {3792--3803},
  year = {2019}
}

@inproceedings{kleinberg2017inherent,
  author = {Jon Kleinberg and Sendhil Mullainathan and Manish Raghavan},
  title = {Inherent trade-offs in the fair determination of risk scores},
  booktitle = {Innovations in Theoretical Computer Science Conference (ITCS)},
  pages = {43:1--43:23},
  year = {2017}
}

@inproceedings{pleiss2017fairness,
  author = {Geoff Pleiss and Manish Raghavan and Felix Wu and Jon Kleinberg and Kilian Q. Weinberger},
  title = {On fairness and calibration},
  booktitle = {Advances in Neural Information Processing Systems (NIPS)},
  pages = {5680--5689},
  year = {2017}
}

@article{shabat2020sample,
  author = {Eliran Shabat and Lee Cohen and Yishay Mansour},
  title = {Sample complexity of uniform convergence for multicalibration},
  journal = {arXiv preprint arXiv:2005.01757},
  year = {2020}
}

@inproceedings{zhao2020individual,
  author = {Shengjia Zhao and Tengyu Ma and Stefano Ermon},
  title = {Individual calibration with randomized forecasting},
  booktitle = {International Conference on Machine Learning (ICML)},
  pages = {8366--8376},
  year = {2020}
}

@article{murphy1977reliability,
 ISSN = {00359254, 14679876},
 URL = {http://www.jstor.org/stable/2346866},
 author = {Allan H. Murphy and Robert L. Winkler},
 journal = {Journal of the Royal Statistical Society. Series C (Applied Statistics)},
 number = {1},
 pages = {41--47},
 publisher = {[Wiley, Royal Statistical Society]},
 title = {Reliability of Subjective Probability Forecasts of Precipitation and Temperature},
 urldate = {2024-05-22},
 volume = {26},
 year = {1977}
}

@article{oakes1985self,
author = {David Oakes},
title = {Self-Calibrating Priors Do Not Exist},
journal = {Journal of the American Statistical Association},
volume = {80},
number = {390},
pages = {339--339},
year = {1985},
publisher = {Taylor \& Francis},
doi = {10.1080/01621459.1985.10478117}}

@article{murphy1967verification,
 ISSN = {00218952, 2163534X},
 URL = {http://www.jstor.org/stable/26173494},
 author = {Allan H. Murphy and Edward S. Epstein},
 journal = {Journal of Applied Meteorology (1962-1982)},
 number = {5},
 pages = {748--755},
 publisher = {American Meteorological Society},
 title = {Verification of Probabilistic Predictions : A Brief Review},
 urldate = {2024-05-22},
 volume = {6},
 year = {1967}
}

@InProceedings{herbert2018multicalibration,
  title = 	 {Multicalibration: Calibration for the ({C}omputationally-Identifiable) Masses},
  author =       {Hebert-Johnson, Ursula and Kim, Michael and Reingold, Omer and Rothblum, Guy},
  booktitle = 	 {Proceedings of the 35th International Conference on Machine Learning},
  pages = 	 {1939--1948},
  year = 	 {2018},
  editor = 	 {Dy, Jennifer and Krause, Andreas},
  volume = 	 {80},
  series = 	 {Proceedings of Machine Learning Research},
  month = 	 {10--15 Jul},
  publisher =    {PMLR},
  url = 	 {https://proceedings.mlr.press/v80/hebert-johnson18a.html}
}

@inproceedings{garg2024oracle,
author = {Sumegha Garg and Christopher Jung and Omer Reingold and Aaron Roth},
title = {Oracle Efficient Online Multicalibration and Omniprediction},
booktitle = {Proceedings of the 2024 Annual ACM-SIAM Symposium on Discrete Algorithms (SODA)},
year={2024},
chapter = {},
pages = {2725-2792},
doi = {10.1137/1.9781611977912.98},
URL = {https://epubs.siam.org/doi/abs/10.1137/1.9781611977912.98},
eprint = {https://epubs.siam.org/doi/pdf/10.1137/1.9781611977912.98}
}

@inproceedings{
gopalan2023swap,
title={Swap Agnostic Learning, or Characterizing Omniprediction via Multicalibration},
author={Parikshit Gopalan and Michael P. Kim and Omer Reingold},
booktitle={Thirty-seventh Conference on Neural Information Processing Systems},
year={2023},
url={https://openreview.net/forum?id=IzlRh5qwmG}
}

@inproceedings{
minderer2021revisiting,
title={Revisiting the Calibration of Modern Neural Networks},
author={Matthias Minderer and Josip Djolonga and Rob Romijnders and Frances Ann Hubis and Xiaohua Zhai and Neil Houlsby and Dustin Tran and Mario Lucic},
booktitle={Advances in Neural Information Processing Systems},
editor={A. Beygelzimer and Y. Dauphin and P. Liang and J. Wortman Vaughan},
year={2021},
url={https://openreview.net/forum?id=QRBvLayFXI}
}

@InProceedings{kleinberg2023ucalibration,
  title = 	 {U-Calibration: Forecasting for an Unknown Agent},
  author =       {Kleinberg, Bobby and Leme, Renato Paes and Schneider, Jon and Teng, Yifeng},
  booktitle = 	 {Proceedings of Thirty Sixth Conference on Learning Theory},
  pages = 	 {5143--5145},
  year = 	 {2023},
  editor = 	 {Neu, Gergely and Rosasco, Lorenzo},
  volume = 	 {195},
  series = 	 {Proceedings of Machine Learning Research},
  month = 	 {12--15 Jul},
  publisher =    {PMLR},
  url = 	 {https://proceedings.mlr.press/v195/kleinberg23a.html}
}

@inproceedings{blasiok2023when,
 author = {Blasiok, Jaroslaw and Gopalan, Parikshit and Hu, Lunjia and Nakkiran, Preetum},
 booktitle = {Advances in Neural Information Processing Systems},
 editor = {A. Oh and T. Naumann and A. Globerson and K. Saenko and M. Hardt and S. Levine},
 pages = {72071--72095},
 publisher = {Curran Associates, Inc.},
 title = {When Does Optimizing a Proper Loss Yield Calibration?},
 url = {https://proceedings.neurips.cc/paper_files/paper/2023/file/e4165c96702bac5f4962b70f3cf2f136-Paper-Conference.pdf},
 volume = {36},
 year = {2023}
}

@article{crowson2016assessing,
  author = {Cynthia S. Crowson and Elizabeth J. Atkinson and Terry M. Therneau},
  title = {Assessing calibration of prognostic risk scores},
  journal = {Statistical Methods in Medical Research},
  volume = {25},
  number = {4},
  pages = {1692--1706},
  year = {2016}
}

@article{jiang2012calibrating,
  author = {Xiaoqian Jiang and Melanie Osl and Jihoon Kim and Lucila Ohno-Machado},
  title = {Calibrating predictive model estimates to support personalized medicine},
  journal = {Journal of the American Medical Informatics Association},
  volume = {19},
  number = {2},
  pages = {263--274},
  year = {2012},
  month = {Mar--Apr},
  doi = {10.1136/amiajnl-2011-000291},
  eprint = {PMC3277613},
  note = {Epub 2011 Oct 7. PMID: 21984587; PMCID: PMC3277613}
}

@book{cesabianchi2006prediction,
author = {Cesa-Bianchi, Nicolo and Lugosi, Gabor},
title = {Prediction, Learning, and Games},
year = {2006},
isbn = {0521841089},
publisher = {Cambridge University Press},
address = {USA}
}

@article{kakade2008deterministic,
  author = {Kakade, Sham M. and Foster, Dean P.},
  title = {Deterministic Calibration and Nash Equilibrium},
  journal = {Journal of Computer and System Sciences},
  volume = {74},
  number = {1},
  pages = {115--130},
  year = {2008}
}

@inproceedings{qiao2024distance,
  author = {Qiao, Mingda and Zheng, Letian},
  title = {On the Distance from Calibration in Sequential Prediction},
  booktitle = {Conference on Learning Theory (COLT)},
  pages = {4307--4357},
  year = {2024}
}

@inproceedings{arunachaleswaran2025elementary,
  author = {Arunachaleswaran, Eshwar Ram and Collina, Natalie and Roth, Aaron and Shi, Mirah},
  title = {An Elementary Predictor Obtaining Distance to Calibration},
  booktitle = {Proceedings of the 2025 Annual ACM-SIAM Symposium on Discrete Algorithms (SODA)},
  pages = {1366--1370},
  year = {2025}
}

@inproceedings{fishelson2025fullswap,
  author = {Fishelson, Maxwell and Kleinberg, Robert and Okoroafor, Princewill and Paes Leme, Renato and Schneider, Jon and Teng, Yifeng},
  title = {Full Swap Regret and Discretized Calibration},
  booktitle = {Algorithmic Learning Theory (ALT)},
  pages = {444--480},
  year = {2025}
}

@misc{luo2025simultaneous,
  title = {Simultaneous Swap Regret Minimization via KL-Calibration},
  author = {Luo, Haipeng and Senapati, Spandan and Sharan, Vatsal},
  year = {2025},
  eprint = {2502.16387},
  archivePrefix = {arXiv},
  primaryClass = {cs.LG}
}

@inproceedings{hu2024predict,
  author = {Hu, Lunjia and Wu, Yifan},
  title = {Predict to Minimize Swap Regret for All Payoff-Bounded Tasks},
  booktitle = {Foundations of Computer Science (FOCS)},
  pages = {244--263},
  year = {2024}
}

@inproceedings{haghtalab2024truthfulness,
  author = {Haghtalab, Nika and Qiao, Mingda and Yang, Kunhe and Zhao, Eric},
  title = {Truthfulness of Calibration Measures},
  booktitle = {Advances in Neural Information Processing Systems (NeurIPS)},
  pages = {117237--117290},
  year = {2024}
}

@inproceedings{qiao2025truthfulness,
  author = {Qiao, Mingda and Zhao, Eric},
  title = {Truthfulness of Decision-Theoretic Calibration Measures},
  booktitle = {Conference on Learning Theory (COLT)},
  pages = {4686--4739},
  year = {2025}
}

@misc{hartline2025perfectly,
  title = {A Perfectly Truthful Calibration Measure},
  author = {Hartline, Jason and Hu, Lunjia and Wu, Yifan},
  year = {2025},
  eprint = {2508.13100},
  archivePrefix = {arXiv},
  primaryClass = {cs.LG}
}

\appendix

\section{Proofs for \Cref{sec:equivalence}}

\subsection{Proof of \Cref{col:main_eq}}
\label{sec:equivalence_proof}
\begin{proof}[Proof of \Cref{col:main_eq}]
Suppose that $\opt(n,n)\le Cn^{1-\eps}$ for a fixed $0<\eps<1$.

For $0<\alpha\le1$, monotonicity in the round budget and the trivial round bound give the exponent $\min\{\alpha,1-\eps\}$. For $\alpha\ge1$, embed the board into a square game with $\lceil n^\alpha\rceil$ cells and rounds, and combine its guarantee with the trivial bound of $n$ signs. Thus, up to a uniform constant, we may use
\[
f(\alpha)=
\begin{cases}
\min\{\alpha,1-\eps\},&0<\alpha\le1,\\
\min\{\alpha(1-\eps),1\},&\alpha\ge1.
\end{cases}
\]
This function satisfies the hypotheses of \Cref{thm:main_upper} and gives
\[
\gamma=\frac{1-\eps}{2-\eps}.
\]
The theorem therefore gives expected calibration error
\[
O\!\left(T^{(4-\eps)/(6-\eps)}\log T\right).
\]
Since
\[
\left(\frac23-\frac{\eps}{18}\right)
-\frac{4-\eps}{6-\eps}
=\frac{\eps^2}{18(6-\eps)}>0,
\]
the logarithm is absorbed into this positive exponent gap. The resulting bound is $O(T^{2/3-\eps/18})$, as claimed.

 The second statement follows immediately from \Cref{thm:adapt_lower}. 
\end{proof}

\subsection{Deferred Proofs for \Cref{sec:upper}}
For $t\in[T]$, let $\mathcal U_t$ be the set of cell and instance pairs $(c,G)$ used to make a prediction by \Cref{alg:upper} in its first $t$ rounds, and let $N_t:=|\mathcal U_t|$. A pair remains in $\mathcal U_t$ after its cell becomes empty.
\begin{lemma}\label{lem:sign_bias}
At the end of every round $t \in [T]$:
\begin{itemize}
    \item The following inequality holds:
\[
\sum_{p \in P} \left|\sum_{s \in [t]} \ind [p_s = p] (e_s - p) \right| \leq \sum_{c, G} |\biasub_t(c,G)|+N_t.
\]
\item For any cell $c$ of an instance $G_{i,j,l}$ (for $i+1 \leq j \leq i+h$)
  \[
\biasub(c, G_{i,j,l}) \in \begin{cases}
    [-1, M] \qquad \text{if cell $c$ contains a plus} \\
    [-M, 1] \qquad \text{if cell $c$ contains a minus} \\
    [-1, 1] \qquad \ \ \text{if cell $c$ is empty}
\end{cases}
\]
where $M = 2^{j-i} + 1$.
\item Both component counters are nonnegative. If a cell contains a plus, its minus component is at most one; if it contains a minus, its plus component is at most one. If the cell is empty, both components are at most one. In particular,
\[
\min\{\biasub_t^+(c,G),\biasub_t^-(c,G)\}\leq 1.
\]
\item For a cell of $G_{i,j,l}$, the magnitude $|\biasub(c,G_{i,j,l})|$ cannot increase during bias removal and increases by at most $2^{-i}$ during bias placement.
\end{itemize}
\end{lemma}
\begin{proof}[Proof of \Cref{lem:sign_bias}]
We first prove the last three statements by induction on the number of rounds. They hold at time zero, since every cell is empty and all component counters are zero. Assume they hold at the end of round $k$.

Suppose round $k+1$ performs bias removal at a cell $\bar c$ of $G=G_{i,j,l}$. If $\bar c$ lies to the right of the cell $c$ containing $e_{k+1}$, the removal test gives $\biasub(\bar c,G)>1$. The plus component is therefore greater than one, and the inductive hypothesis implies that the cell contains a plus and its minus component is at most one. The interval geometry gives
\[
e_{k+1}\leq\frac{2(c-1)+l+1}{2^{i+1}}
\leq\frac{2(\bar c-1)+l-1}{2^{i+1}}
=\probub(\bar c,+,G)=p_{k+1}.
\]
Here $\bar c>c\geq 1$, so the truncation at zero in the definition of $\probub(\bar c,+,G)$ is inactive. The plus component decreases by $p_{k+1}-e_{k+1}\in[0,1]$ and remains nonnegative because it was greater than one. The minus component and all SPR signs are unchanged. Thus the component invariant and the signed bounds are preserved, and the maximum component magnitude cannot increase. The case $\bar c<c$ is symmetric: the minus component is greater than one and decreases by $e_{k+1}-p_{k+1}\in[0,1]$.

Now suppose round $k+1$ performs bias placement at $(c,G)$, where $G=G_{i,j,l}$. Since every applicable bias-removal test failed, each sign erased by a $\simulateGame$ call has its corresponding component at most one. Indeed, a minus to the left with minus component greater than one, or a plus to the right with plus component greater than one, would have triggered bias removal. The opposite component of each such cell is already at most one by the inductive hypothesis. Consequently, both components of every newly empty cell are at most one, and neither counter is reset.

If the target cell was empty, both its components are at most one before the new sign is placed. Whether the target sign is new or was already present, only its corresponding component is increased by the prediction. By the definitions of $\interval$ and $\probub$, this increase is $s(e_{k+1}-p_{k+1})\in[0,2^{-i}]$, where $s$ is the sign in the cell. The opposite component remains at most one. Before the update, the maximum component is less than $2^{j-i}$ by the placement test, so afterward it is at most $2^{j-i}+2^{-i}\leq M$. This proves the component invariant, the signed bounds, and the magnitude-growth statement in the placement case. All other cells are unchanged.

To prove the first statement, group the bias at each prediction value by the cell, instance, and prediction sign to which its rounds were assigned. The triangle inequality and nonnegativity of the components give
\begin{align*}
\sum_{p\in P}\left|\sum_{u\leq t:p_u=p}(e_u-p)\right|
&\leq\sum_{c,G}\bigl(\biasub_t^+(c,G)+\biasub_t^-(c,G)\bigr)\\
&=\sum_{c,G}\left(|\biasub_t(c,G)|+\min\{\biasub_t^+(c,G),\biasub_t^-(c,G)\}\right)\\
&\leq\sum_{c,G}|\biasub_t(c,G)|+N_t.
\end{align*}
The final inequality uses the component invariant and the fact that both counters are zero for pairs outside $\mathcal U_t$.
\end{proof}

Our next objective is to bound the number of rounds played in each simulated game $G$. We first make a simple reduction. Suppose that, in $G$, \PlayerP points to a cell and \PlayerL places a plus there. If \PlayerP's next move in $G$ is to the left of that cell, the plus is immediately erased. The first move was wasted: \PlayerP might as well have gone directly to the second location.

We call the sequence obtained by repeatedly removing these wasted moves the ``reduced transcript.'' Each sign placed in the reduced transcript survives the next move. Thus, before a sign is deleted in this transcript, the pointer must play on both sides of it: first the side that preserves the sign, and later the side that erases it. These visits are what allow us to show that repeatedly placing signs in the same cell requires many calibration rounds.

To formalize this intuition, we make a slight modification to the labeler: it saves its state before each move, including the states of its recursive subroutines. Whenever a new pointer immediately cancels the last recorded move, we rewind to the preceding state, repeating as necessary. Older signs restored by rewinding remain in the labeler's internal record; if the pointer encounters one, we reuse its sign without advancing the strategy. The actual bias counters are unchanged. This gives a legal play of the original strategy whose board contains every actual surviving sign, so its guarantee also bounds the number of signs remaining in the actual game.

\begin{lemma}\label{lem:useful}
For any instance $G=G_{i,j,l}$ with $j\geq i+2$, the reduced transcript at the end of each calibration round $t$ contains at most
\[
\frac{t}{2^{j-1}-2^i}\leq\frac{4t}{2^j}
\]
moves.
\end{lemma}
\begin{proof}
Fix $G=G_{i,j,l}$, and let $G'=G_{i,j-1,l}$ and $q=2^{j-1-i}>1$. These two instances have the same cells and intervals. Whenever the algorithm calls $\simulateGame(c,G)$, it has already failed the placement test at $(c,G')$, so $|\biasub(c,G')|\geq q$ immediately before that call.

Consider the reduced transcript at time $t$, retaining the calibration time at which each of its moves was made. Suppose two consecutive occurrences of a cell $c$ in this transcript were made at times $u<v$. The move following the occurrence at $u$ preserves its sign. Since the occurrence at $v$ points at an empty cell in the reduced game, an intervening move must erase that sign. Thus there are actual calls in $G$ on both sides of $c$ between times $u$ and $v$.

We claim that $|\biasub(c,G')|\leq1$ at some time between $u$ and $v$. At time $u$, its magnitude is at least $q>1$, so \Cref{lem:sign_bias} implies that cell $c$ in $G'$ contains the sign of its unique component exceeding one. If that sign is erased before the two side visits have occurred, both components are at most one at its erasure. Otherwise, one of those visits is on the side that would erase this sign. On that calibration round, the bias-removal test in $G'$ failed, since the algorithm made a call in $G$. The component corresponding to the sign is therefore at most one, and the opposite component is at most one by \Cref{lem:sign_bias}. This proves the claim.

By \Cref{lem:sign_bias}, the magnitude at $(c,G')$ can increase only on bias-placement rounds assigned to that pair, and by at most $2^{-i}$ per round. Rebuilding it from at most one to at least $q$ before time $v$ therefore requires at least
\[
(q-1)2^i=2^{j-1}-2^i
\]
such rounds in $(u,v)$. Before the first occurrence of $c$ in the reduced transcript, the counters start at zero, so reaching $q$ requires at least $q2^i=2^{j-1}$ such rounds.

For each cell, these charges use disjoint intervals: from time zero to its first occurrence, and between successive occurrences. Charges for different cells also use disjoint rounds, since a calibration round is assigned to only one cell and instance. If the reduced transcript has $r$ moves, it follows that
\[
r(2^{j-1}-2^i)\leq t.
\]
Finally, $j\geq i+2$ gives $2^{j-1}-2^i\geq2^{j-2}$, proving the stated bound.
\end{proof}

\begin{corollary}[of \Cref{lem:useful}]
  \label{cor:bound-opt}
For $G=G_{i,j,l}$ with $j\geq i+2$, a deterministic optimal labeler with round budget
\[
M_{i,j}:=\left\lceil\frac{4T}{2^j}\right\rceil
\]
suffices for the reduced game. The number of actual surviving signs at every calibration time $t\leq T$ is at most $\opt(2^i,M_{i,j})$. For $j=i+1$, take the labeler's round budget to be $T$ and use the bound $S_G(t)\leq N_G(t)$, where $S_G(t)$ is the number of actual surviving signs and $N_G(t)$ is the number of distinct cells used in $G$ up to time $t$.
\end{corollary}
\begin{proof}
We may choose deterministic optimal labelers because SPR is a finite perfect-information game. For $j\geq i+2$, \Cref{lem:useful} shows that every reduced transcript has at most $4T/2^j\leq M_{i,j}$ moves. To avoid any issue of defining a response beyond the budget, extend the labeler arbitrarily after $M_{i,j}$ moves. The counting argument uses only the legality of the reduced play and the calibration update rules, so it shows that none of these additional responses is ever used.

Restoring the labeler's full state ensures that the reduced transcript is a legal play of the supplied strategy. Its board contains every actual surviving sign, with the same sign in the same cell. Indeed, when a move is undone, its new sign is eligible for erasure by the incoming pointer, while undoing its other effects only restores older signs. When the pointer encounters a sign already in the restored record, copying it preserves this inclusion. Otherwise, the labeler's query is at an empty recorded cell, and every sign it deletes from its record is also eligible for deletion from the actual board. Thus the inclusion follows by induction over the calls.

Since \PlayerP may stop the game early, the optimal labeler's guarantee applies to every reduced transcript of at most $M_{i,j}$ moves. It bounds the number of signs on the recorded board by $\opt(2^i,M_{i,j})$, and hence also bounds the number on the actual board. At the bottom level, there is at most one new recorded move per calibration round, so budget $T$ suffices. Every actual occupied cell is a used cell, giving $S_G(t)\leq N_G(t)$.
\end{proof}

\begin{lemma}\label{lem:distinct_intervals}
For each $i$, the number of distinct intervals in the $\frac{1}{2^{i+1}}$-discretization that get played by the algorithm is at most $O(\min \{2^i,  2^{\tau - h - i} \})$. Consequently, the number of distinct probability values predicted by the algorithm at this discretization level is bounded by $O(\min \{2^i,  2^{\tau - h - i} \})$. 
\end{lemma}
\begin{proof}
  Let $n$ be the number of intervals in the $\frac{1}{2^{i+1}}$-discretization that get played by the algorithm. 
  By design of \Cref{alg:upper}, at least $n/2$ intervals in the $\frac{1}{2^i}$-discretization must have been played. 
  In turn, in order to play an interval in the $\frac{1}{2^{i}}$-discretization we need that, at level $i' := i-1$, the corresponding cell $c$ for instance $G_{i', i'+h, l}$ (for some $l \in \{0,1\}$) must satisfy $|\biasub(c, G_{i', i'+h,l})| \geq 2^{(i'+h)-i'} = 2^h$. Since $|\biasub(c, G_{i',i'+h,l})|$ increases by at most $2^{-i'}$ on a bias-placement round and cannot increase on a bias-removal round, by \Cref{lem:sign_bias}, it takes $2^{h + i'}$ time steps for its bias to grow to $2^h$. Thus the total number of time steps during which the $n/2$ intervals in the $\frac{1}{2^i}$-discretization were increasing their biases (for instances $G_{i',i'+h,l}$) towards $2^h$ is at least $\frac{n}{2} \cdot 2^{h+i'} \leq T = 2^\tau$. Using $i' = i-1$, we see that $n \leq O(2^{\tau - h - i})$.

  We also have $n \leq 2^{i+1} = O(2^i)$. For any fixed $j$ and $l$, distinct used cells of $G_{i,j,l}$ correspond to distinct intervals, so the same bound applies to the number of distinct used cells in each game. This completes the proof.
\end{proof}

\begin{lemma}\label{lem:calerr_upper}
Suppose $1\le h\le\tau-1$. The total expected calibration error is bounded by
\[
O \left( \sum_{i=1}^{\tau-h} \sum_{j=i+1}^{i+h} 2^{j-i} \opt (2^i, 2^{\tau - j}) + 2^{\frac{3}{4}\tau - \frac{h}{4}} \right)
\]
Note that setting $h = \frac{\tau}{3}$ and assuming the trivial sign player for SPR, we can upper bound the expression by $2^{\frac{2\tau}{3}}$ which is $T^{2/3}$.
\end{lemma}
\begin{proof}
Fix an adversary strategy in the minimax reduction \cite{hart2023minimax}, use deterministic labelers as in \Cref{cor:bound-opt}, and fix all tie-breaking. Let $\mathcal F_{t-1}:=\sigma(H_t)$. Then $p_t,e_t$ are $\mathcal F_{t-1}$-measurable and $\E[y_t\mid\mathcal F_{t-1}]=e_t$. For the horizon parameter $2^\tau$, take the fixed grid
\[
P:=\{k\,2^{-(\tau+1)}:k=0,\ldots,2^{\tau+1}\},
\]
which contains every prediction returned by $\probub$. For $p\in P$, define
\[
B_p:=\sum_{t=1}^T\ind[p_t=p](e_t-p),
\qquad
M_p:=\sum_{t=1}^T\ind[p_t=p](e_t-y_t).
\]
The triangle inequality gives
\[
\E[\calerr(T)]
\leq\E\sum_{p\in P}|B_p|+\E\sum_{p\in P}|M_p|.
\]
We refer to these as the bias and variance terms.

For each fixed $p$, the increments $\ind[p_t=p](e_t-y_t)$ have conditional mean zero given $\mathcal F_{t-1}$, so increments at different times have zero expected product. Consequently,
\begin{align*}
\E\sum_{p\in P}M_p^2
&=\sum_{t=1}^T\sum_{p\in P}
  \E\!\left[\ind[p_t=p](e_t-y_t)^2\right]\\
&=\sum_{t=1}^T\E[e_t(1-e_t)]\leq T.
\end{align*}
Let $A_T:=\{p_1,\ldots,p_T\}$ be the random set of used prediction values. Each used interval at level $i$ supplies at most two prediction values across all choices of $j$, since $\probub(c,s,G_{i,j,l})$ does not depend on $j$. By \Cref{lem:distinct_intervals}, on every run,
\[
|A_T|\leq C\sum_{i=1}^{\tau}\min\{2^i,2^{\tau-h-i}\}
\leq C'2^{(\tau-h)/2}=:K,
\]
for absolute constants $C,C'$. Since $M_p=0$ outside $A_T$, Cauchy--Schwarz gives, pathwise,
\begin{align*}
\sum_{p\in P}|M_p|
&\leq\sqrt{|A_T|}\left(\sum_{p\in P}M_p^2\right)^{1/2}\\
&\leq\sqrt K\left(\sum_{p\in P}M_p^2\right)^{1/2}.
\end{align*}
Taking expectations and applying Jensen's inequality yields
\[
\E\sum_{p\in P}|M_p|
\leq\sqrt{K\,\E\sum_{p\in P}M_p^2}
\leq\sqrt{KT}
=O(2^{3\tau/4-h/4}).
\]
The same estimate applies to the first $T\leq2^\tau$ rounds when the horizon parameter is rounded up.

For a game $G=G_{i,j,l}$, let $S_G$ be the number of cells containing a sign at time $T$, and let $N_G:=|\{c:(c,G)\in\mathcal U_T\}|$ be its number of distinct used cells. By \Cref{lem:sign_bias},
\[
\sum_{c\text{ in }G}\bigl(\biasub_T^+(c,G)+\biasub_T^-(c,G)\bigr)
\leq\sum_{c\text{ in }G}|\biasub_T(c,G)|+N_G
\leq 2^{j-i}S_G+2N_G.
\]
Let $K_i$ be the number of distinct intervals at resolution $2^{-(i+1)}$ used across all the games with discretization index $i$. Since each such interval belongs to at most one cell for each of the $h$ choices of $j$, \Cref{lem:distinct_intervals} gives
\[
N_T=\sum_G N_G\leq h\sum_{i=1}^{\tau}K_i
\leq O\!\left(h\sum_{i=1}^{\tau}\min\{2^i,2^{\tau-h-i}\}\right)
\leq O\!\left(h\,2^{(\tau-h)/2}\right).
\]
This residual contribution is absorbed by the term of order $2^{3\tau/4-h/4}$: since $1\leq h\leq\tau$,
\[
\frac{h\,2^{(\tau-h)/2}}{2^{3\tau/4-h/4}}
=h\,2^{-(\tau+h)/4}\leq h\,2^{-h/2}=O(1).
\]
Only levels $i\le\tau-h$ can be used. Indeed, the first use of a cell and instance pair at a level $i\ge2$ must be for bias placement. Reaching that level requires the corresponding cell in $G_{i-1,i-1+h,l}$ to have bias magnitude at least $2^h$. Since its magnitude starts at zero, cannot increase on removal rounds, and increases by at most $2^{-(i-1)}$ on each placement round, this requires at least $2^{h+i-1}$ preceding calibration rounds. For $i>\tau-h$, this is at least $2^\tau$, which is impossible within the horizon. The assumption $h\le\tau-1$ ensures that the initial level $i=1$ is also included.

For $j=i+1$, \Cref{cor:bound-opt} gives $S_G\leq N_G$, so $2^{j-i}S_G+2N_G\leq4N_G$. For $j\geq i+2$, apply the preserved-sign bound in \Cref{cor:bound-opt}. The total used-cell contribution is at most $4N_T$ and is absorbed as above. Summing over the games therefore bounds the expected bias error by
\[
O\!\left(\sum_{i\leq\tau-h}\sum_{j=i+2}^{i+h}2^{j-i}\opt(2^i,2^{\tau-j+2})+2^{3\tau/4-h/4}\right).
\]
Here $j\leq i+h\leq\tau$, so $M_{i,j}=2^{\tau-j+2}$. Note that $\opt(n,4s)\leq4\opt(n,s)$: split the play into four blocks of at most $s$ rounds and use a fresh optimal $s$-round strategy for the signs placed in each block. A cell empty on the full board is also empty among that block's signs, so these plays are legal; each block leaves at most $\opt(n,s)$ signs, whose number can only decrease afterward. Thus $\opt(2^i,2^{\tau-j+2})\leq4\opt(2^i,2^{\tau-j})$. Adding the nonnegative $j=i+1$ terms and combining with the variance bound above proves the lemma.
\end{proof}

We are now ready to prove \Cref{thm:main_upper}.
\begin{proof}[Proof of \Cref{thm:main_upper}]
Let $C=\max\{1,C_0\}$. For integers $n,s\ge2$, applying the hypothesis with $\alpha=\log_n s$ gives
\[
\opt(n,s)\le C(ns)^\gamma.
\]
This inequality also holds for $s=1$, since $\opt(n,1)\le1$.

For sufficiently large $T$, set $\tau=\lceil\log_2T\rceil$ and use $2^\tau$ as the algorithm's horizon parameter. The preceding estimates apply to its first $T$ rounds. By \Cref{lem:calerr_upper}, its expected calibration error is bounded by
\[
O\!\left(S(\tau,h)+2^{3\tau/4-h/4}\right),
\]
where
\[
S(\tau,h)=\sum_{i=1}^{\tau-h}\sum_{j=i+1}^{i+h}
2^{j-i}\opt(2^i,2^{\tau-j}).
\]
Writing $k=j-i$, we obtain
\begin{align*}
S(\tau,h)
&\le C(\tau-h)2^{\gamma\tau}\sum_{k=1}^{h}2^{(1-\gamma)k}\\
&=O\!\left(\tau\,2^{h+\gamma(\tau-h)}\right).
\end{align*}
The final step is a geometric-series estimate, using $1-\gamma\ge1/2$.

Choose an admissible integer
\[
h=\frac{3-4\gamma}{5-4\gamma}\tau+O(1).
\]
Its coefficient lies between $1/3$ and $3/5$, so $1\le h\le\tau-1$ for all sufficiently large $T$. Substitution gives expected calibration error
\[
O\!\left(T^{\frac{3-2\gamma}{5-4\gamma}}\log T\right).
\]

For $\gamma=1/2$, use the trivial bounds more directly:
\[
\sum_{i=1}^{\tau-h}\opt(2^i,2^{\tau-i-k})
\le\sum_{i=1}^{\tau-h}\min\{2^i,2^{\tau-k-i}\}
=O(2^{(\tau-k)/2}).
\]
Summing over $k$ gives $S(\tau,h)=O(2^{(\tau+h)/2})$. Taking $h=\tau/3+O(1)$ therefore gives $O(T^{2/3})$. Bounded small horizons are absorbed into the implied constant.
\end{proof}

\subsection{Deferred proofs for \Cref{sec:lower}}
\label{sec:adapt_lower_proofs}
\paragraph{Proof overview}
We first show that the SPR simulation finishes before time $T$ with probability $1-o(1)$. We then show that, at its completion time $\tau$, each preserved sign accounts for $\theta/4$ calibration error. The continuation in \Cref{alg:adapt_lower} preserves at least half of $\calerr(\tau)$ through time $T$.

A prediction in an epoch playing cell $i$ is \emph{truthful} if it belongs to $\Int_i$, and \emph{untruthful} otherwise. All estimates below apply also to an unfinished epoch stopped at time $T$. We take $T$ sufficiently large, depending on the fixed parameters $\alpha,\beta$, and suppress routine integer rounding.

\subsubsection{Truthful predictions}
\begin{lemma}[Adaptation of Lemma 11 of \cite{qiao2021stronger}]\label{lem:truthful}
Conditionally on the history at the start of any epoch, except with probability at most $T^{-2}$, the epoch makes at most $(1/2+o(1))T/n^\alpha$ truthful predictions. The same bound holds for every unfinished prefix of that epoch before time $T$.
\end{lemma}
\begin{proof}
Use blocks of $m=T/(324n^\alpha\ln T)$ truthful predictions and $K=162\ln T$ blocks. The block estimate in the proof of Lemma 11 of \cite{qiao2021stronger} applies since $\theta=\sqrt m/80$ and $\mu_i^*\in[1/3,2/3]$. Here is its stopping-time interpretation. Freeze the four prediction classes at the start of a block and give each class an independent tape of Bernoulli outcomes. For each tape, a binomial event of probability at least $1/3$ forces the interval error to reach $\theta$ if that class receives $m/4$ predictions. These events concern fixed initial tape segments, without conditioning on whether the segments are read. Their intersection has probability at least $3^{-4}$; on it, completing a block forces sign placement, since some class receives $m/4$ predictions. Earlier termination, including termination by Condition 2, only shortens the epoch.

Thus the conditional probability of continuing beyond $K$ blocks is at most $(1-3^{-4})^K\le T^{-2}$. Rounding block lengths upward to multiples of four and block counts upward changes $Km$ by $o(T/n^\alpha)$.
\end{proof}

\subsubsection{Untruthful predictions}
For a fixed cell $i$, define
\[
\Psi_t(i)=\sum_{p\in P\cap[0,l_i)}\calD_t^+(p)
             +\sum_{p\in P\cap[r_i,1]}\calD_t^-(p),
\qquad I_t(i)=\sum_{p\in P\cap\Int_i}|\calD_t(p)|,
\]
and write $D_t(i)=\Phi_t(i)-\Psi_t(i)$. Then
\begin{equation}\label{eq:adaptive-potential-decomposition}
\calerr(t)=\Phi_t(i)+\Psi_t(i)+I_t(i)=2\Phi_t(i)-D_t(i)+I_t(i).
\end{equation}
During an epoch playing $i$, a truthful prediction leaves $D_t(i)$ unchanged. For an untruthful prediction, its increment is $y_t-p_t$ when $p_t<l_i$, and $p_t-y_t$ when $p_t\ge r_i$. Thus the increment belongs to $[-1,1]$ and has conditional mean at least $\mu:=1/(6n)$, given the past and the current prediction.

\begin{lemma}[Parameters]\label{lem:params}
The parameters of \Cref{alg:adapt_lower} satisfy
\[
\frac{\theta}{n}=\frac{\ln^2T}{1440},\qquad
\theta n^{\alpha+1}=\frac{T}{1440\ln^3T},\qquad
\frac{T}{n^{\alpha+2}}=\ln^5T.
\]
\end{lemma}
\begin{proof}
Substitute $n^{\alpha+2}=T/\ln^5T$ into the definition of $\theta$.
\end{proof}

We use the following stopped-process calculation for both termination and preservation. Let $A_t\in\{0,1\}$ indicate an active, selected prediction, with activity and selection determined before the outcome at time $t$. Suppose the selected increment $X_t\in[-1,1]$ has conditional mean at least $\mu$ given this pre-outcome information. Starting from any stopping time, put $U_t=\sum_{u\le t}A_uX_u$ and $N_t=\sum_{u\le t}A_u$, where the sums start after that time. Conditional Hoeffding bounds show that, for every $\lambda>0$,
\[
Z_t(\lambda)=\exp\!\left(-\lambda U_t+
                 (\lambda\mu-\lambda^2/2)N_t\right)
\]
is a nonnegative supermartingale starting at $1$. Activity may be set to zero upon stopping. Taking $\lambda=2\mu$ and applying the maximal inequality gives
\begin{equation}\label{eq:adaptive-downcrossing}
\Pr\left[\min_{t\le T}U_t<-\theta/4\right]
\le \exp(-\mu\theta/2)=\exp(-\theta/(12n)).
\end{equation}
For a fixed integer $m$, stop at the $m$th selected observation or at time $T$, whichever comes first. With $\lambda=\mu/2$, the event that the $m$th observation is reached with $U_t<m\mu/2$ implies $Z_t(\lambda)>\exp(m\mu^2/8)$. Optional stopping and Markov's inequality therefore give
\begin{equation}\label{eq:adaptive-count-bound}
\Pr[\text{the $m$th observation is reached with }U_t<m/(12n)]
\le \exp(-m/(288n^2)).
\end{equation}
Both bounds are conditional on the starting history. In particular, they do not condition on a future observation count.

\begin{lemma}[Many untruthful predictions]\label{lem:high-untruthful}
For an epoch playing $i$ and starting after time $u$, except with conditional probability at most
\[
T\exp\!\left(-\frac{T}{1728n^{\alpha+2}}\right)=o(1/T),
\]
every prefix ending at $v\le T$ with $B\ge T/(6n^\alpha)$ untruthful predictions satisfies $D_v(i)-D_u(i)\ge B/(12n)$.
\end{lemma}
\begin{proof}
Select the untruthful predictions and stop activity at the end of the epoch. Apply \Cref{eq:adaptive-count-bound} and take a union bound over $T/(6n^\alpha)\le m\le T$. The final estimate follows from \Cref{lem:params}.
\end{proof}

\begin{lemma}[Uniform preservation within an epoch]\label{lem:low-untruthful}
For an epoch playing $i$ and starting after time $u$, except with conditional probability at most $\exp(-\theta/(12n))=o(1/T)$, every prefix ending at $v\le T$ satisfies $D_v(i)-D_u(i)\ge-\theta/4$.
\end{lemma}
\begin{proof}
Apply \Cref{eq:adaptive-downcrossing} to the same selected increments and use \Cref{lem:params}.
\end{proof}

The stopping conditions give, for every completed epoch or unfinished prefix from $u$ to $v$,
\[
\Phi_v(i)-\Phi_u(i)\le\theta+1,
\qquad I_v(i)-I_u(i)\le\theta+1.
\]
Indeed, before the final outcome neither threshold has been reached, and one outcome changes each quantity by at most one. An epoch with no outcomes has zero increments. Consequently, \Cref{eq:adaptive-potential-decomposition} gives
\begin{equation}\label{eq:adaptive-epoch-increase}
\calerr(v)-\calerr(u)\le 3\theta+3-(D_v(i)-D_u(i)).
\end{equation}

\begin{corollary}\label{cor:calerr-incre}
Except with conditional probability $o(1/T)$, calibration error increases by at most $4\theta$ over every prefix of an epoch.
\end{corollary}
\begin{proof}
Combine \Cref{lem:low-untruthful,eq:adaptive-epoch-increase}; for sufficiently large $T$, $13\theta/4+3\le4\theta$.
\end{proof}

\begin{corollary}\label{lem:untruthful-calerr}
Except with conditional probability $o(1/T)$, every epoch prefix containing $B\ge T/(6n^\alpha)$ untruthful predictions decreases calibration error by at least $B/(12n)-4\theta$.
\end{corollary}
\begin{proof}
Combine \Cref{lem:high-untruthful,eq:adaptive-epoch-increase} and use $3\theta+3\le4\theta$ for sufficiently large $T$.
\end{proof}

\begin{lemma}\label{lem:alg_terminates}
With probability $1-o(1)$, the SPR simulation in \Cref{alg:adapt_lower} finishes before time $T$.
\end{lemma}
\begin{proof}
Take the union of the exceptional events in \Cref{lem:truthful,lem:high-untruthful,lem:low-untruthful} over at most $n^\alpha$ epochs. Its probability is at most
\[
n^\alpha\left(T^{-2}+T e^{-T/(1728n^{\alpha+2})}
                         +e^{-\theta/(12n)}\right)=o(1).
\]
Work outside this event. If the simulation has not finished by time $T$, include the unfinished epoch prefix through time $T$ together with all completed epochs. Otherwise include the completed epochs only. Write $A_j,B_j$ for their truthful and untruthful counts. By \Cref{lem:truthful},
\[
\sum_j A_j\le T/2+o(T).
\]
Call an epoch or prefix long if $B_j\ge T/(6n^\alpha)$. Calibration error starts at zero and remains nonnegative, so \Cref{cor:calerr-incre,lem:untruthful-calerr} imply
\[
\sum_j\II\!\left[B_j\ge\frac{T}{6n^\alpha}\right]
     \left(\frac{B_j}{12n}-4\theta\right)
\le4\theta n^\alpha.
\]
By \Cref{lem:params}, for every long epoch and sufficiently large $T$, $4\theta\le B_j/(24n)$. Therefore
\[
\sum_{j:\,B_j\ge T/(6n^\alpha)}B_j
\le96\theta n^{\alpha+1}=\frac{T}{15\ln^3T}.
\]
The short epochs contribute at most $T/6$ untruthful predictions. It follows that the total number of outcomes used in these epochs and prefixes is at most $2T/3+o(T)<T$. This rules out an unfinished simulation at time $T$ and proves the claim.
\end{proof}

\subsubsection{Preservation and the epoch invariant}
To include the boundary cells without undefined indices, define the rays
\[
R_t(i):=\sum_{p\in P\cap[l_i,1]}\calD_t^+(p),
\qquad L_t(i):=\sum_{p\in P\cap[0,r_i)}\calD_t^-(p).
\]
\begin{lemma}\label{lem:error_preserve}
With probability $1-o(1)$, the following holds simultaneously for all sign placements. If a minus sign is placed in cell $i$ after time $u$, then $R_t(i)\ge R_u(i)-\theta/4$ for every subsequent time $t\le T$ during the SPR simulation while that sign survives. For a plus sign, $L_t(i)\ge L_u(i)-\theta/4$ under the same conditions.
\end{lemma}
\begin{proof}
For a minus sign placed at time $u$, freeze the set
\[
P_i^+:=\{p\in P\cap[l_i,1]:\calE_u(p)>0\},
\qquad W_t:=\sum_{p\in P_i^+}\calE_t(p).
\]
Then $W_u=R_u(i)$ and $R_t(i)\ge W_t$. Because the algorithm removes all legally removable signs before drawing outcomes in each epoch, every later epoch during which this minus survives plays a cell strictly below $i$. Hence its Bernoulli mean is at most $l_i-1/(6n)$, and predicting any $p\in P_i^+$ gives $W_t$ a bounded increment with conditional mean at least $1/(6n)$.

Apply \Cref{eq:adaptive-downcrossing}, selecting predictions in this frozen set and stopping activity at erasure, at the end of the SPR simulation, or at time $T$. The conditional failure probability is at most $\exp(-\theta/(12n))$. For a plus sign, freeze the initially negative predictions in $[0,r_i)$ and track the sum of $-\calE_t(p)$; while it survives, later Bernoulli means are at least $r_i+1/(6n)$. A union bound over at most $n^\alpha$ sign placements completes the proof. No preservation claim is needed during the final continuation.
\end{proof}

\begin{lemma}\label{lem:epoch_invariant}
With probability $1-o(1)$, at the end of every epoch, at time $t$, and for every $k\in[n]$,
\begin{align*}
R_t(k)&\ge\frac{\theta}{4}\,\#\{\text{preserved minus signs in cells }j\ge k\},\\
L_t(k)&\ge\frac{\theta}{4}\,\#\{\text{preserved plus signs in cells }j\le k\}.
\end{align*}
\end{lemma}
\begin{proof}
Work on the event of \Cref{lem:error_preserve} and set $a=\theta/4$. We first prove by induction over placements the stronger assertion that a newly placed minus in cell $i$ has $R_t(i)\ge(M+1)a$, where $M$ counts all preserved minuses at or to the right of $i$, including the new sign. A newly placed plus satisfies the symmetric assertion for $L_t(i)$.

Consider a new minus in cell $i$, in an epoch starting after time $u$ and ending at time $v$. Let $m$ be the number of old surviving minuses strictly to the right of $i$. If $m>0$, let $j>i$ be the nearest such minus, placed at time $w$. Throughout its lifetime, all later pointers lie strictly below $j$. Thus the minuses at or to the right of $j$ are unchanged since its placement, and their number is $m$. The inductive placement margin and \Cref{lem:error_preserve} give
\[
R_u(j)\ge ma,\qquad R_v(j)\ge ma.
\]
If Condition 1 places the new minus, then
\[
R_v(i)\ge R_v(j)+\sum_{p\in P\cap\Int_i}\calD_v^+(p)
\ge ma+\theta/2=(m+2)a.
\]
If Condition 2 places it, then $\Phi_v^+(i)-\Phi_u^+(i)\ge\theta/2$, and $[l_j,1]\subseteq[r_i,1]$ gives
\[
R_v(i)\ge\Phi_v^+(i)
\ge\Phi_u^+(i)+\theta/2
\ge R_u(j)+\theta/2\ge(m+2)a.
\]
When $m=0$, the same inequalities with the neighbor term omitted give $R_v(i)\ge2a$. Since $M=m+1$, this establishes the stronger assertion, including the no-neighbor case. Reflecting left and right and exchanging positive and negative error proves the assertion for a new plus.

Finally, fix any $k$ and choose the nearest preserved minus $j\ge k$, if one exists. Since its placement, the count of minuses at or to its right has been constant. Its extra placement margin pays the loss of at most $a$ from \Cref{lem:error_preserve}; also $R_t(k)\ge R_t(j)$. This proves the first claimed inequality. Choosing the nearest preserved plus $j\le k$ proves the second in exactly the same way. Both inequalities therefore hold regardless of the sign most recently placed.
\end{proof}

\begin{proof}[Proof of \Cref{thm:adapt_lower}]
Choose a deterministic optimal strategy for \PlayerP, available by the perfect-information observation in \Cref{sec:spr}. Its guarantee of at least $C_0n^\beta$ signs holds against every legal sequence of labeler actions. On the joint event of \Cref{lem:alg_terminates,lem:error_preserve}, whose probability is $1-o(1)$, let $\tau<T$ be the completion time. Applying \Cref{lem:epoch_invariant} at $k=1$ for minuses and $k=n$ for pluses gives
\[
\calerr(\tau)\ge R_\tau(1)+L_\tau(n)
\ge\frac{\theta}{4}C_0n^\beta.
\]
The first inequality holds because the two rays count positive and negative error, respectively.

If total positive error at $\tau$ is at least total negative error, the continuation outputs $0$ thereafter, so every $\calE_t(p)$ is nondecreasing and total positive error cannot decrease. Otherwise it outputs $1$, so total negative error cannot decrease. Hence, on every path where the simulation finishes,
\[
\calerr(T)\ge\tfrac12\calerr(\tau).
\]
Nonnegativity on the exceptional event now yields $\E[\calerr(T)]\ge(1-o(1))C_0\theta n^\beta/8$. Writing $\eta=(\beta+1)/(\alpha+2)$, the parameter substitution is
\[
\theta n^\beta=\frac1{1440}T^\eta(\ln T)^{2-5\eta}
=\tilde\Omega(T^\eta),
\]
which proves the theorem.
\end{proof}
\subsection{Concentration inequality}
We state the following corollary of the Azuma-Hoeffding inequality for submartingales, for convenience:
\begin{theorem}[Azuma-Hoeffding inequality; Lemma 13 of \cite{qiao2021stronger}]
  \label{thm:ah}
  Suppose that the random variables $X_1, \ldots, X_m$ satisfy the following for each $t \in [m]$: (a) $X_t \in [-1,1]$ almost surely, and (b) $\E[X_t \mid X_1, \ldots, X_{t-1}] \geq \mu $. Then for any $c < m \mu$, we have
  \begin{align}
\Pr \left[ \sum_{t=1}^m X_t \leq c \right] \leq \exp \left( - \frac{(m\mu - c)^2}{2m} \right)\nonumber.
  \end{align}
\end{theorem}

\section{Deferred proofs for \Cref{sec:oblivious-main}}
\subsection{Proof of \Cref{thm:oblivious-lb}}
\label{sec:oblsign-to-oblcal}
The bulk of the proof of \Cref{thm:oblivious-lb} is accomplished by the following lemma. 
\begin{lemma}\label{lem:lb-main}
   Fix $T \in \BN$, $\ep, v \in (0,1)$, and $\Delta \in (0,1/4]$. Suppose $\MQ \in \Delta([0,1]^T)$  is a distribution satisfying the following properties: 
    \begin{itemize}
        \item With probability $1$ over $(q_1, \ldots, q_T) \sim \MQ$, for all $t \in [T]$ we have $q_t(1-q_t) \ge v$. 
        \item For all $t \in [T]$ and all $p\in[0,1]$, almost surely over $q_{1:t}$,
          \begin{align}
            \Pr_{q_{1:T} \sim \MQ}\left[\forall s> t,~ \sgn(p-q_s)=\sgn(p-q_t)\mid q_{1:t}\right] \ge \epsilon.\label{eq:q-eps-asm}
          \end{align}
        \item For all $p\in [0,1]$, $|\{t \colon |p-q_t|\le \Delta\}| \le \Delta^{-2}$ with probability $1$ over $q_{1:T} \sim \MQ$. 
    \end{itemize}
    Then, any forecaster suffers expected calibration error of at least $\Omega(\epsilon \Delta v\cdot T)$ against the adversarial sequence $X_1, \ldots, X_T$ defined by sampling $q_{1:T} \sim \MQ$ and drawing $X_t \sim \Ber(q_t)$ independently for each $t \in [T]$. 
  \end{lemma}

  \begin{proof}[Proof of \Cref{thm:oblivious-lb}]
First suppose $T<s$. The no-reuse condition implies $s\le n$, so the claimed lower bound is at most $\epsilon T/n<1$. Independent fair bits force expected calibration error at least $1/2$: condition on the history and the last prediction, and write $z=\calE_{T-1}(p_T)+p_T$. The expected absolute error in that bucket is $(|z|+|z-1|)/2\ge1/2$.

Now suppose $T\ge s$, and let $\MQ$ be the distribution of the means in \Cref{alg:oblivious-lb}. Set
\[
v=3/16,\qquad \Delta=\min\{1/(8n),\sqrt{s/(2T)}\}.
\]
Every mean is in $[1/4,3/4]$, so its Bernoulli variance is at least $v$. Distinct grid means are separated by $1/(2n)>2\Delta$, so a fixed prediction is within $\Delta$ of at most one grid mean. Since cells are not reused and each batch has at most $2T/s$ rounds,
\[
|\{t:|p-q_t|\le\Delta\}|\le2T/s\le\Delta^{-2}.
\]
The batch boundaries are deterministic, and observing the means through a round in batch $i$ reveals precisely the pointer prefix through $k_i$. The remaining means in that batch equal its current mean. Thus \Cref{def:wc-pp} implies, almost surely over $q_{1:t}$,
\[
\min\{\Pr[\forall u>t,\ q_u\ge q_t\mid q_{1:t}],
        \Pr[\forall u>t,\ q_u\le q_t\mid q_{1:t}]\}\ge\epsilon.
\]
For $p<q_t$, the first event preserves $\sgn(p-q_t)=-1$; for $p\ge q_t$, the second preserves $\sgn(p-q_t)=+1$. This proves \Cref{eq:q-eps-asm}. Applying \Cref{lem:lb-main}, and using $\Delta\ge\frac18\min\{1/n,\sqrt{s/T}\}$, gives the asserted bound.
\end{proof}

  \subsection{Setup for proof of \Cref{lem:lb-main}}
  Fix $T \in \BN$, and  consider any distribution $\MQ \in \Delta([0,1]^T)$ as in \Cref{lem:lb-main} together with $\ep, v \in (0,1), \ \Delta \in (0,1/4]$. Throughout this section and the following one, we consider the following oblivious adversary for the calibration problem: it first samples $(q_1, \ldots, q_T) \sim \MQ$, then draws $X_t \sim \Ber(q_t)$ for $t \in [T]$, and plays the sequence $X_1, \ldots, X_T$. Recall that we denote the forecasters' predictions by $p_1, \ldots, p_T \in P$, for the fixed finite set $P$ (formally, $p_t$ is a randomized function of $X_1, \ldots, X_{t-1}$). 
  We define the following quantities to aid in the analysis:
  \begin{itemize}[leftmargin=0.4cm]
  \item   For $t \in [T]$ and $p \in [0,1]$, define the random variable $n_t(p) = |\{i \le t\colon p_i=p,\ |p-q_i|\le \Delta\}|$ to denote the number of time steps $i\le t$ at which the forecaster predicts $p$ and $q_i$ is $\Delta$-close to $p$.
  \item  For $p \in [0,1]$, define the random variable $F(p) \in [T]$ to be $F(p) := \max\bigl(\{1\}\cup\{t\in\{2,\ldots,T\}:\sgn(p-q_t)\ne\sgn(p-q_{t-1})\}\bigr)$.
  \end{itemize}

\paragraph{Construction of the potential function} To lower bound the calibration error, we use a potential-function based argument. To construct our potential function, we define a function $h : \BR \to \BR$ by 
\[
  h(x) =
  \begin{cases} 
    \frac{\Delta x^2}{2} & |x| \le \frac{1}{\Delta}\\
    |x|-\frac{1}{2\Delta} & |x| \ge \frac{1}{\Delta}
  \end{cases}
\]
and functions $\Rbias : \BR \times \BR_{\geq 0} \to \BR$, $\Rvar: \BR \times \BR_{\geq 0}  \to \BR$ by
\begin{align}
  \Rbias(x,a) :=  \max\{ (a - \ep/3) \cdot x, \ep x / 3\}, 
  \qquad \Rvar(x, m) := \frac{\epsilon}{12}h(x) + \frac{\epsilon}{12} m\Delta^2|x|\label{eq:rbias-rvar}.
\end{align}
and a function $R : \BR^3 \to \BR$ by 
\begin{align}
  \label{eq:rxam}
R(x,a,m) &=  \Rbias(x,a) + \Rvar(x,m) \\
&= \max\{ (a - \ep/3) \cdot x, \ep x / 3\} +\frac{\epsilon}{12}h(x) + \frac{\epsilon}{12} m\Delta^2|x|. 
\end{align}
Note that, whenever $a \geq 2\ep/3$, we equivalently have
\begin{align}
  \Rbias(x,a) = \max\{ x, 0 \} \cdot a - \frac{\ep}{3} \cdot |x|\label{eq:rbias-alt}.
\end{align}

Let the \emph{history} up to time $t$ be the tuple $H_t:=(p_{1:t},q_{1:t},X_{1:t-1})$. With a slight abuse of notation, we will sometimes write $H_t \sim \MQ$ to denote the process which draws $q_{1:t} \sim \MQ$, draws $X_{1:t-1}$ from the appropriate Bernoulli distributions, and applies the forecaster's algorithm to produce $p_{1:t}$.  Note that for any $p \in [0,1]$, the random variables $\calE_{t-1}(p), n_{t-1}(p)$ are measurable with respect to $H_t$, and that $\calE_t(p), n_t(p)$ are measurable with respect to $(H_t, X_t)$. 
Our potential functions $\Psi_t(p; (H_t,X_t)), \Phi_t(p; H_t)$ (for $t \in [T], p \in [0,1]$) are defined below: 
\begin{align}\label{eq:phit-define}
\Phi_t(p; H_t) = R(\calE_{t-1}(p) \cdot \sgn(p-q_t), \Pr[F(p) \le t \mid q_{1:t}], n_{t-1}(p))
\end{align}
\begin{align}\label{eq:psit-define}
\Psi_t(p; (H_t, X_t)) = R(\calE_{t}(p) \cdot \sgn(p-q_t), \Pr[F(p) \le t \mid q_{1:t}], n_t(p))
\end{align}
To simplify notation, we omit the dependence of $\Phi_t$ on $H_t$ and of $\Psi_t$ on $(H_t, X_t)$; that is, we will write $\Phi_t(p) = \Phi_t(p; H_t)$ and $\Psi_t(p) = \Psi_t(p; (H_t, X_t))$.
  
Overall, we will show that the expected calibration error is lower bounded by $\frac67\sum_{p \in P} \E[\Psi_T(p)]$ (in \Cref{lem:et-psit}); to lower bound this latter quantity, we will argue in two steps:  namely, for $2\le t\le T$, we have 
\begin{align}
  \sum_{p \in P} \E[\Psi_{t}(p)] &\stackrel{\text{\Cref{lem:phi-psi}}}{\geq} \sum_{p \in P} \E[\Phi_t(p)] + \Omega(\epsilon \Delta v) \\ 
  &\stackrel{\text{\Cref{lem:phi1-psi}}}{\geq} \sum_{p \in P} \E[\Psi_{t-1}(p)] + \Omega(\ep \Delta v).
  \label{eq:psi-t-t1}
\end{align}

    \subsubsection{Proof of \Cref{lem:lb-main}}
  \label{sec:obliv-lb-proof}

  Recall the definition of $R(x,a,m)$ in \Cref{eq:rxam}. First, we establish some basic properties of $R$ in the below lemma; the proof is deferred to the end of the section.
\begin{lemma}\label{lem:R-prop}
    Let $a \ge \epsilon$ and $m \in [0,\frac{1}{\Delta^2}]$. Let $x\in \mathbb{R}$ and let $Y$ be a random variable such that $\E[Y] \ge x$. Then,
    \[
    \E[R(Y,a,m)] - R(x,a,m) \ge \epsilon(\E[Y]-x)/6~.
    \]
    If further $Y$ is supported on $[-\Delta^{-1},\Delta^{-1}]$, then,
        \[
        \mathbb{E}[R(Y,a,m)] - R(x,a,m) \ge \frac{\epsilon(\mathbb{E}[Y]-x)}{6} + \frac{\epsilon\Delta \mathrm{Var}(Y)}{24}~.
        \]
\end{lemma}

Recall the definitions of the potential functions $\Psi_t(p), \Phi_t(p)$ in \Cref{eq:psit-define,eq:phit-define}, respectively. Using \Cref{lem:R-prop}, we may now lower bound $\Psi_t(p)$ in terms of $\Phi_t(p)$. 
\begin{lemma}
  \label{lem:phi-psi}
  Fix any $t \in [T]$ and $p \in [0,1]$. Then with probability $1$ over $H_t \sim \MQ$, we have
    \[
    \E[\Psi_t(p)\mid H_t] - \Phi_t(p) \ge \mathds{1}[p_t=p]\cdot \frac{\epsilon\Delta v}{24}~.
    \]
  \end{lemma}
  \begin{proof}
Fix $t,p$ as in the lemma statement and work conditionally on $H_t = (p_{1:t}, q_{1:t}, X_{1:t-1})$, outside the null set where the assumptions fail.
From definition of $R(x,a,m)$ (\Cref{eq:rxam}): 
\begin{align}
  \Psi_t(p)-\Phi_t(p) &= R(\calE_t(p)\sgn(p-q_t), \Pr[F(p) \le t \mid q_{1:t}], n_t(p))\nonumber\\
  & \qquad - R(\calE_{t-1}(p)\sgn(p-q_t), \Pr[F(p) \le t \mid q_{1:t}], n_{t-1}(p))\nonumber\\
  &= R(\calE_t(p)\sgn(p-q_t), \Pr[F(p) \le t \mid q_{1:t}], n_{t-1}(p))\nonumber\\
  &\qquad - R(\calE_{t-1}(p)\sgn(p-q_t), \Pr[F(p) \le t \mid q_{1:t}], n_{t-1}(p)) \nonumber\\&\qquad+ \frac{(n_t(p)-n_{t-1}(p))\epsilon\Delta^2|\calE_t(p)|}{12}~\\
  &= R(\calE_t(p)\sgn(p-q_t), \Pr[F(p) \le t \mid q_{1:t}], n_{t-1}(p))\nonumber\\
  &\qquad - R(\calE_{t-1}(p)\sgn(p-q_t), \Pr[F(p) \le t \mid q_{1:t}], n_{t-1}(p)) \nonumber\\&\qquad+\frac{\mathds{1}[p_t=p,|p-q_t|\le\Delta] \epsilon\Delta^2|\calE_t(p)|}{12}~.\label{eq:psiphi-1}
\end{align}
We would like to take conditional expectation conditioned on $H_t$.
Notice that 
\[
(\calE_t(p) - \calE_{t-1}(p))\cdot \sgn(p-q_t) = (p-X_t)\mathds{1}[p_t=p]\cdot  \sgn(p-q_t)~.
\]
Consequently,
\begin{align}
\mathbb{E}[\calE_t(p)\sgn(p-q_t)\mid H_t]
  =& \calE_{t-1}(p)\sgn(p-q_t) + (p-q_t) \mathds{1}[p_t=p] \sgn(p-q_t) \nonumber\\
  =& \calE_{t-1}(p)\sgn(p-q_t) + |p-q_t| \mathds{1}[p_t=p]  \label{eq:exp-diff}
\end{align}
and due to the assumption that $q_t(1-q_t) \geq v$ with probability $1$,
\begin{align}\label{eq:var-diff}
\var[\calE_t(p)\sgn(p-q_t)\mid H_t]
= \var[X_t\mid H_t] \mathds{1}[p_t=p]
\ge v\mathds{1}[p_t=p]~.
\end{align}
We will apply \Cref{lem:R-prop} to the random variable $Y = \calE_t(p) \cdot \sgn(p-q_t)$ with its conditional distribution given the full history $H_t$, $x$ set to $\calE_{t-1}(p) \cdot \sgn(p-q_t)$, $a$ set to  $ \Pr[F(p) \leq t \mid q_{1:t}]$, and $m$ set to $ n_{t-1}(p)$. We have $m \in [0,1/\Delta^2]$ by assumption in the statement of \Cref{lem:lb-main}, and moreover $a \geq \ep$ almost surely over the draw of $H_t \sim \MQ$, since \Cref{eq:q-eps-asm} can be rephrased as stating that $\Pr_\MQ\left[ F(p) \not \in \{t+1, \ldots, T \} \mid q_{1:t}\right] \geq \ep$, almost surely over the draw of $H_t$. Finally, \Cref{eq:exp-diff} gives that $\E[Y] \geq x$.  Therefore, we may now compute 
\begin{align}
&\mathbb{E}\left[
\Psi_t(p)-\Phi_t(p) \mid H_t 
\right] \nonumber\\
&\ge \frac{\epsilon(\E[\calE_t(p)\sgn(p-q_t)\mid H_t] - \calE_{t-1}(p)\sgn(p-q_t))}{6} 
                      \nonumber\\
  & \qquad + \frac{\mathds{1}[p_t=p]\mathds{1}[|\calE_{t-1}(p)| \le \Delta^{-1}-1] \var[\calE_t(p)\sgn(p-q_t)\mid H_t] \epsilon\Delta}{24}\nonumber \\
&\qquad+ \frac{\mathds{1}[p_t=p,|p-q_t|\le\Delta] \epsilon\Delta^2\E[|\calE_t(p)|\mid H_t]}{12}\nonumber\\
&\ge\mathds{1}[p_t=p] \cdot\nonumber \\
&\quad \left(\frac{\epsilon|p-q_t|}{6} + \frac{\mathds{1}[|\calE_{t-1}(p)| \le \Delta^{-1}-1] v \epsilon\Delta}{24}
+ \frac{\mathds{1}[|p-q_t|\le \Delta] \epsilon\Delta^2\E[|\calE_t(p)|\mid H_t]}{12}\right)\label{eq:psiphi-2}
\end{align}
where the first inequality uses \Cref{lem:R-prop} (in the manner described above) together with \Cref{eq:psiphi-1} and the fact that if $|\calE_{t-1}(p)|\le \Delta^{-1}-1$ then $|\calE_t(p)|\le \Delta^{-1}$, and the second inequality uses \Cref{eq:exp-diff,eq:var-diff}. 
We will analyze the last two terms in the above expression. Notice that 
\[
|\calE_t(p)|
\ge \mathds{1}[|\calE_t(p)| \ge \Delta^{-1}-2](\Delta^{-1}-2)
\ge \mathds{1}[|\calE_{t-1}(p)| \ge \Delta^{-1}-1](\Delta^{-1}-2)~.
\]
Consequently,
\[
\E[|\calE_t(p)|\mid H_t] \ge \mathds{1}[|\calE_{t-1}(p)| \ge \Delta^{-1}-1](\Delta^{-1}-2)~.
\]
Hence,
\begin{align*}
&\frac{\mathds{1}[|\calE_{t-1}(p)| \le \Delta^{-1}-1] v \epsilon\Delta}{24}
+ \frac{\mathds{1}[|p-q_t|\le \Delta] \epsilon\Delta^2\E[|\calE_t(p)|\mid H_t]}{12}\\
&\ge \frac{\mathds{1}[|\calE_{t-1}(p)| \le \Delta^{-1}-1] v \epsilon\Delta}{24} \\
& \qquad + \frac{\mathds{1}[|p-q_t|\le \Delta] \epsilon\Delta^2\mathds{1}[|\calE_{t-1}(p)|\ge \Delta^{-1}-1](\Delta^{-1}-2)}{12}\\
&\ge \min\left(
\frac{ v \epsilon\Delta}{24}, \frac{\mathds{1}[|p-q_t|\le \Delta]\epsilon\Delta^2(\Delta^{-1}-2)}{12}
\right)\\
&\ge \mathds{1}[|p-q_t|\le \Delta] \min\left(
\frac{ v \epsilon\Delta}{24}, \frac{\epsilon\Delta^2(\Delta^{-1}-2)}{12}
\right)\\
&= \mathds{1}[|p-q_t|\le \Delta] \min\left(
\frac{ v \epsilon\Delta}{24}, \frac{\epsilon\Delta(1-2\Delta)}{12}
\right)\\
&\ge \frac{\mathds{1}[|p-q_t|\le \Delta]\epsilon\Delta v}{24}~,
\end{align*}
where the final inequality uses that $\Delta\le 1/4$ and $v \le 1$.
Combining the above display with \Cref{eq:psiphi-2}, we derive that
\begin{align*}
\E[\Psi_t(p)-\Phi_t(p)\mid H_t] & \ge  
\mathds{1}[p_t=p]\left(\frac{\epsilon|p-q_t|}{6}+\frac{\mathds{1}[|p-q_t|\le \Delta]\epsilon\Delta v}{24}\right)\\
&\ge \mathds{1}[p_t=p]\left(\frac{\epsilon\Delta\mathds{1}[|p-q_t|\ge\Delta]}{6}+\frac{\mathds{1}[|p-q_t|\le \Delta]\epsilon\Delta v}{24}\right)\\
&\ge\mathds{1}[p_t=p]\frac{\epsilon\Delta v}{24},
\end{align*}
as desired.
\end{proof}
\begin{lemma}
  \label{lem:phi1-psi}
  Fix any $1 \leq t \leq T-1$ and $p \in [0,1]$. Then almost surely over $(H_t,X_t)$, we have
    \[
    \mathbb{E}[\Phi_{t+1}(p)\mid H_t,X_t] \ge \Psi_t(p)~.
    \]
\end{lemma}
\begin{proof}
  Fix $t,p$ as in the lemma statement and work with $H_{t+1} = (p_{1:t+1}, q_{1:t+1}, X_{1:t})$, outside the null set where the assumptions fail.
  Recall (from \Cref{eq:rxam}) that $R(x,a,m) = \Rbias(x,a) + \Rvar(x,m)$, and note that $\Rvar(x,m) = \Rvar(-x,m)$ for all $x,m$. Moreover, by assumption (in the statement of \Cref{lem:lb-main}) we have that $\Pr_\MQ\left[F(p) \leq t \mid q_{1:t} \right] \geq \ep$ and $\Pr_\MQ\left[ F(p) \leq t+1 \mid q_{1:t+1} \right] \geq \ep$ almost surely over the draw of $q_{1:t+1} \sim \MQ$, which implies, using \Cref{eq:rbias-alt} together with the fact that $|\calE_t(p) \sgn(p-q_{t+1})| = |\calE_t(p) \sgn(p-q_t)|$, that 
    \begin{align}
    \Phi_{t+1}(p)-\Psi_t(p) &= R(\calE_t(p)\sgn(p-q_{t+1}),\Pr[F(p)\le t+1\mid q_{1:t+1}],n_t(p))\nonumber\\
    &\qquad - R(\calE_t(p)\sgn(p-q_{t}),\Pr[F(p)\le t\mid q_{1:t}],n_t(p))\nonumber\\
    &= \max\{\calE_t(p)\sgn(p-q_{t+1}),0\}\Pr[F(p)\le t+1\mid q_{1:t+1}]\nonumber\\
    &\qquad - \max\{\calE_t(p)\sgn(p-q_{t}),0\}\Pr[F(p)\le t\mid q_{1:t}]~.\label{eq:phit1-psit-1}
    \end{align}
    We claim that
    \begin{align}
&\max\{ \calE_t(p) \sgn(p-q_{t+1}), 0 \} \cdot \Pr[F(p) \leq t+1 \mid q_{1:t+1}]\nonumber\\
&\geq \max\{ \calE_t(p) \sgn(p-q_t), 0 \} \cdot \Pr[F(p) \leq t \mid q_{1:t+1}]\label{eq:casework-ineq}. 
    \end{align}
    To establish \Cref{eq:casework-ineq} we consider the following two cases depending on $\sgn(p-q_t), \sgn(p-q_{t+1})$:

    \paragraph{Case 1: $\sgn(p-q_t) = \sgn(p-q_{t+1})$} In this case $F(p)\ne t+1$ with probability $1$ conditioned on $q_{1:t+1}$, consequently
    \begin{align*}
    \max\{\calE_t(p)\sgn(p-q_{t+1}),0\}\Pr[F(p)\le t+1\mid q_{1:t+1}]\\
    = \max\{\calE_t(p)\sgn(p-q_{t}),0\}\Pr[F(p)\le t\mid q_{1:t+1}]
    \end{align*}
    \paragraph{Case 2: $\sgn(p-q_t) \ne \sgn(p-q_{t+1})$} In this case, $\Pr[F(p)\le t \mid q_{1:t+1}]=0$, consequently
    \begin{align*}
    \max\{\calE_t(p)\sgn(p-q_{t+1}),0\}\Pr[F(p)\le t+1\mid q_{1:t+1}]
    \ge 0\\ = \max\{\calE_t(p)\sgn(p-q_{t}),0\}\Pr[F(p)\le t\mid q_{1:t+1}]~.
    \end{align*}
    Hence \Cref{eq:casework-ineq} is established, and we conclude, using \Cref{eq:phit1-psit-1,eq:casework-ineq}, that
    \begin{align*}
    &\E\left[\Phi_{t+1}(p)-\Psi_t(p)  \mid H_t,X_t\right]\\
    &\ge \E\Big[ 
        \max\{\calE_t(p)\sgn(p-q_{t}),0\}\Pr[F(p)\le t\mid q_{1:t+1}]\\
        & \qquad- \max\{\calE_t(p)\sgn(p-q_{t}),0\}\Pr[F(p)\le t\mid q_{1:t}]
    \mid H_t,X_t\Big]\\
    &= \max\{\calE_t(p)\sgn(p-q_{t}),0\} \cdot \left(\E\left[ 
        \Pr[F(p)\le t\mid q_{1:t+1}]
    \mid H_t,X_t\right] - \Pr[F(p)\le t\mid q_{1:t}]\right)\\
    &= 0,
    \end{align*}
    where the final equality uses the tower law and the following conditional independence: given $q_{1:t}$, the outcomes $X_{1:t}$ and the independent private randomness of the forecaster reveal no further information about future means. Thus,
    \begin{align*}
\Pr[F(p) \leq t \mid q_{1:t}] = \Pr[F(p) \leq t \mid H_t, X_t ] = \E \left[ \Pr[F(p) \leq t \mid q_{1:t+1}] \mid H_t, X_t \right].
    \end{align*}
    This completes the proof.
  \end{proof}
  Using the above lemmas, we can lower bound the expected calibration error: 
\begin{lemma}
  \label{lem:et-psit}
  For any $p \in [0,1]$, we have 
\[
    \mathbb{E}[|\calE_T(p)|] \ge \frac{\epsilon \Delta v}{28}\E\left[ \sum_{t=1}^T \mathds{1}[p_t=p]\right]
\]
\end{lemma}
\begin{proof}
  Taking expectation over $H_t \sim \MQ$ in \Cref{lem:phi-psi} gives that for each $t \in [T], p \in [0,1]$, we have $\E[\Psi_t(p) - \Phi_t(p)] \geq \E \left[ \mathds{1}[p_t = p] \cdot \frac{\ep \Delta v}{24}\right]$. Similarly, taking expectation over $(H_t, X_t) \sim \MQ$ in \Cref{lem:phi1-psi} gives that for each $t \in [T-1], p \in [0,1]$, we have $\E[\Phi_{t+1}(p) - \Psi_t(p)] \geq 0$. Combining these facts, summing over $t \in [T]$, and using the fact that $\Phi_1(p) = 0$ for all $p$ gives that 
\begin{align}
\E[\Psi_T(p)] &= \E[\Phi_1(p)] + \sum_{t=1}^{T} \E[\Psi_{t}(p) - \Phi_t(p)] + \sum_{t=1}^{T-1} \E[\Phi_{t+1}(p) - \Psi_t(p)] \\
&\ge \frac{\epsilon \Delta v}{24}\E\left[ \sum_{t=1}^T \mathds{1}[p_t=p]\right].
\end{align}
Note that since $n_T(p)\le \Delta^{-2}$ by assumption in \Cref{lem:lb-main}, \begin{align}
\Psi_T(p) &= R(\calE_T(p)\sgn(p-q_T),\Pr[F(p)\le T],n_T(p))\nonumber\\
&\le R(|\calE_T(p)|,1,\Delta^{-2})\nonumber\\
&\leq |\calE_T(p)| \cdot \max\{ 1, \ep/3 \} + \frac{\ep}{12} |\calE_T(p)| + \frac{\ep}{12} |\calE_T(p)|\nonumber\\  
&\leq \frac{7|\calE_T(p)|}{6}~.\nonumber
\end{align}
Then,
\[
\E[|\calE_T(p)|]
\ge \frac{6\E[\Psi_T(p)]}{7}
\ge \frac{6\epsilon \Delta v}{24 \cdot 7}\E\left[ \sum_{t=1}^T \mathds{1}[p_t=p]\right],
\]
which gives the desired result since $6/(24\cdot7)=1/28$. 
\end{proof}

Finally, \Cref{lem:lb-main} follows as a direct consequence of \Cref{lem:et-psit}:
\begin{proof}[Proof of \Cref{lem:lb-main}]
We sum the lower bound of \Cref{lem:et-psit} over the fixed finite set $p \in P$, noting that for each $t$, $\sum_{p \in P} \mathds{1}[p_t = p] = 1$ almost surely. This yields that the calibration error is lower bounded by $\sum_{p \in P} \E[|\calE_T(p)|] \geq T \cdot \frac{\ep \Delta v}{28}$, as desired.
\end{proof}

\subsection{Proof of \Cref{col:obl_lower}}
\begin{proof}[Proof of \Cref{col:obl_lower}]
Fix $\lambda=0.15229$. For each sufficiently large integer $d$, take $k_d$ to be $\lambda d$ rounded down, and set
\[
s_d={d\choose k_d},\qquad n_d=s_d2^{d-k_d},\qquad
\epsilon_d=2^{-k_d},\qquad H_d=n_d^2s_d.
\]
The construction in \Cref{lem:tree_spr} gives these parameters. Consecutive values of $k_d$ differ by zero or one, and
\[
\frac{H_{d+1}}{H_d}=
\begin{cases}
4\bigl((d+1)/(d+1-k_d)\bigr)^3,&k_{d+1}=k_d,\\
\bigl((d+1)/(k_d+1)\bigr)^3,&k_{d+1}=k_d+1.
\end{cases}
\]
Thus $H_d$ increases to infinity and its consecutive ratios are bounded by a constant depending only on $\lambda$. For any sufficiently large $T$, choose the largest $d$ such that $H_d\le T$. Then $H_d=\Theta(T)$. Applying \Cref{thm:oblivious-lb} at the actual horizon $T$ gives
\[
\E[\calerr(T)]\ge\Omega\!\left(\min\{
\epsilon_dT/n_d,\epsilon_d\sqrt{Ts_d}\}\right)
\ge\Omega(\epsilon_dn_ds_d).
\]
Let $h(x)=-x\log x-(1-x)\log(1-x)$, with base-two logarithms. Stirling's formula and $k_d=\lambda d+O(1)$ give
\begin{align*}
\log T&=d(3h(\lambda)+2-2\lambda)+O(\log d),\\
\log(\epsilon_dn_ds_d)&=d(2h(\lambda)+1-2\lambda)+O(\log d).
\end{align*}
Consequently,
\[
\log\E[\calerr(T)]\ge
\left(1-\frac{h(\lambda)+1}{3h(\lambda)+2-2\lambda}\right)\log T
-O(\log\log T).
\]
The coefficient is approximately $0.5438957625>0.54389$. Its strict slack absorbs the logarithmic term, proving $\E[\calerr(T)]\ge\Omega(T^{0.54389})$ for every sufficiently large $T$.
\end{proof}

\subsection{Missing proofs}
\begin{proof}[Proof of \Cref{lem:R-prop}]
%
%
%
    Fix $a,m$ as in the lemma statement; for simplicity, let us write $R(x) = R(x,a,m)$. We claim that for real numbers $y \geq z$, we have
    \begin{align}
      \label{eq:yx-ineq}
      R(y) - R(z) \geq \frac{\epsilon \cdot (y-z)}{6}.
    \end{align}
    To prove the above inequality, we first note that $R(x)$ is increasing in $x$: this follows from $a \geq \epsilon$ and the fact that for $x \leq 0$, $\frac{\epsilon}{12} \cdot h'(x) + \frac{\epsilon}{12} \cdot m\Delta^2 \frac{d}{dx}|x| \geq -\epsilon/6$, where we have used $m\Delta^2 \leq 1$. Thus, to prove \Cref{eq:yx-ineq}, it suffices to consider the cases $y \geq z \geq 0$ and $0 \geq y \geq z$. For the case $y \geq z \geq 0$, \Cref{eq:yx-ineq} holds since $x \mapsto \frac{\epsilon}{12} h(x) + \frac{\epsilon}{12} m\Delta^2 |x|$ is increasing for $x \geq 0$ and since $\max\{(a-\ep/3) \cdot x, \ep x/3 \} = (a-\ep/3) \cdot x$ for $x \geq 0$. For the case $0 \geq y \geq z$, we note that for $x \leq 0$, $R'(x) \geq \frac{\ep}{3} - \epsilon/12 - m\Delta^2 \epsilon/12 \geq \epsilon/6$, since $|h'(x)| \leq 1$ for all $x$. This establishes \Cref{eq:yx-ineq}.

    Applying \Cref{eq:yx-ineq} with $y = \E[Y]$ and $z = x$, we conclude that 
    \[
    R(\E Y) - R(x) \ge \frac{\epsilon (\E Y - x)}{6}~.
    \]
    Further, since $R$ is convex, by Jensen's inequality,
    $
    \mathbb{E}[R(Y)] \ge R(\E Y)~.
    $ 
    Combining this fact with the above display proves the first part of the lemma.
    
    The second statement follows from the $\epsilon\Delta/12$-strong convexity of $R$ in $[-\Delta^{-1},\Delta^{-1}]$: for any $y,z \in [-\Delta^{-1},\Delta^{-1}]$, 
    \[
    R(z) \ge R(y) + \partial R(y) \cdot (z-y) + \frac{\epsilon \Delta(z-y)^2}{24}~,
    \]
    where $\partial R(y)$ denotes the subgradient of $R$ at $y$. Consequently, almost surely,
    \[
    R(Y)
    \ge R(\E(Y)) + \partial R(\E Y) \cdot (Y-\E Y) + \frac{\epsilon \Delta(Y-\E Y)^2}{24} ~.
    \]
    Taking expectation over $Y$, we get
    \[
    \E R(Y) \ge R(\E(Y)) + \frac{\epsilon \Delta\mathrm{Var}(Y)}{24}~.
    \]
    Applying the first statement of the lemma, we obtain that
    \[
    R(\E Y) - R(x) \ge \frac{\epsilon(\E Y - x)}{6}~.
    \]
    Combining the last two inequalities, the result follows.
\end{proof}


\section{Deferred proofs for \Cref{sec:signflipping-ub}}
\begin{lemma}
  \label{lem:p-alpha-beta}
For any $t, t_1, t_0 \geq 0$ and $p \in [0,1/2]$ for which $t = t_0 + t_1$, $\max\{t_0, t_1\} \geq (1-p) t$, and $\beta \in [0,1]$, we have $t_0^\beta + t_1^\beta \leq (p^\beta + (1-p)^\beta) \cdot t^\beta$. 
\end{lemma}
\begin{proof}
If $t=0$ or $\beta\in\{0,1\}$, the claim is immediate.
Otherwise, put $x:=\min\{t_0,t_1\}/t$. The hypothesis gives
$0\leq x\leq p\leq1/2$.
The function $g(x):=x^\beta+(1-x)^\beta$ is continuous
and nondecreasing on $[0,1/2]$, since
$g'(x)=\beta(x^{\beta-1}-(1-x)^{\beta-1})\geq0$
for $0<x<1/2$. Therefore
\[
t_0^\beta+t_1^\beta=t^\beta g(x)
\leq t^\beta g(p).
\]
\end{proof}
\begin{lemma}
  \label{lem:2-alpha-beta}
For any $t, t_1, t_0 \geq 0$ satisfying $t=t_0 + t_1$ and $\beta \in [0,1]$, we have $t_0^\beta + t_1^\beta \leq 2^{1-\beta} \cdot t^\beta$. 
\end{lemma}
\begin{proof}
Follows immediately from \Cref{lem:p-alpha-beta} with $p=1/2$.
\end{proof}

\begin{lemma}
  \label{lem:ABCf}
For any $A,B,C > 0$ and $\beta \in (0,1)$, the function $f(p) := \frac{A + C p^\beta}{(B+p)^\beta}$ is increasing for $p \in (0, p^\star)$ and decreasing for $p > p^\star$, where $p^\star := \left( \frac{CB}{A} \right)^{1/(1-\beta)}$. 
\end{lemma}
\begin{proof}
It is straightforward to check that $f'(p) > 0$ for $p \in (0,p^\star)$ and $f'(p) < 0$ for $p > p^\star$. 
\end{proof}

\begin{lemma}
  \label{lem:fpab}
  Given $A,B > 0$ and $\beta \in (0,1)$, the function $f(p) := A \cdot (1-p)^\beta + B \cdot p^\beta$ satisfies
  \begin{align}
    \max_{p \in [0,1]} f(p) = B \cdot ((A/B)^{1/(1-\beta)} + 1)^{1-\beta}\nonumber. 
  \end{align}
\end{lemma}
\begin{proof}
  The function $f(p)$ is concave for $p \in [0,1]$ with maximum at $p^\star := \frac{1}{(A/B)^{1/(1-\beta)} + 1}$. The value of its maximum is therefore
\begin{align*}
\frac{A\cdot (A/B)^{\beta/(1-\beta)}+B}{((A/B)^{1/(1-\beta)} + 1)^\beta}= B \cdot \frac{(A/B)^{1/(1-\beta)} + 1}{((A/B)^{1/(1-\beta)} + 1)^\beta} = B \cdot ((A/B)^{1/(1-\beta)} + 1)^{1-\beta}. 
\end{align*}
This completes the proof.
\end{proof}

\begin{lemma}
  \label{lem:geometric-ti}
  For any $\beta \in (0,1)$ and sequence $t_1, \ldots, t_k$ of nonnegative real numbers satisfying $t_{i+1} \geq 2t_i$ for each $i \in [k-1]$, we have
  \begin{align}
\sum_{i=1}^k t_i^\beta \leq \frac{ \left( \sum_{i=1}^k t_i \right)^\beta}{2^\beta-1}\nonumber.
  \end{align}
\end{lemma}
\begin{proof}
  We use induction on $k$, noting that the base case $k=1$ is immediate since $0<2^\beta-1\leq 1$. To establish the inductive step, fix $\ell > 1$, and suppose that the conclusion of the lemma holds for $k = \ell-1$. Note that $t_\ell \geq \sum_{i=1}^{\ell-1} t_i$ by virtue of the assumption that $t_{i+1} \geq 2t_i$ for each $i$.
  If $\sum_{i=1}^{\ell}t_i=0$, the claim is immediate.
  Otherwise, set $p:=t_\ell/\sum_{i=1}^{\ell}t_i\in[1/2,1]$.
  Thus, we may compute
  \begin{align}
    \sum_{i=1}^\ell t_i^\beta \leq & t_\ell^\beta + \frac{ \left( \sum_{i=1}^{\ell-1} t_i \right)^\beta}{2^\beta-1}
    \leq  \left( \frac{\sum_{i=1}^\ell t_i}{2} \right)^\beta + \frac{ \left( \frac{\sum_{i=1}^\ell t_i}{2} \right)^\beta}{2^\beta-1} = \frac{ \left( \sum_{i=1}^\ell t_i \right)^\beta}{2^\beta-1}\nonumber,
  \end{align}
  where the second inequality uses the fact that the function $p \mapsto p^\beta + (1-p)^\beta/(2^\beta-1)$ is continuous and decreasing for $p \in [1/2,1]$ (as it is concave for $p \in [0,1]$ with a global maximum $p^\star \in [0,1/2]$). 
\end{proof}

In the below lemma, we bound a quantity involving the constants $D_{\ref{eq:large-ratio}}, D_{\ref{eq:phase4}}, D_{\ref{eq:b-perp}}, D_{\ref{eq:small-M}}$ defined in \Cref{eq:def-d1,eq:def-d2,eq:def-d3,eq:def-d4}. Note that these quantities depend on $\beta, \delta$, though to keep notation clean we suppress these dependencies. \noah{am thinking of making dependencies explicit, since the $\delta = 1/100$ in the statement of the below lemma doesn't clearly relate to the paramters, as things are written now}
\begin{lemma}
  \label{lem:bound-d-constants}
Fix $\lambda\in(1,2)$ and set $\delta:=1/100$. There are $\beta\in(0,1)$ and $\vep>0$ such that
\begin{align}
\max\Bigg\{&
\max_{p\in[0,1]}
\left(
\frac{D_{\ref{eq:b-perp}}(1-p)^\beta}{2^\beta-1}
+p^\beta\max\{D_{\ref{eq:large-ratio}},D_{\ref{eq:phase4}}\}
\right),
\nonumber\\
&
\max_{p\in[0,9/10]}
\left(
\frac{D_{\ref{eq:b-perp}}(1-p)^\beta}{2^\beta-1}
+p^\beta D_{\ref{eq:small-M}}
\right)
\Bigg\}
\leq2^{1-\beta-\vep}.
\label{eq:D-terms}
\end{align}
\end{lemma}
\begin{proof}
  Fix $\lambda \in (1,2)$.  Note that $D_{\ref{eq:large-ratio}}, D_{\ref{eq:phase4}}, D_{\ref{eq:b-perp}}, D_{\ref{eq:small-M}}$ are still functions of $\beta \in (0,1)$. For fixed $\lambda$, we have
  \begin{align}
\lim_{\beta \uparrow 1}  \frac{\max \left\{ (3/4)^\beta, (1/2)^\beta + (1/4)^\beta + (1/4)^\beta/\lambda, (1+4\lambda/3)/4^\beta \right\}}{2^\beta-1} < 1\label{eq:dbperp-l1}.
  \end{align}
  Thus, there are constants $\nu_0 > 0$ and $\nu_1 < 1$ (depending only on $\lambda$) so that for all $\beta \geq 1-\nu_0$, we have $\frac{D_{\ref{eq:b-perp}}}{2^\beta-1} \leq \nu_1$. 

  We bound each of the terms in the lemma statement in turn:

  \paragraph{Bounding the first term} We observe
\begin{align}
&\max\{ D_{\ref{eq:large-ratio}}, D_{\ref{eq:phase4}} \}\nonumber\\ 
&\leq \max \left\{ \frac{1 + 4\lambda/3}{4^\beta} , \left(\frac{1-\delta}{3}\right)^\beta + \left(\frac{2(1-\delta)}{3} \right)^\beta + \delta^\beta, \max_{p \in [\delta/9, 1]} \left\{ (1-p)^\beta 2^{1-\beta} + \frac{p^\beta}{\lambda} \right\} \right\}\nonumber
\end{align}
Note that 
 \begin{align}
&\lim_{\beta \uparrow 1} \max \left\{ \frac{1+4\lambda/3}{4^\beta},  \max_{p \in [\delta/9, 1]} \left\{ (1-p)^\beta 2^{1-\beta} + p^\beta/\lambda \right\} \right\} \nonumber\\
&< 1 <= \lim_{\beta\uparrow 1} ((1-\delta)/3)^\beta + (2(1-\delta)/3)^\beta + \delta^\beta\nonumber.
  \end{align}
  Thus, combining the above with \Cref{eq:dbperp-l1}, we see that there is some $\eta > 0$ so that so that for all $\beta > 1-\eta$, we have 
  \begin{align}
\max\{ D_{\ref{eq:large-ratio}}, D_{\ref{eq:phase4}} \} \leq ((1-\delta)/3)^\beta + (2(1-\delta)/3)^\beta + \delta^\beta, \qquad \frac{D_{\ref{eq:b-perp}}}{2^\beta-1} \leq \nu_1<1\nonumber.
  \end{align}
  Thus, for all $\beta > 1-\eta$, we have
  \begin{align}
    & \max_{p \in [0,1]} \frac{D_{\ref{eq:b-perp}} \cdot (1-p)^\beta}{2^\beta-1} + p^\beta \cdot \max\{ D_{\ref{eq:large-ratio}}, D_{\ref{eq:phase4}} \} \nonumber\\
    \leq & \max_{p \in [0,1]} \nu_1 \cdot (1-p)^\beta + (((1-\delta)/3)^\beta + (2(1-\delta)/3)^\beta + \delta^\beta) \cdot p^\beta\nonumber\\
    =& (((1-\delta)/3)^\beta + (2(1-\delta)/3)^\beta + \delta^\beta)\nonumber\\ 
    &\quad \cdot \left( \left(\frac{\nu_1}{((1-\delta)/3)^\beta + (2(1-\delta)/3)^\beta + \delta^\beta}\right)^{1/(1-\beta)} + 1\right)^{1-\beta}\nonumber\\
    \leq & (((1-\delta)/3)^\beta + (2(1-\delta)/3)^\beta + \delta^\beta)
    \cdot \left( \nu_1^{1/(1-\beta)} + 1\right)^{1-\beta}\label{eq:main-D-ineq},
  \end{align}
  where the equality uses \Cref{lem:fpab} with $A = \nu_1$ and $B = ((1-\delta)/3)^\beta + (2(1-\delta)/3)^\beta + \delta^\beta$. Recall our choice $\delta := 1/100$. Thus, there is some $\eta' > 0$ so that as long as $\beta > 1-\eta'$ 
  \begin{align}
((1-\delta)/3)^\beta + (2(1-\delta)/3)^\beta + \delta^\beta < 2^{0.99 (1-\beta)}\label{eq:main-limit}.
  \end{align}
  Indeed, \Cref{eq:main-limit} holds because for $\delta = 1/100$, \noah{check} 
  \begin{align}
\lim_{\beta \to 1} \left( ((1-\delta)/3)^\beta + (2(1-\delta)/3)^\beta + \delta^\beta \right)^{1/(1-\beta)} < 2^{0.99}\nonumber.
  \end{align}
  Note that, as long as $1-\beta \leq \frac{ \ln(1/\nu_1)}{\ln(2000)}$, we have
  \begin{align}
\left( \nu_1^{1/(1-\beta)} + 1\right)^{1-\beta} \leq \exp \left((1-\beta) \cdot \nu_1^{1/(1-\beta)}\right) \leq 2^{0.001 \cdot (1-\beta)}\label{eq:nu-bound}.
  \end{align}
  Using our choice $\delta = 1/100$ and   combining \Cref{eq:main-D-ineq,eq:main-limit,eq:nu-bound} gives that for all $\beta > 1 - \min \{ \eta, \eta', \ln(1/\nu_1) / \ln(2000) \}$,
  \begin{align}
\max_{p \in [0,1]} \frac{D_{\ref{eq:b-perp}} \cdot (1-p)^\beta}{2^\beta-1} + p^\beta \cdot \max\{ D_{\ref{eq:large-ratio}}, D_{\ref{eq:phase4}} \} \leq & 2^{0.991(1-\beta)}\nonumber.
  \end{align}
  In particular, for any $\beta > 1 - \min \{ \eta, \eta', \ln(1/\nu_1) / \ln(2000) \}$ there is some $\vep > 0$ for which the left-hand side of the above expression is bounded above by $2^{1-\beta-\vep}$.

  \paragraph{Bounding the second term}
Write
\[
a_\beta:=\frac{D_{\ref{eq:b-perp}}}{2^\beta-1},
\qquad
d_\lambda:=\lim_{\beta\uparrow1}a_\beta<1,
\]
where the strict inequality is \Cref{eq:dbperp-l1}. Since $D_{\ref{eq:small-M}}=2^{1-\beta}$, the functions
$p\mapsto a_\beta(1-p)^\beta+2^{1-\beta}p^\beta$
converge uniformly on $[0,9/10]$ to
$p\mapsto d_\lambda(1-p)+p$. Therefore
\begin{align*}
&\lim_{\beta\uparrow1}\max_{p\in[0,9/10]}
\left(
\frac{D_{\ref{eq:b-perp}}(1-p)^\beta}{2^\beta-1}
+p^\beta D_{\ref{eq:small-M}}
\right)\\
&\qquad =
\max_{p\in[0,9/10]}\bigl(d_\lambda(1-p)+p\bigr)
=\frac{9+d_\lambda}{10}<1.
\end{align*}
Thus the second maximum is less than one for all $\beta$ sufficiently close to one. The first maximum is at most $2^{0.991(1-\beta)}$ for all $\beta$ sufficiently close to one, by the preceding argument. Choosing such a $\beta$ and setting $\vep:=0.009(1-\beta)>0$ proves \Cref{eq:D-terms}.
\end{proof}

\begin{lemma}
  \label{lem:plus-bound}
  Let $t_0, t_1, t_2$ be non-negative integers satisfying $1 \leq t_1 = \lfloor t_0/2\rfloor$ and $t_2 \leq \lceil t_0/2 \rceil$. Write $t = t_0 + t_1 + t_2$, and let $\lambda \in (1,2)$ be given. Then for any $\beta \in (0,1)$, 
  \begin{align}
t_1^\beta + \lambda t_2^\beta \leq t^\beta \cdot \frac{1+4\lambda/3}{4^\beta}\nonumber.
  \end{align}
\end{lemma}
\begin{proof}
Let $\omega = 1$ if $t_0$ is odd else $\omega = 0$.  Write $p_2 := t_2/t_0$, so that $p_2 \leq 1/2 + \omega/(2t_0)$. Note also that $t = t_0 + t_0/2 - \omega/2 + p_2 t_0 = t_0 \cdot (3/2 + p_2 - \omega/(2t_0))$.

The function $f(p) := \frac{(1/2 - \omega/(2t_0))^\beta + \lambda \cdot p^\beta}{(3/2 - \omega/(2t_0) + p)^\beta}$ is increasing for $p \in \left(0, \left(\frac{\lambda \cdot (3/2 - \omega/(2t_0))}{(1/2 - \omega/(2t_0))^\beta}\right)^{1/(1-\beta)} \right)$ by \Cref{lem:ABCf}; in particular, it is increasing for $p \in (0,1)$. Hence
\begin{align}
\frac{t_1^\beta + \lambda t_2^\beta}{t^\beta} = \frac{(1/2 - \omega/(2t_0))^\beta + \lambda p_2^\beta}{(3/2 - \omega/(2t_0) + p_2)^\beta} = f(p_2) \leq f(1/2 + \omega/(2t_0))\nonumber.
\end{align}

  Then we have
  \begin{align}
    \frac{t_1^\beta + \lambda t_2^\beta}{t^\beta} \leq &  \frac{(1/2 - \omega/(2t_0))^\beta + \lambda (1/2 + \omega/(2t_0))^\beta}{2^\beta} \nonumber\\
    = &  \frac{(1-\omega/t_0)^\beta + \lambda (1 + \omega/t_0)^\beta}{4^\beta}\nonumber\\
    \leq & \frac{1 + 4\lambda/3}{4^\beta}\nonumber,
  \end{align}
  where the final inequality uses that either $\omega = 0$, in which case the penultimate expression is bounded above by $\frac{1+\lambda}{4^\beta}$, or else $t_0 \geq 3$, in which case $(1+\omega/t_0)^\beta \leq (4/3)^\beta < 4/3$. 
\end{proof}

\begin{lemma}
  \label{lem:minus-bound}
  For any value $\delta \in (0,1/3)$, the following holds.  Let $t_0, t_1, t_2$ be non-negative integers satisfying $1 \leq t_1 = \lfloor t_0/2\rfloor$ and $t_2 \leq \lceil t_0/2 \rceil$. Write $t = t_0 + t_1 + t_2$, and let $\lambda \in (1,2)$ be given. Define
  \begin{align}
    \label{eq:f-delta-beta}
F_{\beta,\lambda}(\delta) := \max \left\{ ((1-\delta)/3)^\beta + (2(1-\delta)/3)^\beta + \delta^\beta , \max_{p \in [\delta/9, 1]} \left\{ (1-p)^\beta 2^{1-\beta} + p^\beta/\lambda \right\} \right\}
  \end{align}
  Then for any $\beta \in (0,1)$, 
  \begin{align}
t_0^\beta+ t_1^\beta +  t_2^\beta/\lambda  \leq t^\beta \cdot F_{\beta,\lambda}(\delta)\nonumber.
  \end{align}
\end{lemma}
\begin{proof}
  Since $t_1 = \lfloor t_0/2 \rfloor \geq 1$, we have $t_1 \geq t_0/3$. 
  We consider two cases, depending on the value of $t_2$:

  \paragraph{Case 1: $t_2 \geq \delta \cdot t_1$} We have $t_2 \geq \delta \cdot t_1 \geq \delta t_0/3$ and $t \leq 3t_0/2 + t_2$, meaning that  $\frac{t_2}{t-t_2} \geq 2\delta/9$, and so $p_2 := t_2/t \geq \delta/9$. \Cref{lem:2-alpha-beta} gives that $t_0^\beta + t_1^\beta \leq (t_0+t_1)^\beta \cdot 2^{1-\beta} = t^\beta  (1-p_2)^\beta 2^{1-\beta}$, and so we have
  \begin{align}
    t_0^\beta + t_1^\beta + t_2^\beta / \lambda \leq & t^\beta (1-p_2)^\beta 2^{1-\beta} + t^\beta p_2^\beta/\lambda \nonumber\\
    \leq & t^\beta \cdot \max_{p \in [\delta/9, 1]} \left\{ (1-p)^\beta 2^{1-\beta} + p^\beta/\lambda \right\}\nonumber.
  \end{align}

  \paragraph{Case 2: $t_2 < \delta \cdot t_1$}
Write $q_2:=t_2/t$, so that $q_2<\delta$.
Note that $t_1\leq(t_0+t_1)/3=t(1-q_2)/3$.
Thus, in this case, we may compute
  \begin{align}
    t_0^\beta + t_1^\beta + t_2^\beta/\lambda \leq & t_0^\beta + t_1^\beta + t_2^\beta \nonumber\\
    \leq & ((t_0+t_1)/3)^\beta + (2(t_0+t_1)/3)^\beta + t_2^\beta\nonumber\\
    \leq & t^\beta \cdot (((1-q_2)/3)^\beta + (2(1-q_2)/3)^\beta + q_2^\beta)\nonumber,
  \end{align}
  where the second inequality uses \Cref{lem:p-alpha-beta} with $p=1/3$ and total $t_0+t_1$.  It is straightforward to see by differentiating that $q_2 \mapsto  (((1-q_2)/3)^\beta + (2(1-q_2)/3)^\beta + q_2^\beta)$ is increasing for $q_2 \in [0,1/3]$ for any $\beta < 1$, meaning that the above expression may be bounded above as follows: 
  \begin{align}
    t_0^\beta + t_1^\beta + t_2^\beta/\lambda \leq  t^\beta \cdot (((1-\delta)/3)^\beta + (2(1-\delta)/3)^\beta + \delta^\beta)\nonumber,
  \end{align}
  as desired.

\end{proof}


\end{document}